\documentclass[10pt,twocolumn,letterpaper]{article}

\usepackage{iccv}              

\usepackage{graphicx}
\usepackage{amsmath}
\usepackage{amssymb}
\usepackage{booktabs}
\usepackage{tikz}
\usepackage{bbm}
\usepackage{bm}
\usepackage{pifont}
\usepackage{cuted}
\usepackage{colortbl}
\usepackage{algorithm}
\usepackage{algpseudocode}
\usepackage{xifthen}
\usepackage{wrapfig}
\usepackage{multirow}
\usepackage{comment}
\usepackage{xparse}
\usepackage{tikz}
\usetikzlibrary{backgrounds}
\usetikzlibrary{arrows,shapes}
\usetikzlibrary{tikzmark}
\usetikzlibrary{calc}
\usetikzlibrary{backgrounds}
\usetikzlibrary{arrows,shapes}
\usetikzlibrary{calc}
\newcommand{\highlight}[2]{\colorbox{#1!17}{$\displaystyle #2$}}

\usepackage[table]{xcolor}

\usepackage{mathtools, nccmath}

\usepackage{annotate-equations}
\usepackage[skins,listings]{tcolorbox}
\newtcblisting{LTXexample}[1][]{%
  colframe=black!20,
  colback=black!10,
  coltitle=red!50!yellow!3!white,
  bicolor,colbacklower=white,
  fonttitle=\sffamily\bfseries,
  text above listing,
  #1}

\usepackage[labelfont=bf,font=small,skip=3pt]{caption}

\definecolor{iccvblue}{rgb}{0.21,0.49,0.74}
\usepackage[pagebackref,breaklinks,colorlinks,allcolors=iccvblue]{hyperref}

\usepackage[capitalize]{cleveref}
\crefname{section}{Sec.}{Secs.}
\Crefname{section}{Section}{Sections}
\Crefname{table}{Table}{Tables}
\crefname{table}{Tab.}{Tabs.}

\usepackage[accsupp]{axessibility}  %

\newcommand{\customfootnotetext}[2]{{
  \renewcommand{\thefootnote}{#1}
  \footnotetext[0]{#2}}}

\def\paperID{8284} %
\def\confName{ICCV}
\def\confYear{2025}

\title{Ponimator: Unfolding Interactive Pose for Versatile
Human-human \\ Interaction Animation}

\author{
Shaowei Liu$^{*}$\hspace{4mm}
Chuan Guo$^2\dagger$\hspace{4mm}
Bing Zhou$^2\dagger$\hspace{4mm}
Jian Wang$^2\dagger$\hspace{4mm} \\
$^1$University of Illinois Urbana-Champaign\hspace{4mm}
$^2$Snap Inc.\\
{ 
\href{https://stevenlsw.github.io/ponimator/}{\textcolor{magenta}{https://stevenlsw.github.io/ponimator/}}}
}

\begin{document}

\newcommand{\tree}{\bm{\Gamma}}
\newcommand{\shenlong}[1]{\textcolor{magenta}{#1}}
\newcommand{\todocite}[1]{\textcolor{red}{\textit{Citation needed []}}}
\newcommand{\shenlongsay}[1]{\textcolor{blue}{[Shenlong: #1]}}
\newcommand{\sg}[1]{\textcolor{blue}{[Saurabh: #1]}}
\newcommand{\todo}[1]{\textcolor{red}{\textit{TODO: #1}}}
\newcommand{\shaowei}[1]{\textcolor{magenta}{[Shaowei: #1]}}

\newcommand{\imgtile}[2]{
    {\tikz{
    \node[draw=black, draw opacity=1.0, line width=.3mm, fill opacity=0.7,fill=white, inner sep=0pt](gt) at (0, 0) {\includegraphics[width=#2\linewidth]{#1}};
    \node[draw=black, draw opacity=1.0, line width=.3mm, fill opacity=0.7,fill=white, inner sep=0pt](gt) at (-0.1, 0.1) {\includegraphics[width=#2\linewidth]{#1}};
    \node[draw=black, draw opacity=1.0, line width=.3mm, fill opacity=0.7,fill=white, inner sep=0pt](gt) at (-0.2, 0.2) {\includegraphics[width=#2\linewidth]{#1}};
    \node[draw=black, draw opacity=1.0, line width=.3mm, fill opacity=0.7,fill=white, inner sep=0pt](gt) at (-0.3, 0.3) {\includegraphics[width=#2\linewidth]{#1}}; }}
}

\newcommand{\xpar}[1]{\noindent\textbf{#1}\ \ }
\newcommand{\vpar}[1]{\vspace{3mm}\noindent\textbf{#1}\ \ }

\newcommand{\sect}[1]{Section~\ref{#1}}
\newcommand{\sects}[1]{Sections~\ref{#1}}
\newcommand{\eqn}[1]{Equation~\ref{#1}}
\newcommand{\eqns}[1]{Equations~\ref{#1}}
\newcommand{\fig}[1]{Figure~\ref{#1}}
\newcommand{\figs}[1]{Figures~\ref{#1}}
\newcommand{\tab}[1]{Table~\ref{#1}}
\newcommand{\tabs}[1]{Tables~\ref{#1}}

\newcommand{\ignorethis}[1]{}
\newcommand{\norm}[1]{\lVert#1\rVert}
\newcommand{\fcseven}{$\mbox{fc}_7$}

\renewcommand*{\thefootnote}{\fnsymbol{footnote}}

\def\naive{na\"{\i}ve\xspace}
\def\Naive{Na\"{\i}ve\xspace}

\makeatletter
\DeclareRobustCommand\onedot{\futurelet\@let@token\@onedot}
\def\@onedot{\ifx\@let@token.\else.\null\fi\xspace}

\def\iid{\emph{i.i.d}\onedot}
\def\eg{\emph{e.g}\onedot} \def\Eg{\emph{E.g}\onedot}
\def\ie{\emph{i.e}\onedot} \def\Ie{\emph{I.e}\onedot}
\def\cf{\emph{c.f}\onedot} \def\Cf{\emph{C.f}\onedot}
\def\etc{\emph{etc}\onedot} \def\vs{\emph{vs}\onedot}
\def\wrt{w.r.t\onedot} \def\dof{d.o.f\onedot}
\def\etal{\emph{et al}\onedot}
\makeatother

\definecolor{citecolor}{RGB}{34,139,34}
\definecolor{mydarkblue}{rgb}{0,0.08,1}
\definecolor{mydarkgreen}{rgb}{0.02,0.6,0.02}
\definecolor{mydarkred}{rgb}{0.8,0.02,0.02}
\definecolor{mydarkorange}{rgb}{0.40,0.2,0.02}
\definecolor{mypurple}{RGB}{111,0,255}
\definecolor{myred}{rgb}{1.0,0.0,0.0}
\definecolor{mygold}{rgb}{0.75,0.6,0.12}
\definecolor{myblue}{rgb}{0,0.2,0.8}
\definecolor{mydarkgray}{rgb}{0.66,0.66,0.66}

\newcommand{\myparagraph}[1]{\vspace{-6pt}\paragraph{#1}}

\newcommand{\bbR}{{\mathbb{R}}}
\newcommand{\bK}{\mathbf{K}}
\newcommand{\bX}{\mathbf{X}}
\newcommand{\bY}{\mathbf{Y}}
\newcommand{\bk}{\mathbf{k}}
\newcommand{\bx}{\mathbf{x}}
\newcommand{\by}{\mathbf{y}}
\newcommand{\bhy}{\hat{\mathbf{y}}}
\newcommand{\bty}{\tilde{\mathbf{y}}}
\newcommand{\bG}{\mathbf{G}}
\newcommand{\bI}{\mathbf{I}}
\newcommand{\bg}{\mathbf{g}}
\newcommand{\bS}{\mathbf{S}}
\newcommand{\bs}{\mathbf{s}}
\newcommand{\bM}{\mathbf{M}}
\newcommand{\bw}{\mathbf{w}}
\newcommand{\eye}{\mathbf{I}}
\newcommand{\bU}{\mathbf{U}}
\newcommand{\bV}{\mathbf{V}}
\newcommand{\bW}{\mathbf{W}}
\newcommand{\bn}{\mathbf{n}}
\newcommand{\bv}{\mathbf{v}}
\newcommand{\bq}{\mathbf{q}}
\newcommand{\bR}{\mathbf{R}}
\newcommand{\bi}{\mathbf{i}}
\newcommand{\bj}{\mathbf{j}}
\newcommand{\bp}{\mathbf{p}}
\newcommand{\bt}{\mathbf{t}}
\newcommand{\bJ}{\mathbf{J}}
\newcommand{\bu}{\mathbf{u}}
\newcommand{\bB}{\mathbf{B}}
\newcommand{\bD}{\mathbf{D}}
\newcommand{\bz}{\mathbf{z}}
\newcommand{\bP}{\mathbf{P}}
\newcommand{\bC}{\mathbf{C}}
\newcommand{\bA}{\mathbf{A}}
\newcommand{\bZ}{\mathbf{Z}}
\newcommand{\bff}{\mathbf{f}}
\newcommand{\bF}{\mathbf{F}}
\newcommand{\bo}{\mathbf{o}}
\newcommand{\bc}{\mathbf{c}}
\newcommand{\bT}{\mathbf{T}}
\newcommand{\bQ}{\mathbf{Q}}
\newcommand{\bL}{\mathbf{L}}
\newcommand{\bl}{\mathbf{l}}
\newcommand{\ba}{\mathbf{a}}
\newcommand{\bE}{\mathbf{E}}
\newcommand{\bH}{\mathbf{H}}
\newcommand{\bd}{\mathbf{d}}
\newcommand{\br}{\mathbf{r}}
\newcommand{\bb}{\mathbf{b}}
\newcommand{\bh}{\mathbf{h}}

\newcommand{\btheta}{\bm{\theta}}

\newcommand{\bhh}{\hat{\mathbf{h}}}
\newcommand{\ci}{{\cal I}}
\newcommand{\ct}{{\cal T}}
\newcommand{\co}{{\cal O}}
\newcommand{\ck}{{\cal K}}
\newcommand{\cu}{{\cal U}}
\newcommand{\cS}{{\cal S}}
\newcommand{\cQ}{{\cal Q}}
\newcommand{\cT}{{\cal S}}
\newcommand{\cC}{{\cal C}}
\newcommand{\cE}{{\cal E}}
\newcommand{\cF}{{\cal F}}
\newcommand{\cL}{{\cal L}}
\newcommand{\X}{{\cal{X}}}
\newcommand{\Y}{{\cal Y}}
\newcommand{\cH}{{\cal H}}
\newcommand{\cP}{{\cal P}}
\newcommand{\cN}{{\cal N}}
\newcommand{\cU}{{\cal U}}
\newcommand{\cV}{{\cal V}}
\newcommand{\cX}{{\cal X}}
\newcommand{\cY}{{\cal Y}}
\newcommand{\graph}{{\cal H}}
\newcommand{\bayes}{{\cal B}}
\newcommand{\cx}{{\cal X}}
\newcommand{\cg}{{\cal G}}
\newcommand{\cm}{{\cal M}}
\newcommand{\cM}{{\cal M}}
\newcommand{\cG}{{\cal G}}
\newcommand{\cR}{\cal{R}}
\newcommand{\R}{\cal{R}}
\newcommand{\eig}{\mathrm{eig}}

\newcommand{\D}{{\cal D}}
\newcommand{\bfp}{{\bf p}}
\newcommand{\bfd}{{\bf d}}

\newcommand{\cv}{{\cal V}}
\newcommand{\ce}{{\cal E}}
\newcommand{\cy}{{\cal Y}}
\newcommand{\cz}{{\cal Z}}
\newcommand{\cb}{{\cal B}}
\newcommand{\cq}{{\cal Q}}
\newcommand{\cd}{{\cal D}}
\newcommand{\bcf}{{\cal F}}
\newcommand{\cI}{\mathcal{I}}

\newcommand{\ut}{^{(t)}}
\newcommand{\up}{^{(t-1)}}

\newcommand{\bpi}{\boldsymbol{\pi}}
\newcommand{\bphi}{\boldsymbol{\phi}}
\newcommand{\bPhi}{\boldsymbol{\Phi}}
\newcommand{\bmu}{\boldsymbol{\mu}}
\newcommand{\bSigma}{\boldsymbol{\Sigma}}
\newcommand{\bGamma}{\boldsymbol{\Gamma}}
\newcommand{\bbeta}{\boldsymbol{\beta}}
\newcommand{\bomega}{\boldsymbol{\omega}}
\newcommand{\blambda}{\boldsymbol{\lambda}}
\newcommand{\bkappa}{\boldsymbol{\kappa}}
\newcommand{\btau}{\boldsymbol{\tau}}
\newcommand{\balpha}{\boldsymbol{\alpha}}
\def\bgamma{\boldsymbol\gamma}

\newcommand{\prox}{{\mathrm{prox}}}

\newcommand{\pardev}[2]{\frac{\partial #1}{\partial #2}}
\newcommand{\dev}[2]{\frac{d #1}{d #2}}
\newcommand{\dw}{\delta\bw}
\newcommand{\lab}{\mathcal{L}}
\newcommand{\unlab}{\mathcal{U}}
\newcommand{\ind}{1{\hskip -2.5 pt}\hbox{I}}
\newcommand{\ff}[2]{   \cf_{\prec (#1 \rightarrow #2)}}
\newcommand{\vv}[2]{   \cv_{\prec (#1 \rightarrow #2)}}
\newcommand{\dd}[2]{   \delta_{#1 \rightarrow #2}}
\newcommand{\ld}[2]{   \lambda_{#1 \rightarrow #2}}
\newcommand{\en}[2]{  \bD(#1|| #2)}
\newcommand{\ex}[3]{  \bE_{#1 \sim #2}\left[ #3\right]} 
\newcommand{\exd}[2]{  \bE_{#1 }\left[ #2\right]}

\newcommand{\se}[1]{\mathfrak{se}(#1)}
\newcommand{\SE}[1]{\mathbb{SE}(#1)}
\newcommand{\so}[1]{\mathfrak{so}(#1)}
\newcommand{\SO}[1]{\mathbb{SO}(#1)}

\newcommand{\poselow}{\xi}
\newcommand{\pose}{\bm{\poselow}}
\newcommand{\linpose}{\pose^\ell}
\newcommand{\cbpose}{\pose^c}
\newcommand{\rateparam}{v_i}
\newcommand{\bapose}{\bm{\poselow}_i}
\newcommand{\trackingpose}{\bm{\poselow}}
\newcommand{\rotlow}{\omega}
\newcommand{\rot}{\bm{\rotlow}}
\newcommand{\translow}{v}
\newcommand{\trans}{\bm{\translow}}
\newcommand{\hnorm}[1]{\left\lVert#1\right\rVert_{\gamma}}
\newcommand{\lnorm}[1]{\left\lVert#1\right\rVert}
\newcommand{\barate}{v_i}
\newcommand{\trackingrate}{v}
\newcommand{\imgpt}{\mathbf{u}_{i,k,j}}
\newcommand{\mappt}{\mathbf{X}_{j}}
\newcommand{\timet}[1]{\bar{t}_{#1}}
\newcommand{\mf}[1]{\text{MF}_{#1}}
\newcommand{\kmf}[1]{\text{KMF}_{#1}}
\newcommand{\Exp}{\text{Exp}}
\newcommand{\Log}{\text{Log}}

\NewDocumentCommand{\visvideo}{m m m m m m m O{0} O{green}}{%
  \ifnum#8=1
    \fcolorbox{#9}{white}{\includegraphics[width=#7\linewidth]{src_figs/#1/#2/#2.pdf}}%
  \else
    \includegraphics[width=#7\linewidth]{src_figs/#1/#2/#2.pdf}%
  \fi
  & 
  \ifnum#8=2
    \fcolorbox{#9}{white}{\includegraphics[width=#7\linewidth]{src_figs/#1/#2/video/#3.pdf}}%
  \else
    \includegraphics[width=#7\linewidth]{src_figs/#1/#2/video/#3.pdf}%
  \fi
  & 
  \ifnum#8=3
    \fcolorbox{#9}{white}{\includegraphics[width=#7\linewidth]{src_figs/#1/#2/video/#4.pdf}}%
  \else
    \includegraphics[width=#7\linewidth]{src_figs/#1/#2/video/#4.pdf}%
  \fi
  & 
  \ifnum#8=4
    \fcolorbox{#9}{white}{\includegraphics[width=#7\linewidth]{src_figs/#1/#2/video/#5.pdf}}%
  \else
    \includegraphics[width=#7\linewidth]{src_figs/#1/#2/video/#5.pdf}%
  \fi
  & 
  \ifnum#8=5
    \fcolorbox{#9}{white}{\includegraphics[width=#7\linewidth]{src_figs/#1/#2/video/#6.pdf}}%
  \else
    \includegraphics[width=#7\linewidth]{src_figs/#1/#2/video/#6.pdf}%
  \fi \\%
}

\NewDocumentCommand{\viscontact}{m m m m m m O{0 0 0 0}}{%
\includegraphics[width=0.2\linewidth,trim=#7,clip]{src_figs/#1/#2/contact_pose.pdf} &
\includegraphics[width=0.2\linewidth,trim=#7,clip]{src_figs/#1/#2/video/#3.pdf} &
\includegraphics[width=0.2\linewidth,trim=#7,clip]{src_figs/#1/#2/video/#4.pdf} &
\includegraphics[width=0.2\linewidth,trim=#7,clip]{src_figs/#1/#2/video/#5.pdf} &
\includegraphics[width=0.2\linewidth,trim=#7,clip]{src_figs/#1/#2/video/#6.pdf} \\
}

\NewDocumentCommand{\vistm}{m m m m m m m O{0 0 0 0}}{%
    \includegraphics[width=0.19\linewidth,trim=#8,clip]{src_figs/#1/#2/video/#3.pdf} &
    \includegraphics[width=0.19\linewidth,trim=#8,clip]{src_figs/#1/#2/video/#4.pdf} &
    \includegraphics[width=0.19\linewidth,trim=#8,clip]{src_figs/#1/#2/video/#5.pdf} &
    \includegraphics[width=0.19\linewidth,trim=#8,clip]{src_figs/#1/#2/video/#6.pdf} &
    \includegraphics[width=0.19\linewidth,trim=#8,clip]{src_figs/#1/#2/video/#7.pdf} \\
}

\NewDocumentCommand{\vispm}{m m m m m m O{0 0 0 0}}{%
\includegraphics[width=0.19\linewidth,trim=#7,clip]{src_figs/#1/#2/single_pose.pdf} &
\includegraphics[width=0.19\linewidth,trim=#7,clip]{src_figs/#1/#2/video/#3.pdf} &
\includegraphics[width=0.19\linewidth,trim=#7,clip]{src_figs/#1/#2/video/#4.pdf} &
\includegraphics[width=0.19\linewidth,trim=#7,clip]{src_figs/#1/#2/video/#5.pdf} &
\includegraphics[width=0.19\linewidth,trim=#7,clip]{src_figs/#1/#2/video/#6.pdf} \\
}

\definecolor{Gray}{rgb}{0.9,0.9,0.9}
\newcolumntype{g}{>{\columncolor{Gray}}c}

\definecolor{iccvblue}{rgb}{0.21,0.49,0.74}

\newcommand{\supp}{{Supp. }}
\maketitle

\setlength{\fboxrule}{1.5pt}   %
\setlength{\fboxsep}{0pt}  

\begin{strip}
    \vspace{-3em}
    \centering
    \footnotesize
    \setlength{\tabcolsep}{0.1em} %
    \resizebox{\linewidth}{!}{
    \begin{tabular}{c|cccc}

        {\textbf{\normalsize Input}} & \multicolumn{4}{c}{{\textbf{\normalsize Generated Interaction Animation} (left$\rightarrow$right: time steps)}} \\

    \includegraphics[width=0.19\linewidth]{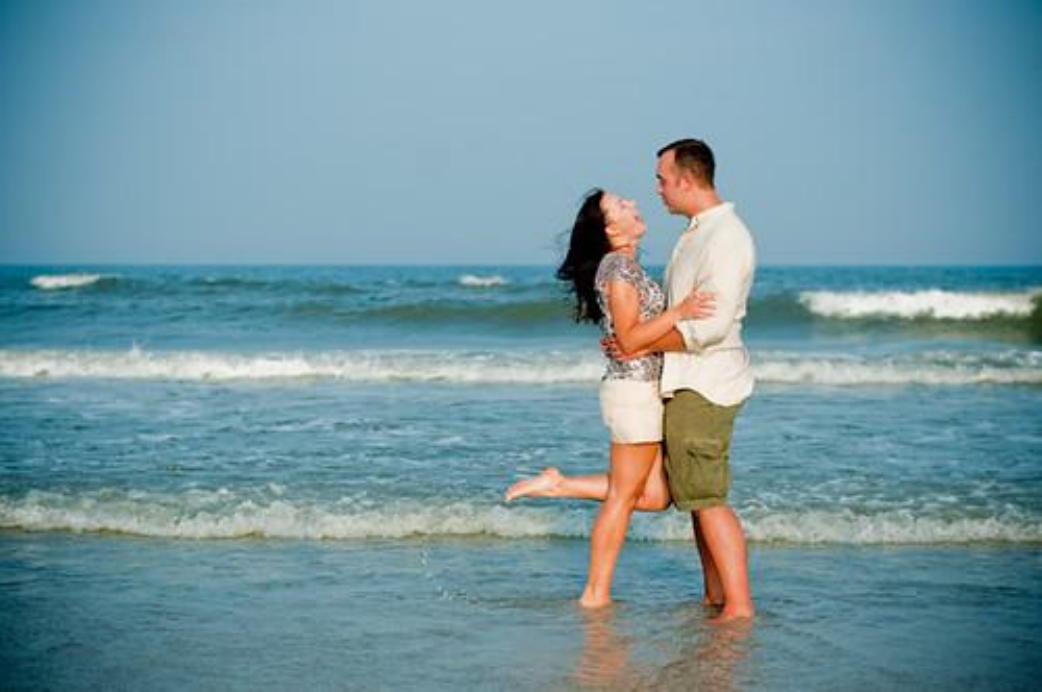} & 
    \includegraphics[width=0.19\linewidth]{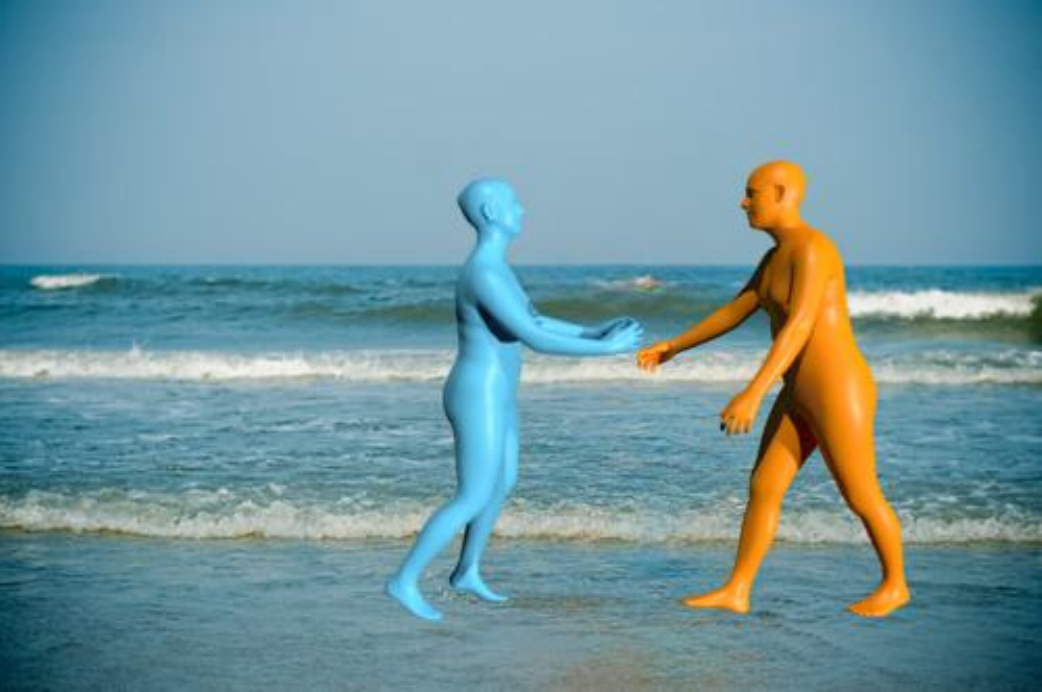}
    & 
     \includegraphics[width=0.19\linewidth]{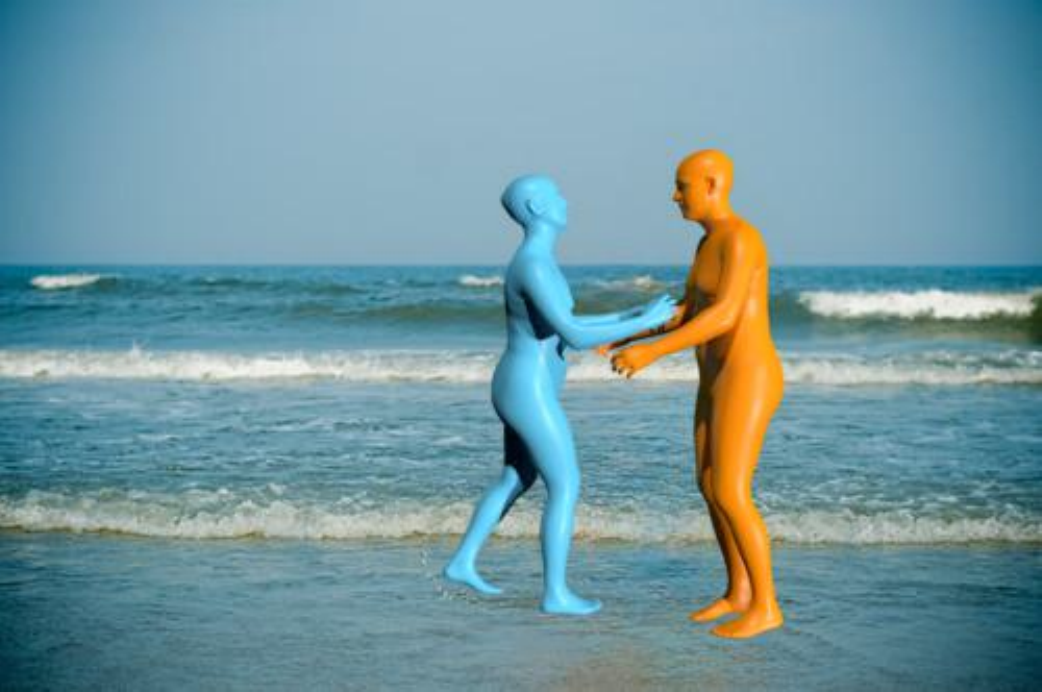} & 
      \fcolorbox{green}{white}{%
      \includegraphics[width=0.19\linewidth]{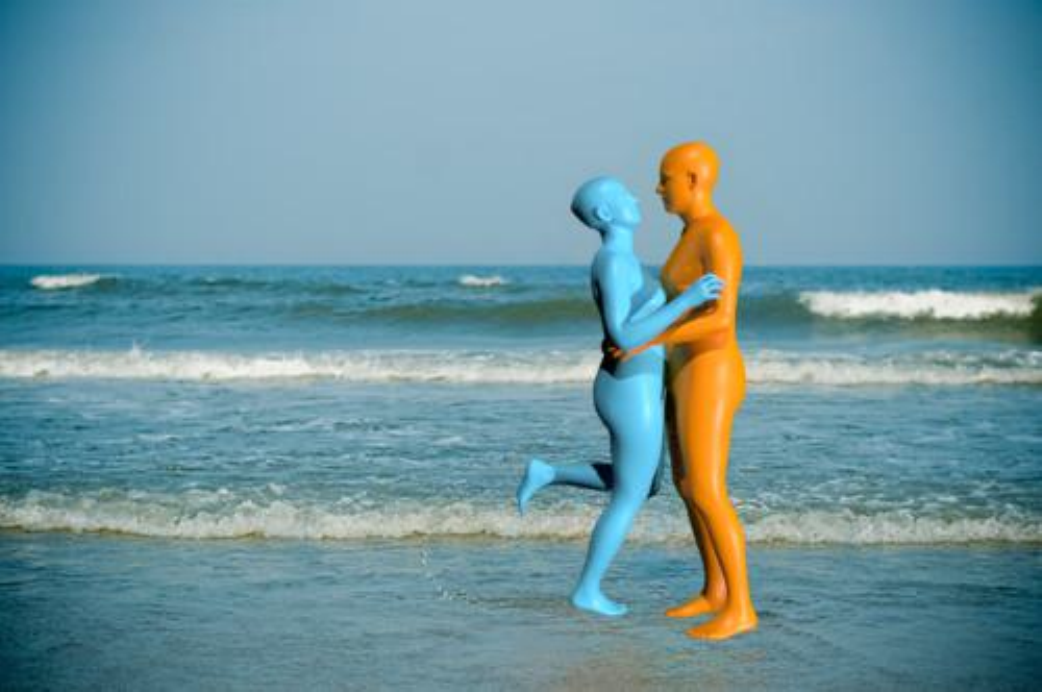}} & 
       \includegraphics[width=0.19\linewidth]{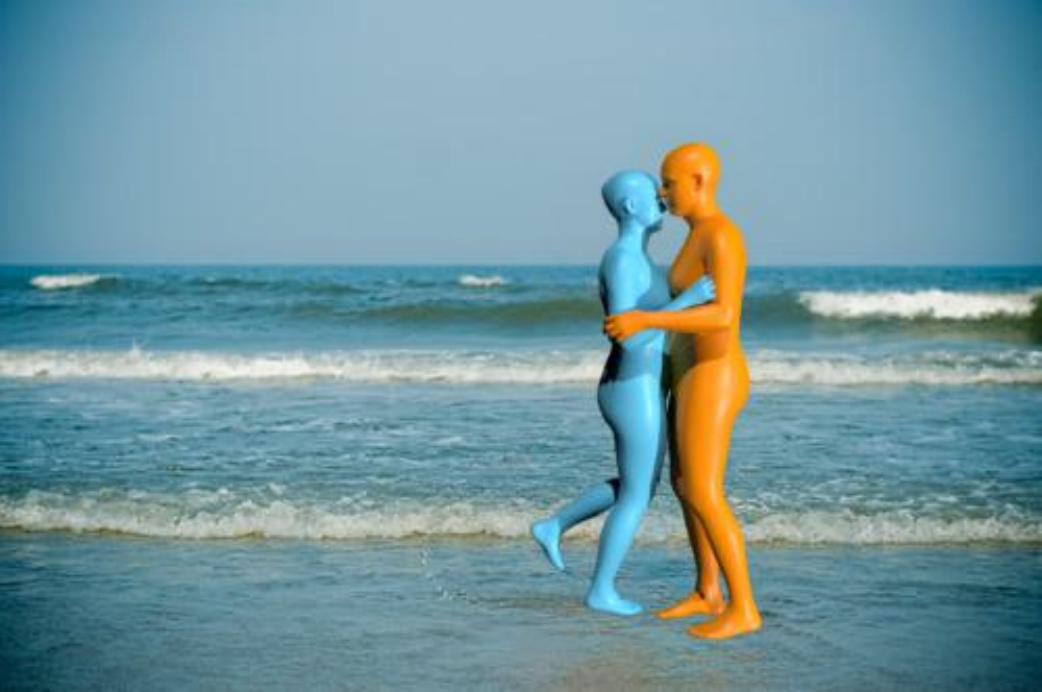} \\

       Two-person image  &  &  & Estimated interactive pose \\

     \includegraphics[width=0.19\linewidth]{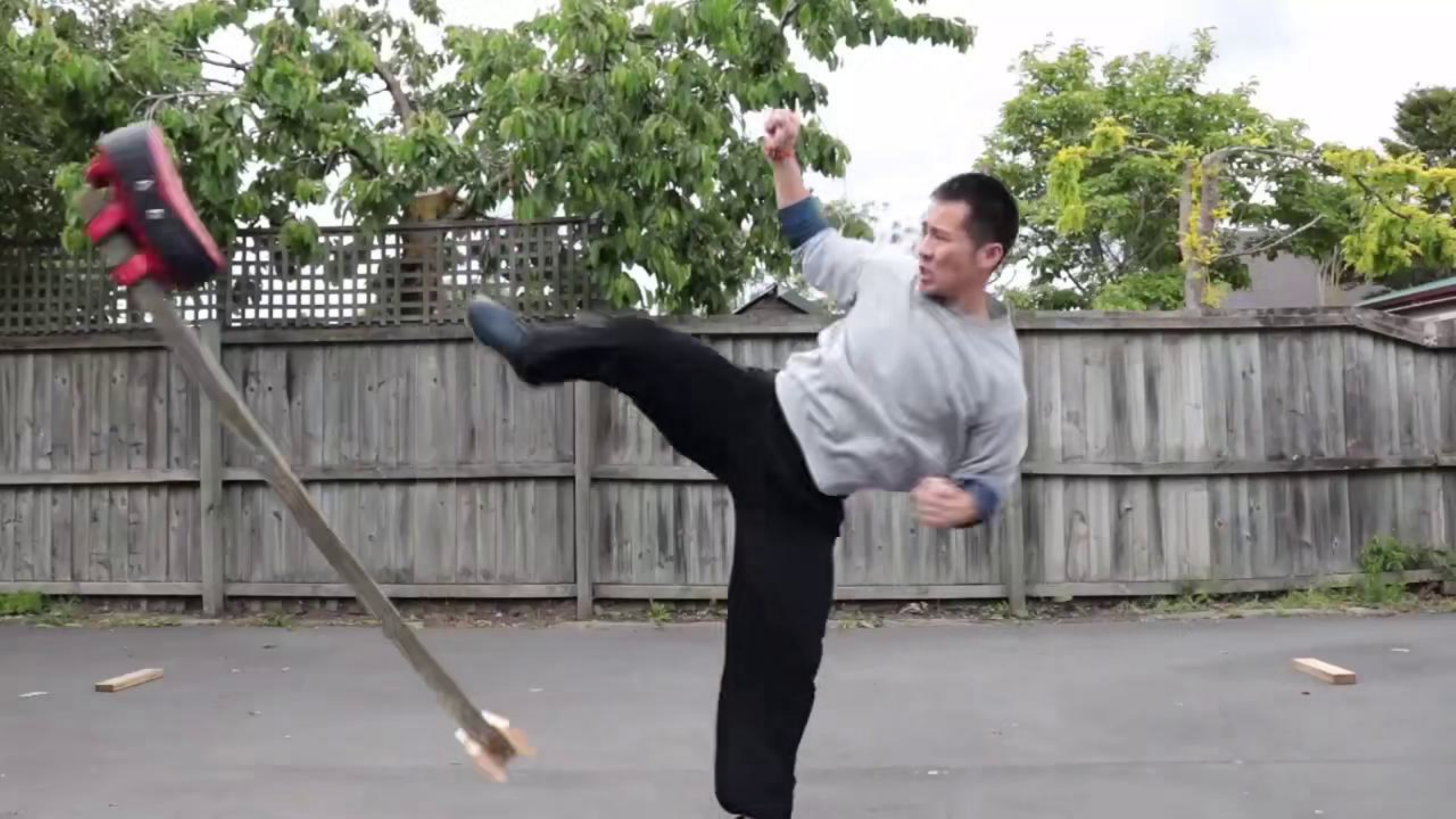} &
     \includegraphics[width=0.19\linewidth]{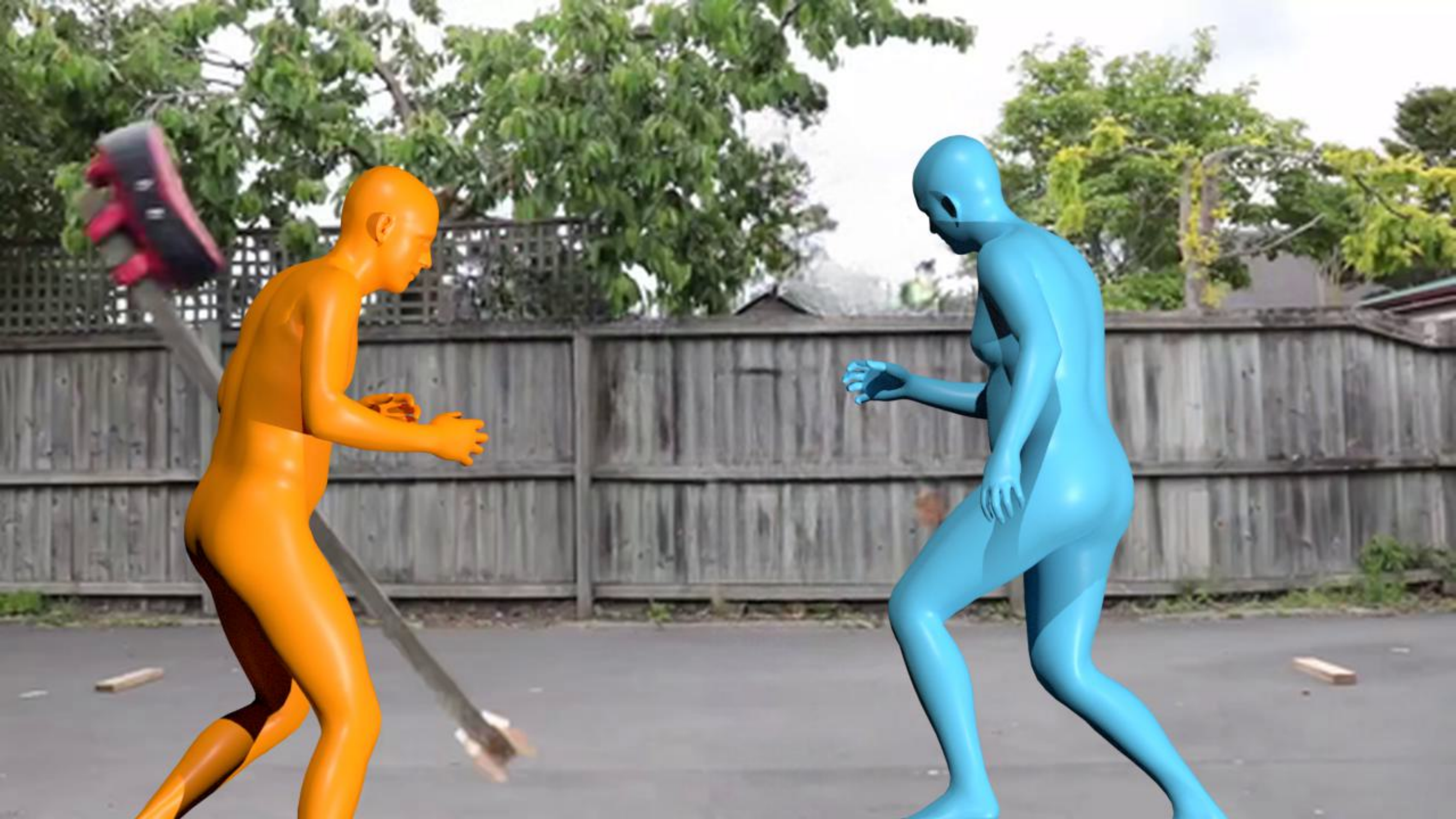} &
      \includegraphics[width=0.19\linewidth]{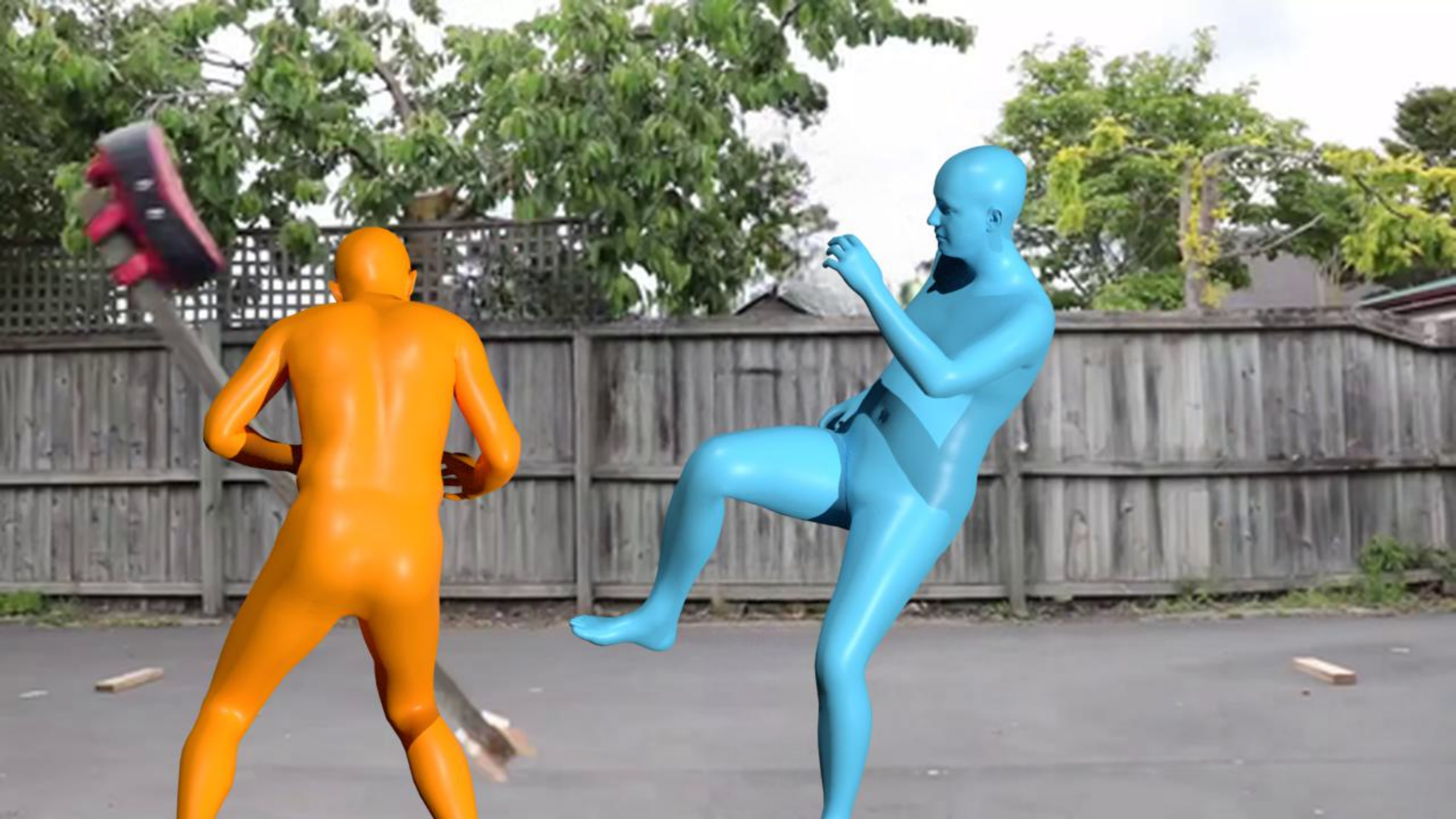} &
      \fcolorbox{magenta}{white}{\includegraphics[width=0.19\linewidth]{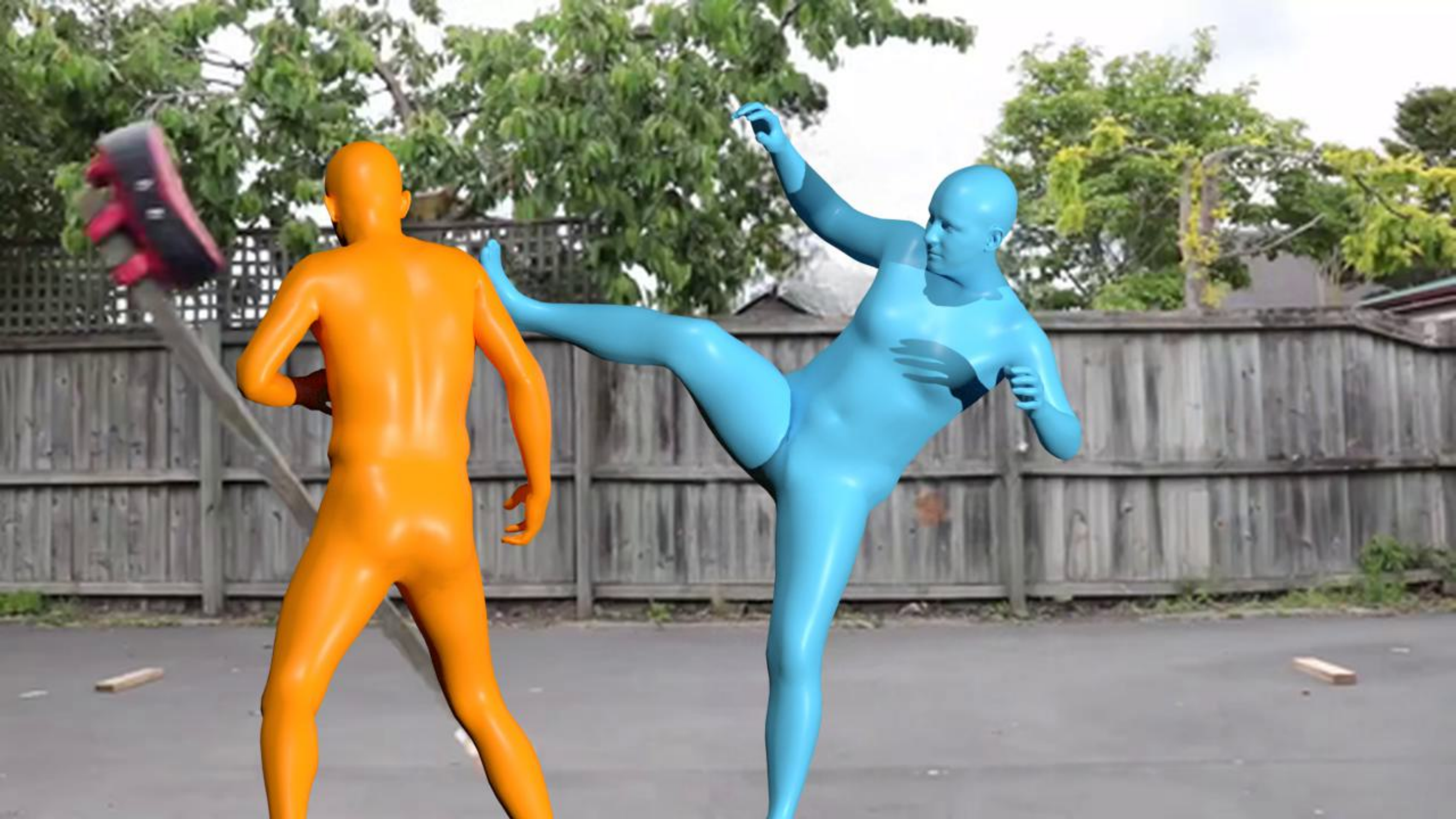}} &
      \includegraphics[width=0.19\linewidth]{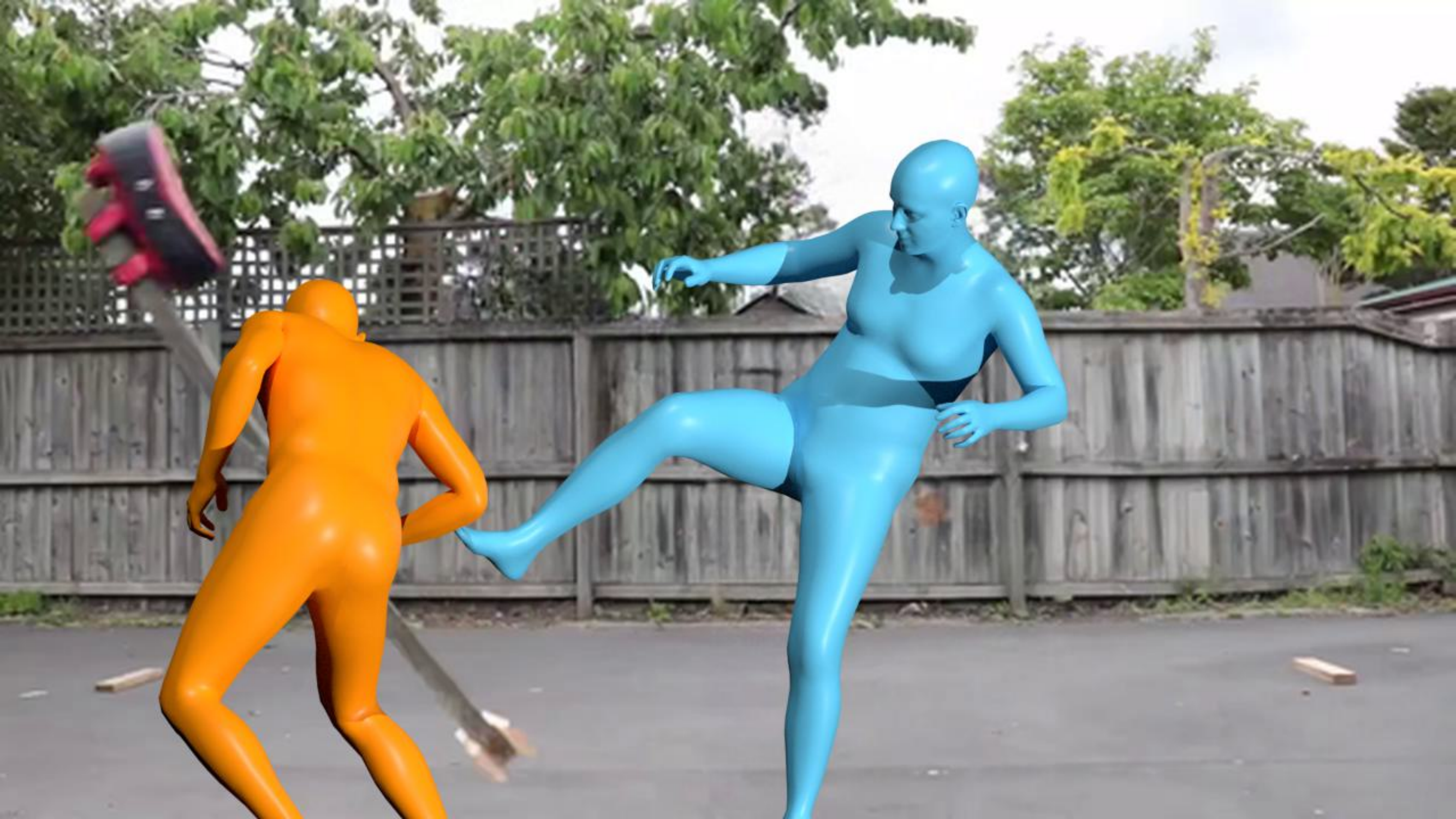} \\

       Single-person image  &  &  & Generated interactive pose \\

    \includegraphics[width=0.19\linewidth]{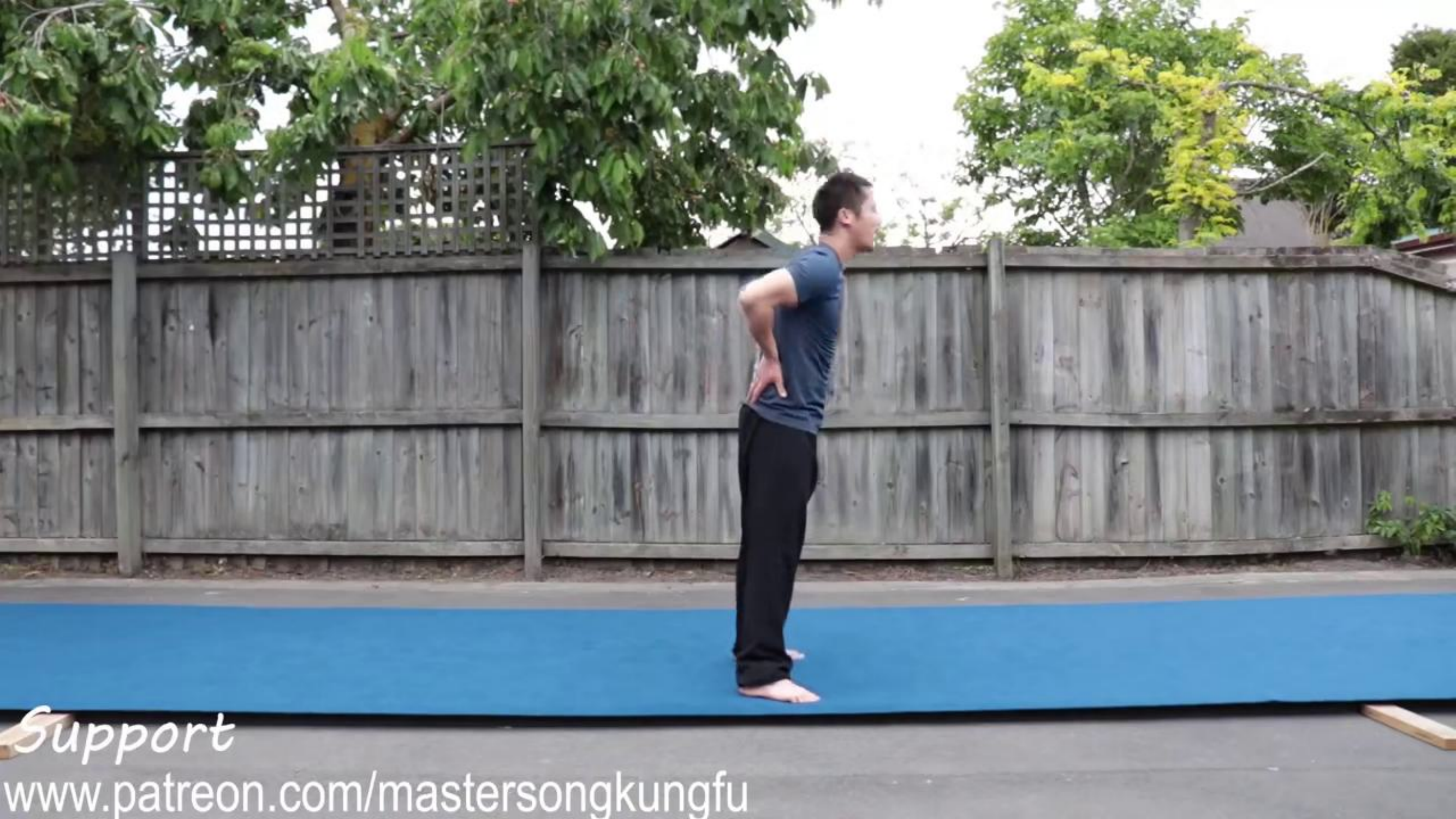} &
    \includegraphics[width=0.19\linewidth]{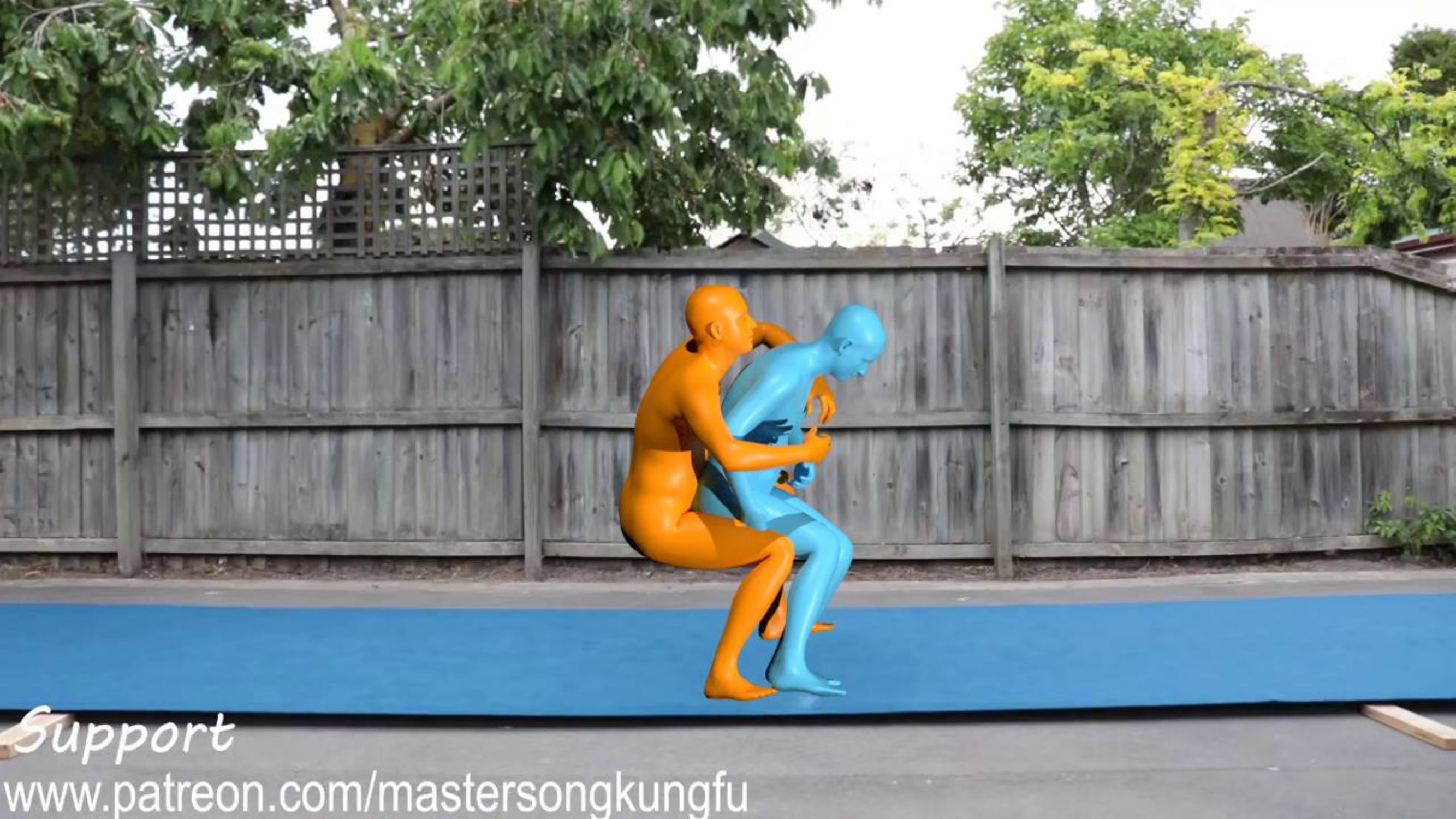} &
     \includegraphics[width=0.19\linewidth]{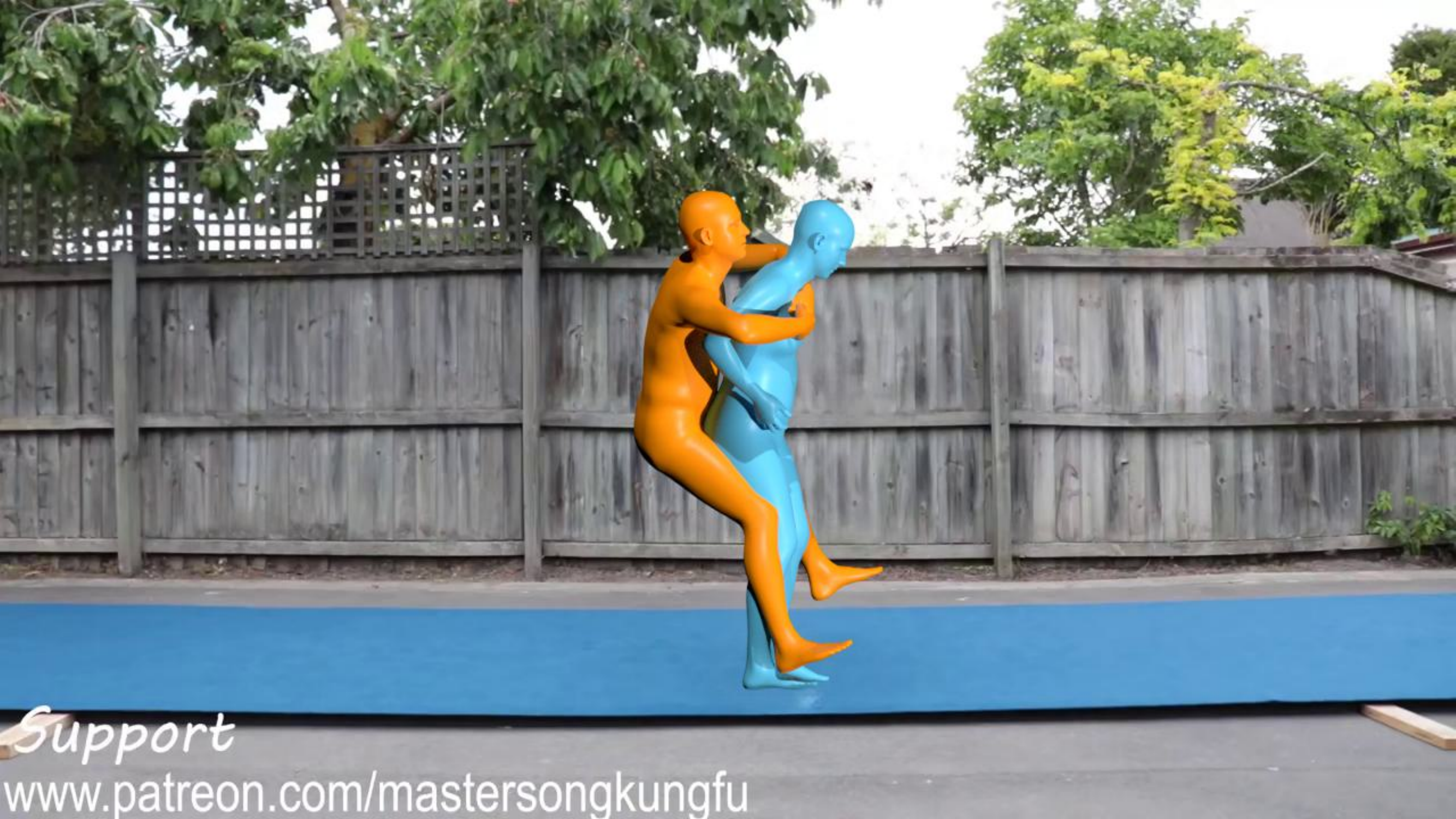} &
      {\setlength{\fboxsep}{-1.5pt}\fcolorbox{magenta}{white}{
      \includegraphics[width=0.19\linewidth]{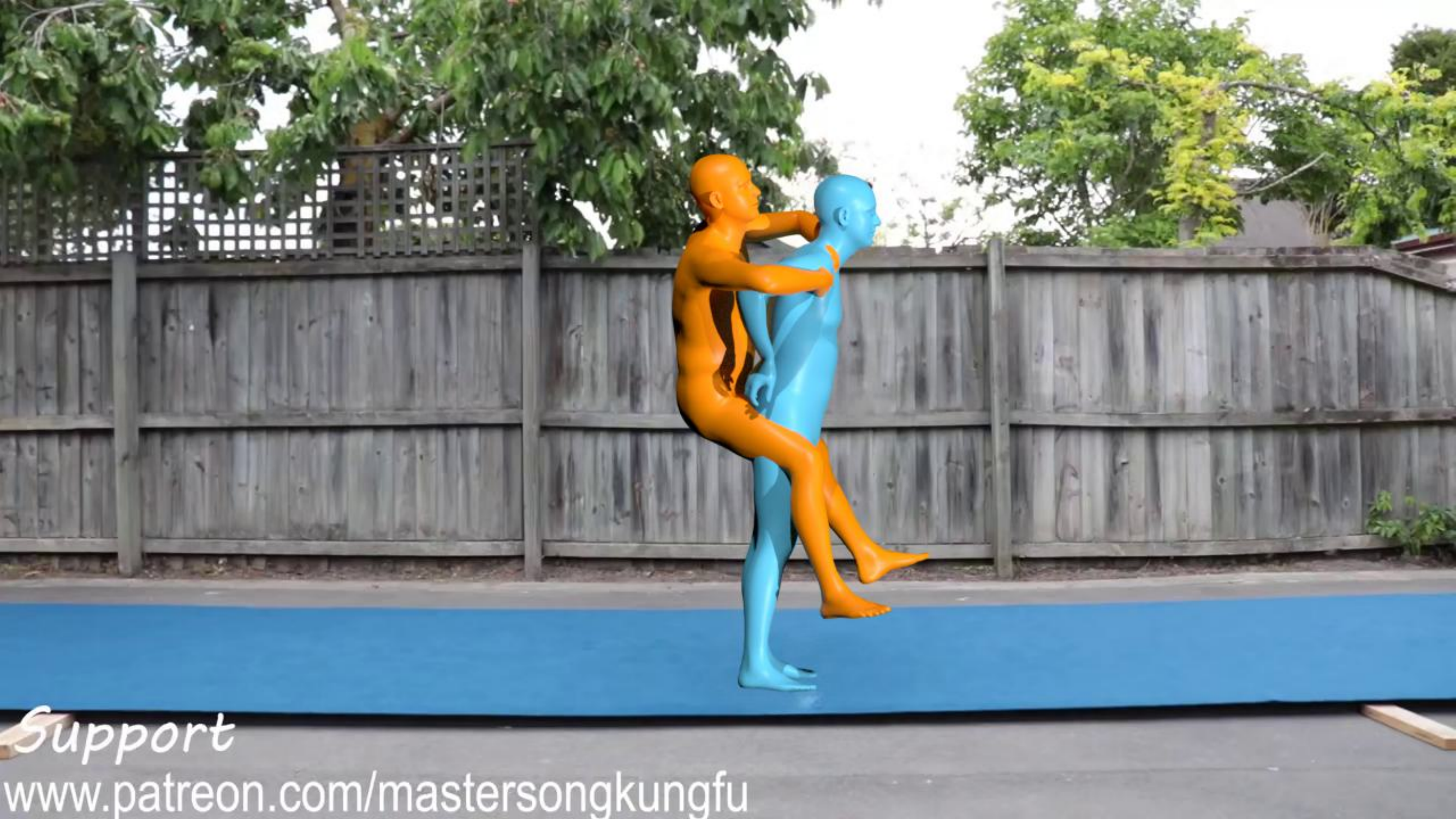}}} &
       \includegraphics[width=0.19\linewidth]{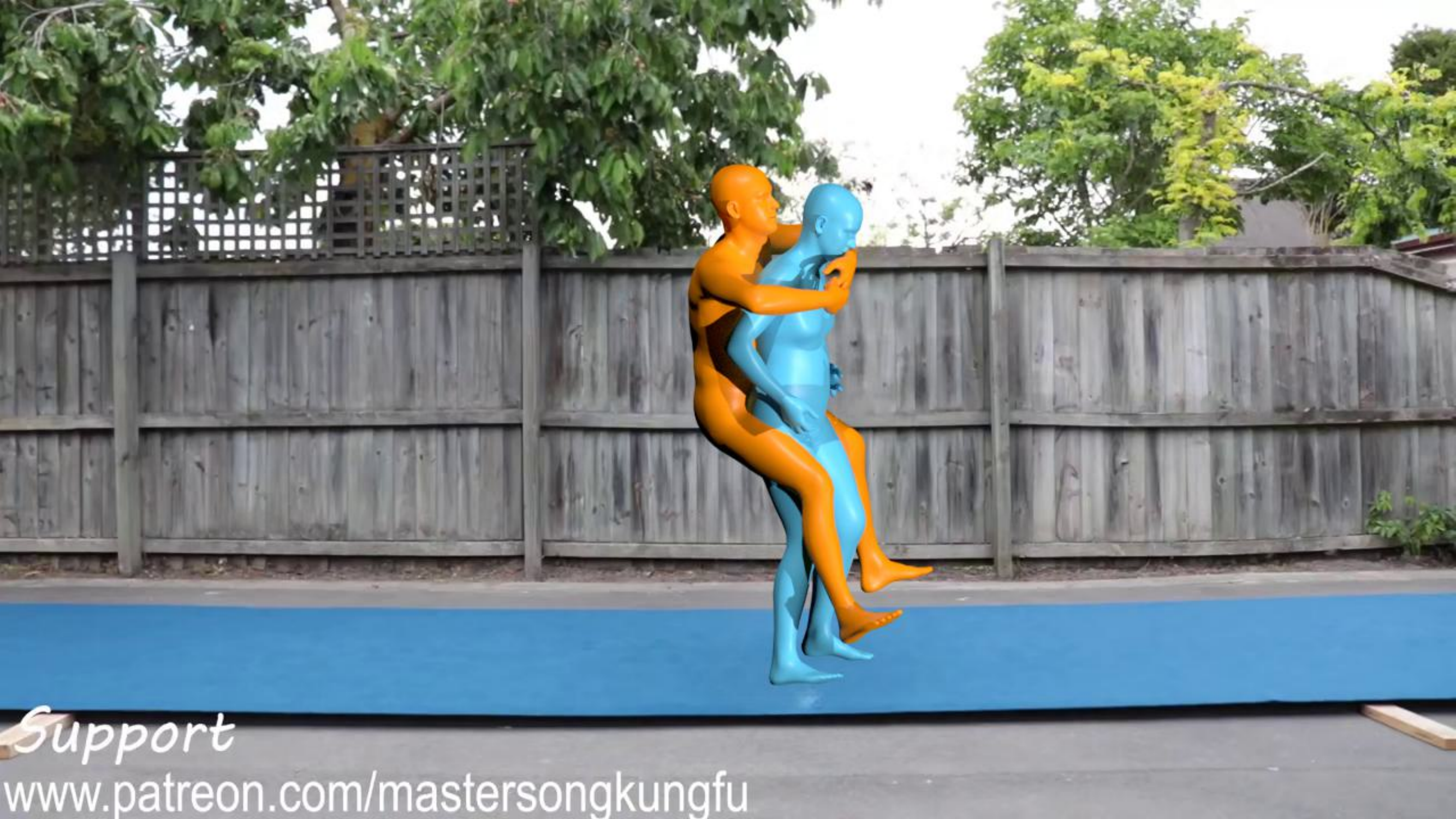} \\

       Single-person image  &  &  & \multirow{2}{*}{Generated interactive pose} \\
       + "Lift the other onto his back" \\

    \end{tabular}
    }
\captionof{figure}{\textbf{Ponimator} enables versatile interaction animation applications anchored on \textit{interactive poses}. For two-person images (top), Ponimator generates contextual dynamics from estimated interactive poses (\textcolor{green}{green box}). For single-person images (middle) with optional text prompts (bottom), Ponimator first generates partner interactive poses (\textcolor{magenta}{magenta box}) and then fulfill the interaction dynamics. }

\label{teaser}
\end{strip}

\customfootnotetext{*}{Work done at an internship at Snap Research NYC, Snap Inc.}
\customfootnotetext{$\dagger$}{Co-corresponding author}

\begin{abstract}

Close-proximity human-human interactive poses convey rich contextual information about interaction dynamics. Given such poses, humans can intuitively infer the context and anticipate possible past and future dynamics, drawing on strong priors of human behavior. Inspired by this observation, we propose Ponimator, a simple framework anchored on proximal interactive poses for versatile interaction animation. Our training data consists of close-contact two-person poses and their surrounding temporal context from motion-capture interaction datasets. Leveraging interactive pose priors, Ponimator employs two conditional diffusion models: (1) a pose animator that uses the temporal prior to generate dynamic motion sequences from interactive poses, and (2) a pose generator that applies the spatial prior to synthesize interactive poses from a single pose, text, or both when interactive poses are unavailable. Collectively, Ponimator supports diverse tasks, including image-based interaction animation, reaction animation, and text-to-interaction synthesis, facilitating the transfer of interaction knowledge from high-quality mocap data to open-world scenarios. Empirical experiments across diverse datasets and applications demonstrate the universality of the pose prior and the effectiveness and robustness of our framework. Codes and video visualization can be found at \href{https://stevenlsw.github.io/ponimator/}{https://stevenlsw.github.io/ponimator/}

\end{abstract}  
\section{Introduction}
\label{sec:intro}

The interplay between humans plays a crucial role in our daily lives. These interactions convey key social signals that reflect relationships and intentions. For example, a simple hug typically expresses closeness, a handshake serves as a formal greeting, while combat indicates opposing stances. A key observation is that interactive poses in close proximity (e.g., handshake) carry rich prior information about interaction dynamics. Specifically, a pair of such poses reveals contextual cues about spatial relationships, constraints, and intent, often suggesting probable ranges of past and future motions. These interactive poses can act as a bridge for modeling interaction dynamics with reduced complexity while inherently preserving prior knowledge of close interactions.

In this paper, we present \textit{Ponimator}, a novel framework that leverages the dynamics priors embedded in interactive poses through a generative model, demonstrating its versatility across various interaction animation tasks. We develop this interaction prior using a combination of two high-quality human-human interaction datasets: Inter-X~\cite{xu2024inter} and Dual-Human~\cite{fang2024capturing}. From these datasets, we construct a collection of two-person poses in close proximity, as shown in~\cref{fig:interactive-pose}, along with their preceding and subsequent interaction motions. Using this collection, we train a conditional diffusion model to generate contextual interaction dynamics given a pair of closely interactive poses.

We first demonstrate the application of our learned pose-to-dynamic interactive priors for open-domain images. Social interactions are frequently depicted in images, yet existing works~\cite{muller2024generative,fieraru2021remips,fieraru2020three,fang2024capturing} typically focus only on reconstructing static interactive poses, lacking the temporal dynamics of these interactions. Meanwhile, video diffusion models~\cite{guo2023animatediff,ho2022video,blattmann2023align} can animate images over time but often struggle to maintain motion and interaction integrity. In contrast, Ponimator seamlessly transfers learned interaction prior knowledge from high-quality 3D mocap datasets to these in-the-wild images through estimated interactive poses, as shown in \cref{teaser} (top). For broader applications, we developed an additional conditional diffusion model that leverages the spatial prior to generate interactive poses from multiple input types, including text descriptions, single poses, or both. Thus, when only a single person appears in an image, Ponimator can first generate a partner pose with an optional text prompt, and then animate the interactive poses over time (see \cref{teaser}). Furthermore, by anchoring on these interactive poses, Ponimator is able to generate short-clip two-person motions with proximal contact (see \cref{fig:exp-t2m-comparison}) directly from text input.

\begin{table}[t]
    \centering
    \footnotesize
    \setlength{\tabcolsep}{0.2em} %
    \resizebox{\linewidth}{!}{
    \begin{tabular}{cccc}

          \fcolorbox{red}{white}{\includegraphics[width=0.25\linewidth]{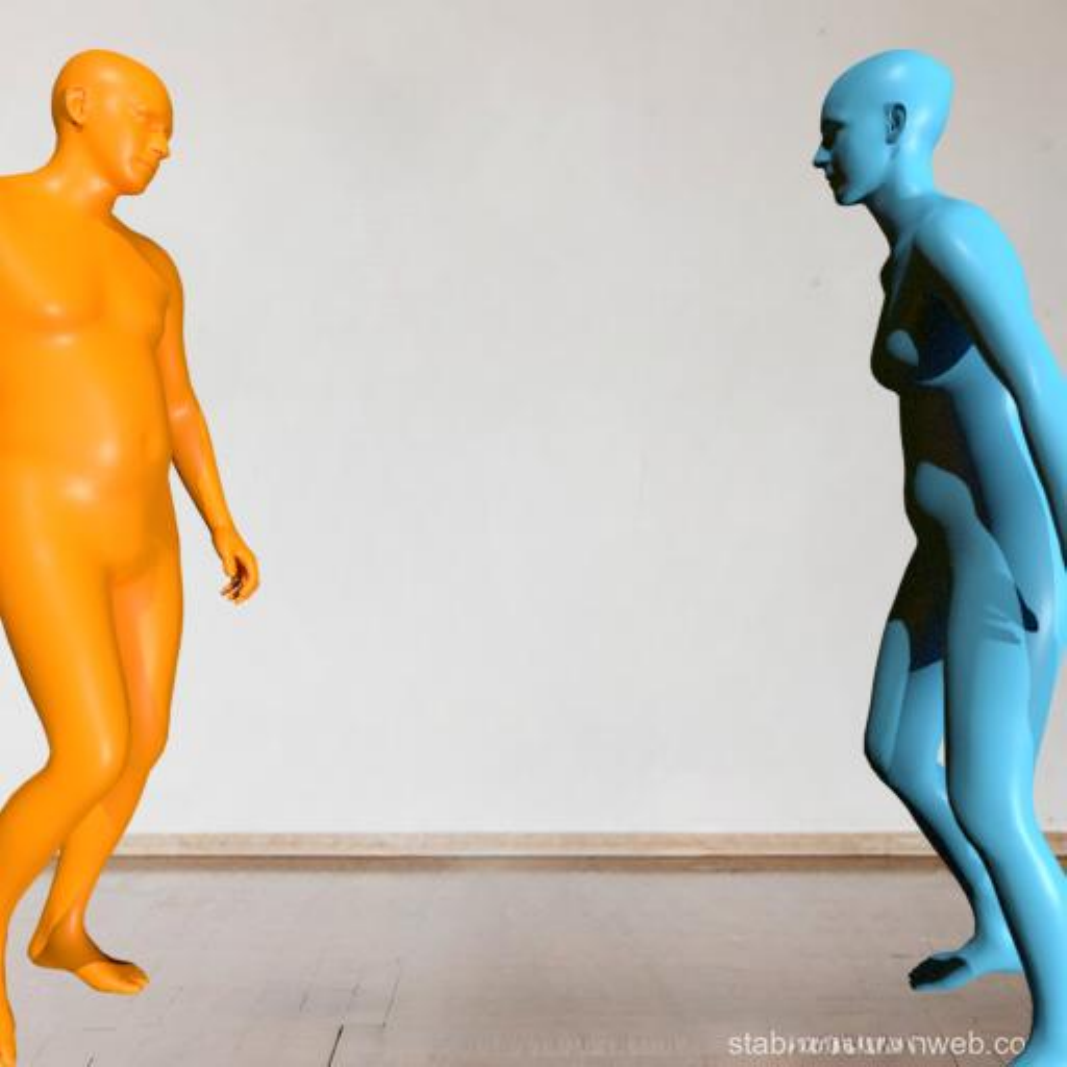}} &

          \fcolorbox{green}{white}{\includegraphics[width=0.25\linewidth]{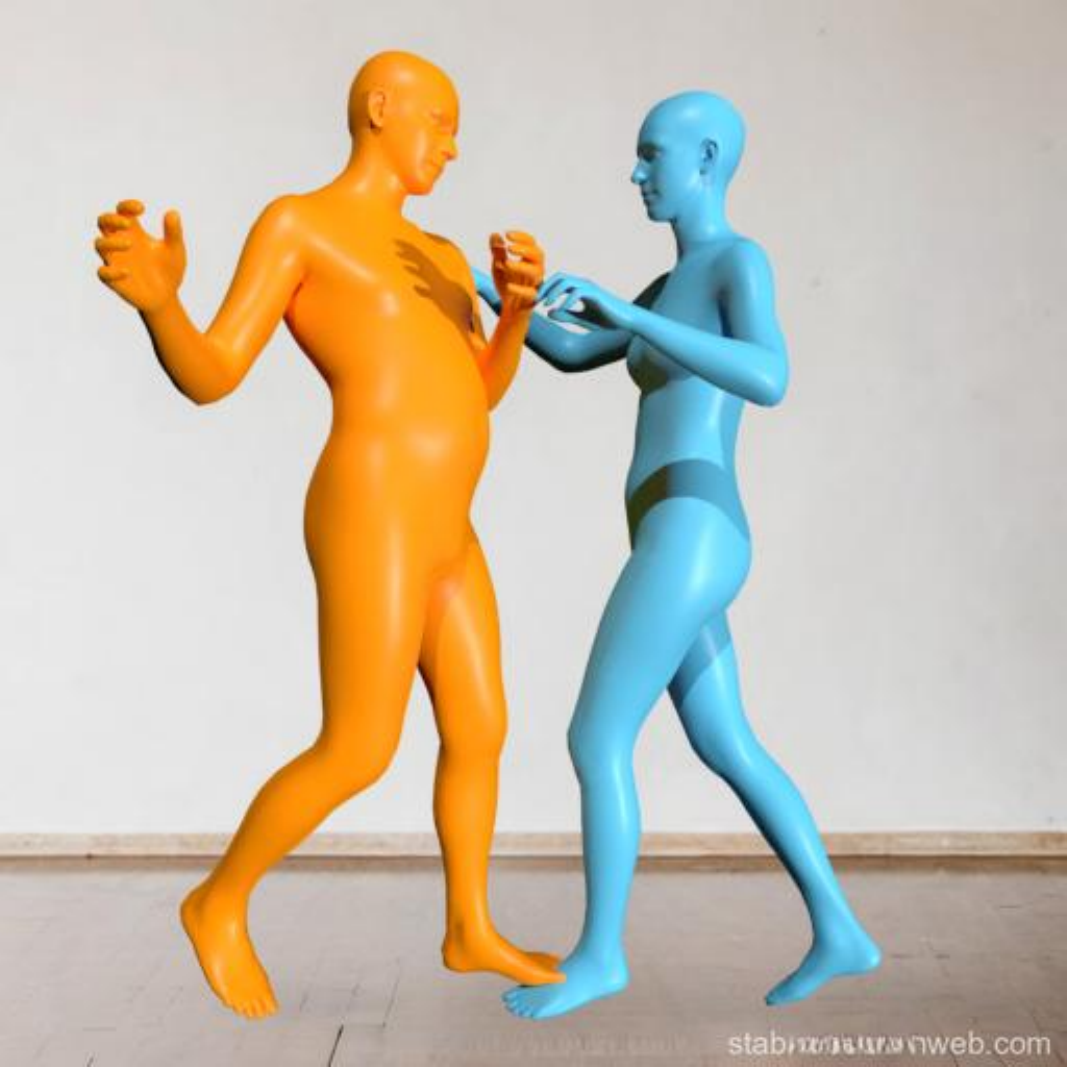}} &

           \fcolorbox{green}{white}{\includegraphics[width=0.25\linewidth]{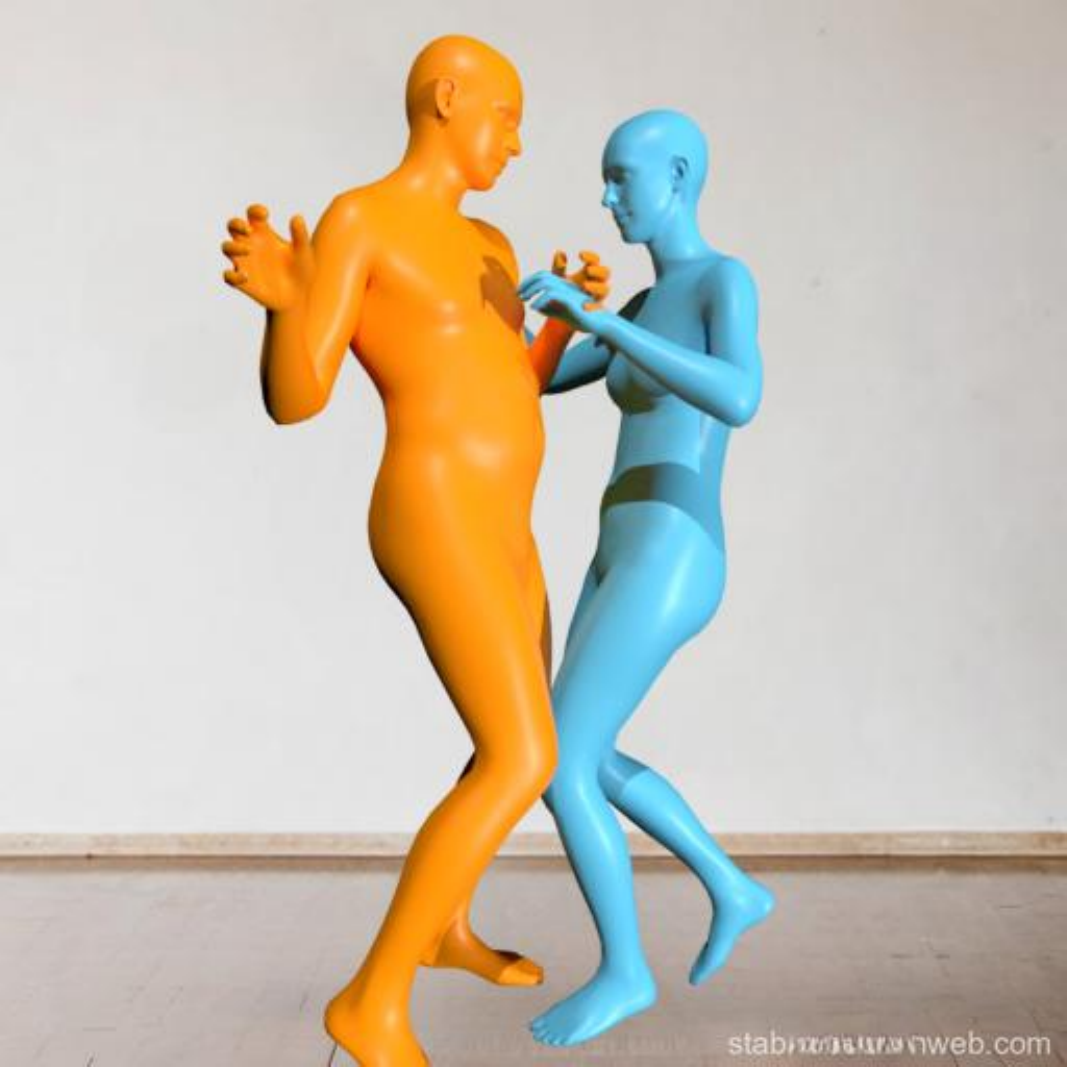}} &
           
         \fcolorbox{red}{white}{\includegraphics[width=0.25\linewidth]{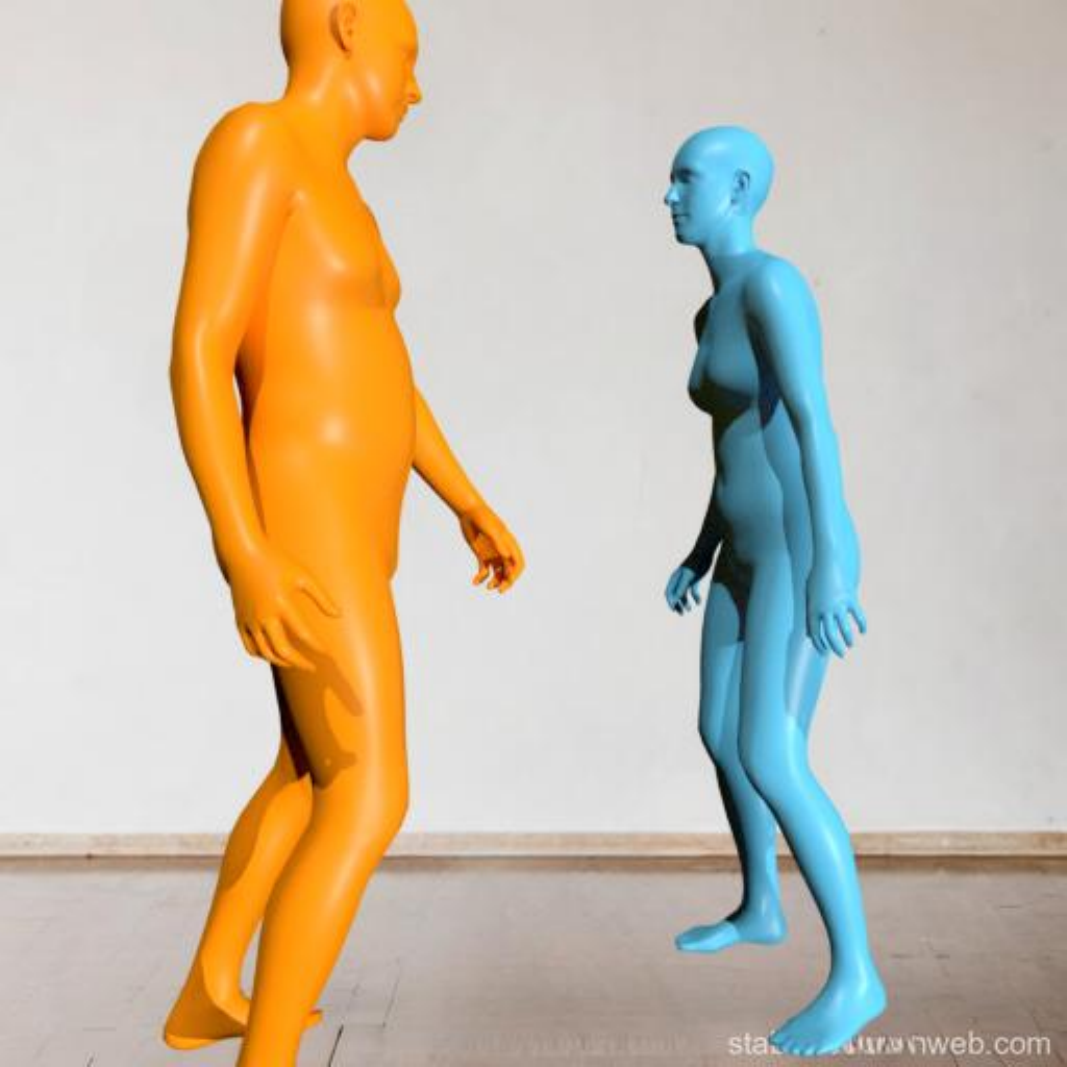}} \\

         \includegraphics[width=0.25\linewidth, trim={5cm 5cm 8cm 5cm},clip]{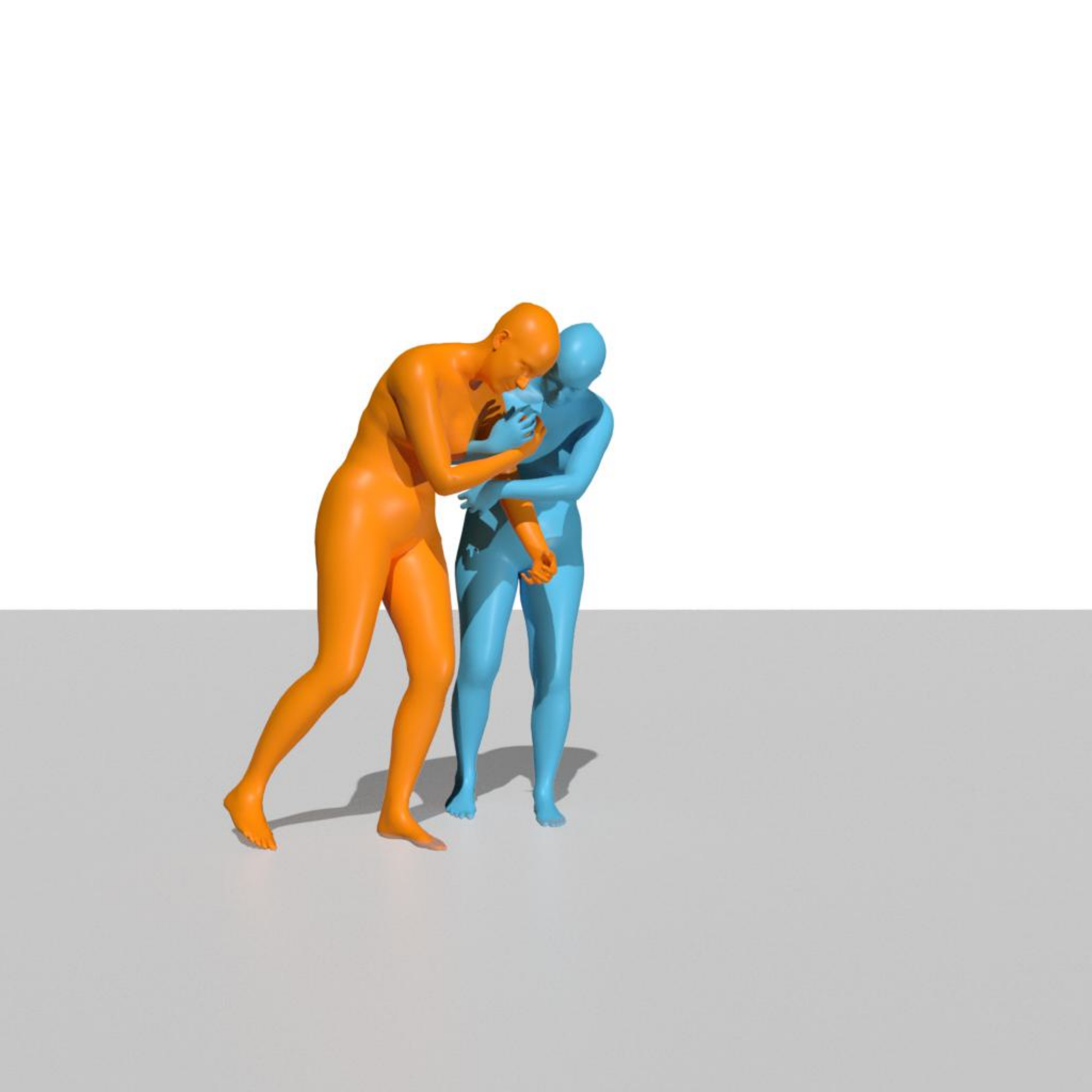} &

        \includegraphics[width=0.25\linewidth, trim={5cm 5cm 5cm 5cm},clip]{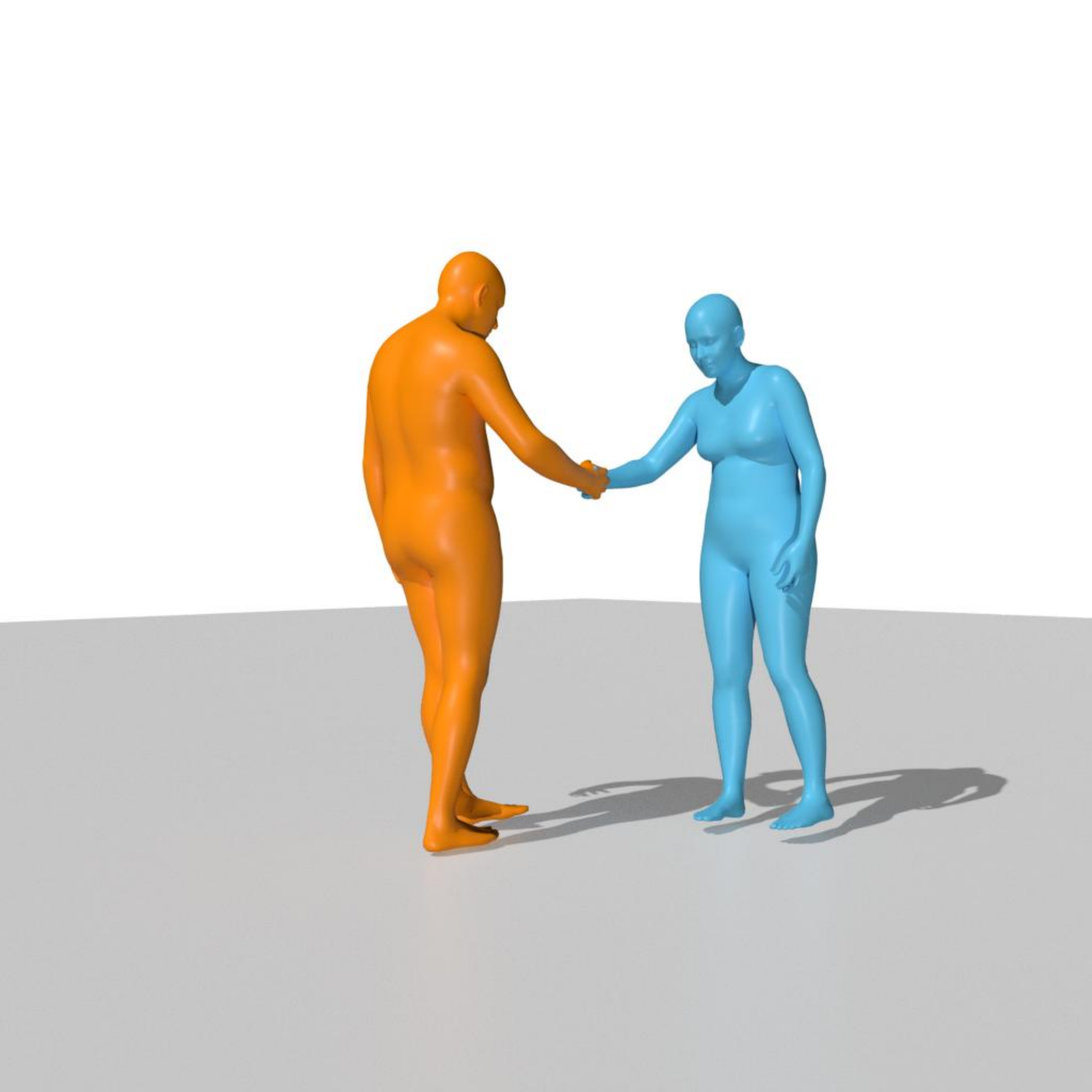} &

        \includegraphics[width=0.25\linewidth, trim={5cm 5cm 5cm 5cm},clip]{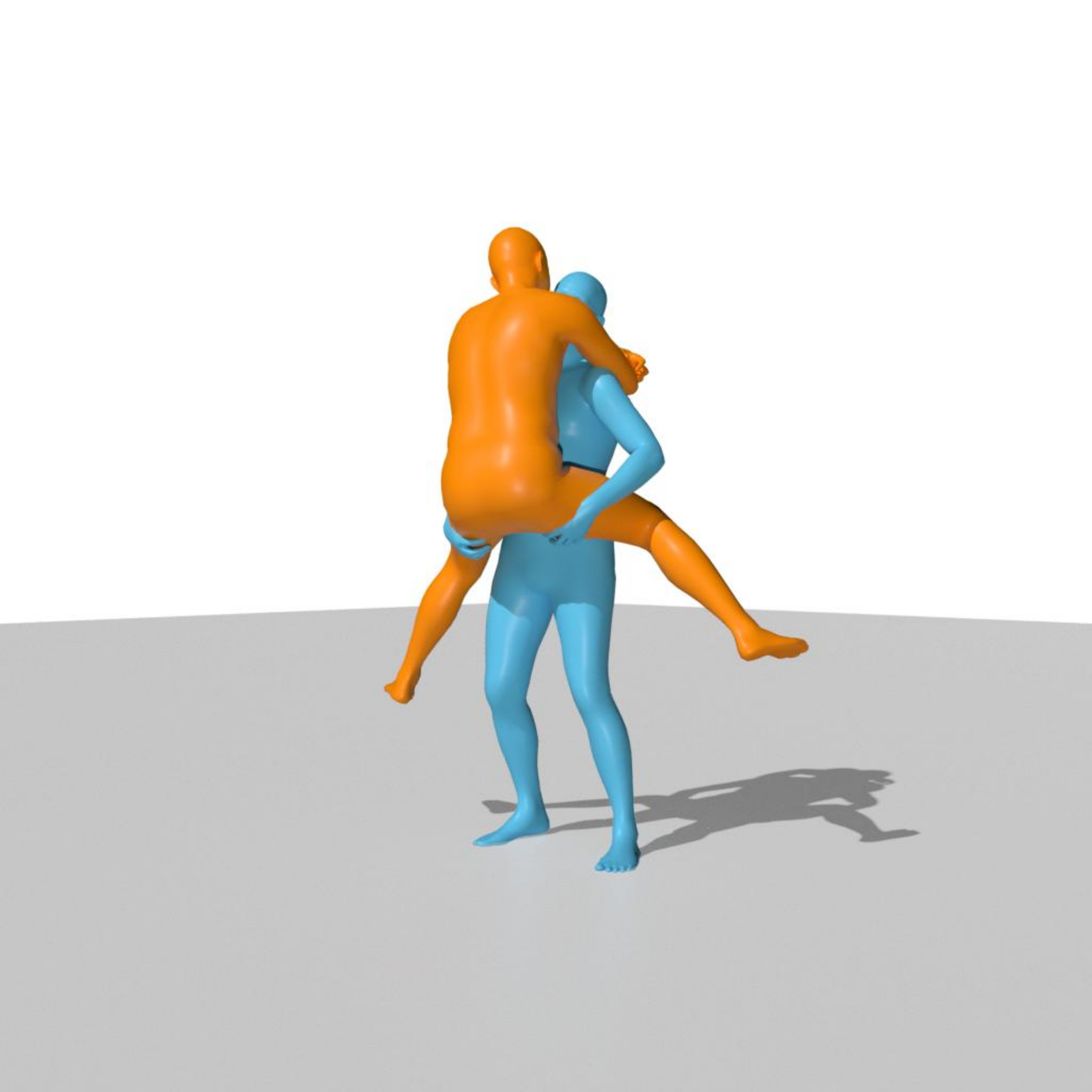} & 

        \includegraphics[width=0.25\linewidth]{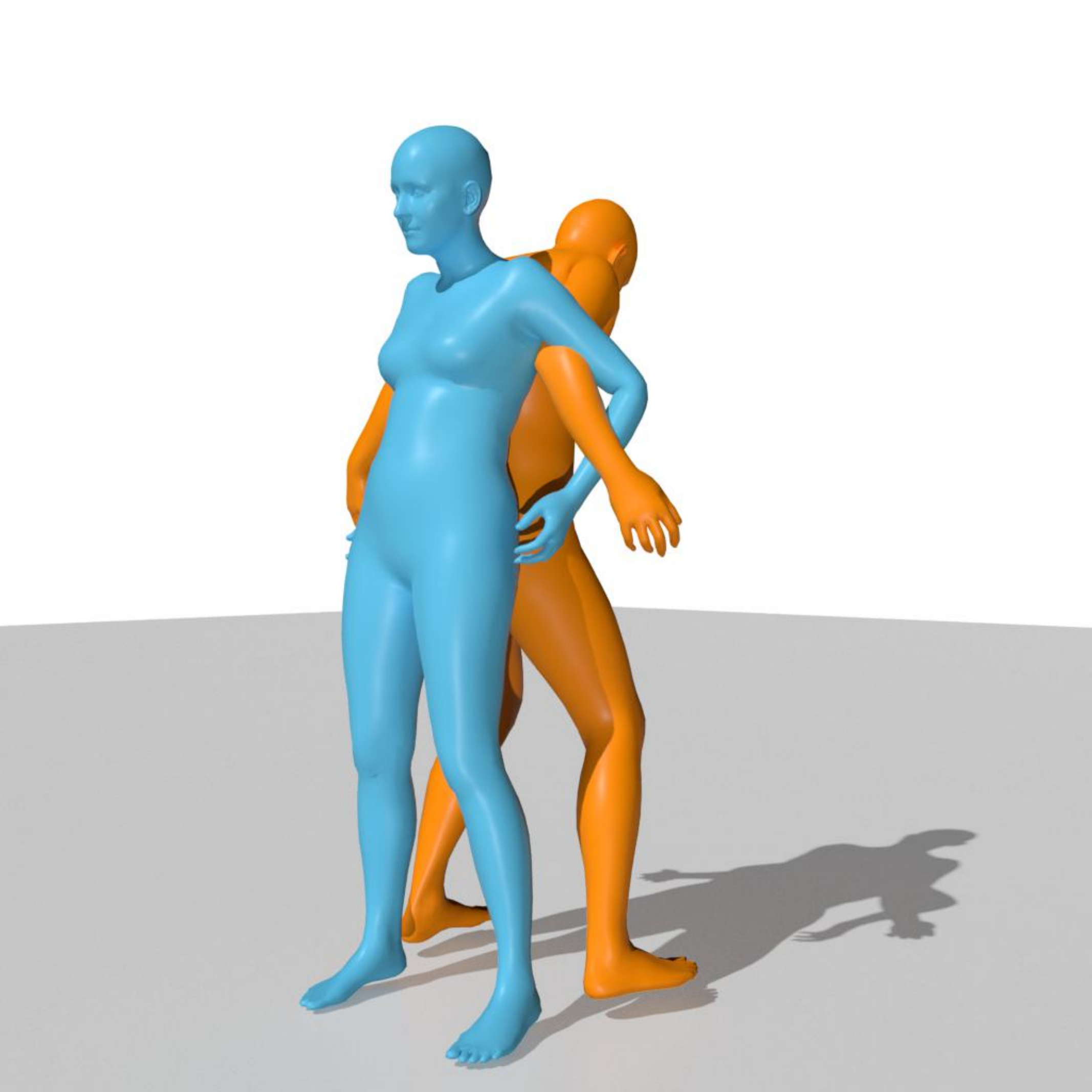} \\
         
    \end{tabular}
    }
\captionof{figure}{Interactive poses refer to two-person poses in proximity and close contact. The top row displays interactive (\textcolor{green}{green}) and non-interactive (\textcolor{red}{red}) poses within one sequence. Interactive poses allow observers to intuitively infer the temporal context, while non-interactive poses are more ambiguous and difficult to interpret. The bottom row showcases common daily interactive poses.}
\label{fig:interactive-pose}
\vspace{-2em}
\end{table}

Our key contributions are summarized as follows: 1) We present Ponimator, a simple framework designed to learn the dynamics prior of interactive poses from motion capture data, particularly focusing on proximal human-human interaction animations; 2) The learned prior is universal and generalizes effectively to poses extracted from open-world images, enabling animation of social interactions in human images; 3) Ponimator can generate interactive poses from a single-person pose, text, or both, combined with interactive pose animation, enabling diverse applications including reaction animation and text-to-interaction synthesis.
\section{Related work}
\label{sec:related}

\begin{figure*}[t]
\centering
\includegraphics[width=0.95\textwidth]{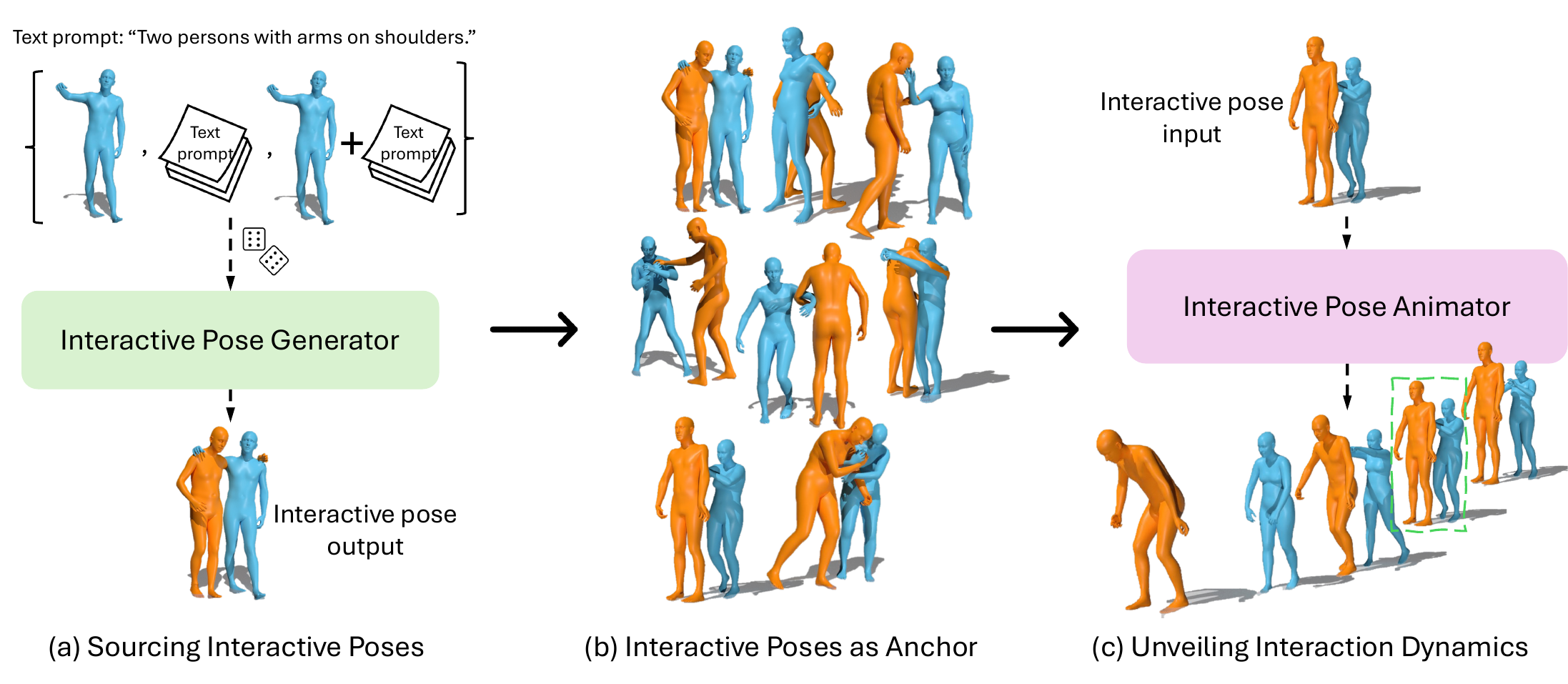}
\caption{\textbf{Framework overview.} Ponimator consists of a pose generator and animator, bridged by interactive poses. The generator takes a single pose, text, or both as input to produce interactive poses, while the animator unleashes interaction dynamics from static poses.
}
\label{fig:archi}
\vspace{-1.5em}
\end{figure*}

\noindent\textbf{Human-human Interactions in Images.}
Human–human interactions are prevailing in social images. Significant progress has been made in interactive pose estimation~\cite{fieraru2020three, fieraru2021remips, muller2024generative} and interaction sequence reconstruction~\cite{huang2024closely, ugrinovic2024multiphys}. Ugrinovic \etal integrate a physical simulator into the human mesh recovery pipeline to capture the physical significance of interactive poses. Huang \etal~\cite{huang2024closely} use Vector-Quantised representation learning and specialized losses to learn a discrete interaction prior, but suffer from limited interpretability and generalization. In contrast, our method directly anchors on interactive poses for interaction modeling without relying on additional physical simulators or intricate model designs. Our simple and interpretable prior generalizes well to in-the-wild settings, adhering to the principle that simplicity leads to robustness. The interactive pose prior is also explored in BUDDI~\cite{muller2024generative}, which estimates two-person static poses from images but is limited to static pose modeling and overlooks the rich dynamics of interactions. In contrast, our work unlocks interactive motions for both animation and generation in arbitrary open-world images.

\noindent\textbf{Human-human Motion Synthesis.} 
Generating human motion dynamics has been a long-standing task~\cite{kovar2023motion, arikan2002interactive, kovar2003flexible, ahuja2019language2pose, martinez2017human, lin2018generating}. Utilizing generative models have gained widespread popularity recently~\cite{petrovich2022temos, guo2020action2motion, petrovich2021action, guo2022generating, li2022ganimator, wang2021scene, wang2022towards, guo2022tm2t, zhang2023generating, jiang2023motiongpt, tanaka2023role, tevet2023human, zhang2022motiondiffuse, karunratanakul2023guided}. With the success of applying generative models in single-person motion synthesis and the release of large-scale two-person interaction datasets, such as InterGen~\cite{liang2024intergen} and Inter-X~\cite{xu2024inter}, there has been a surge in research~\cite{tevet2023human, tanaka2023role, javed2024intermask, ponce2024in2in, shan2024towards, fan2024freemotion, chopin2023interaction, liu2023interactive, xu2024regennet, ghosh2023remos, siyao2024duolando, shafir2024human, liu2023contactgen} focused on multi-person motion generation. However, most existing studies generate two-person motions following input text, but often overlooking close-contact dynamics. For example, Liang \etal~\cite{liang2024intergen} proposed a diffusion model for two-person motion generation, but it relies on detailed text input and struggles with realistic interaction. In contrast, our framework focuses on short-range interactions by leveraging generalizable interaction priors from static interactive poses, naturally ensuring physical contact between individuals and seamlessly generalizes to open-world scenarios.

\noindent\textbf{Human-human Motion Prediction.}
A body of work focuses on tracking multi-person motions from videos~\cite{insafutdinov2017arttrack, shuai2022large, iqbal2017posetrack}, forecasting future multi-person motions based on past movements~\cite{xu2023joint, wang2021multi, peng2022somoformer, xu2023stochastic, guo2022multi, starke2020local, starke2021neural} and generating reactive motion based on an individual’s full motion sequence~\cite{chopin2023interaction, liu2023interactive, xu2024regennet, ghosh2023remos, rahman2023best, fang2023pgformer, siyao2024duolando}. However, existing methods rely on long history context or full individual motions while treating interactive poses and human dynamics separately. In contrast, our approach bridges these two modalities by anchoring on interactive poses and leveraging their prior for dynamics forecasting. This integration enables our model to generate both past and future interaction dynamics while supporting flexible inputs with fewer constraints, such as text, single-pose, or both, unlocking diverse applications in animation and generation.

\vspace{-0.5em}
\section{Approach}
\label{sec:approach}
\vspace{-0.5em}

\begin{table*}
    \centering
    \footnotesize
    \setlength{\tabcolsep}{0.8em} %
    \resizebox{\linewidth}{!}{
        \begin{tabular}{c}

       \textbf{Two-person Image Interaction Animation} \\
       
        \includegraphics[width=\linewidth]{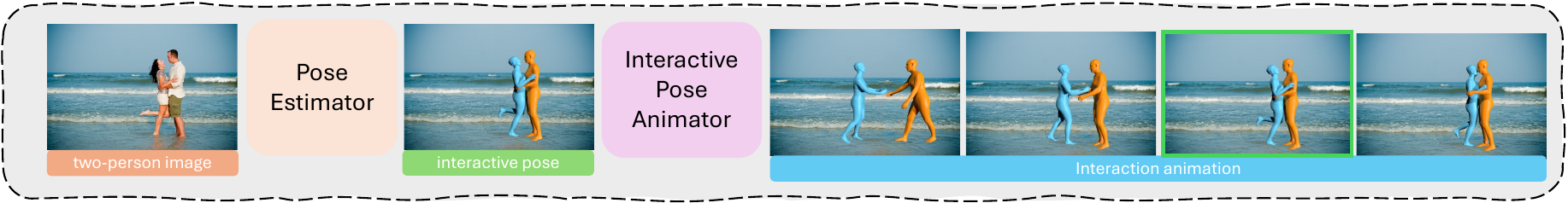} \\

      \textbf{Single-person Image Interaction Animation} \\

        \includegraphics[width=\linewidth]{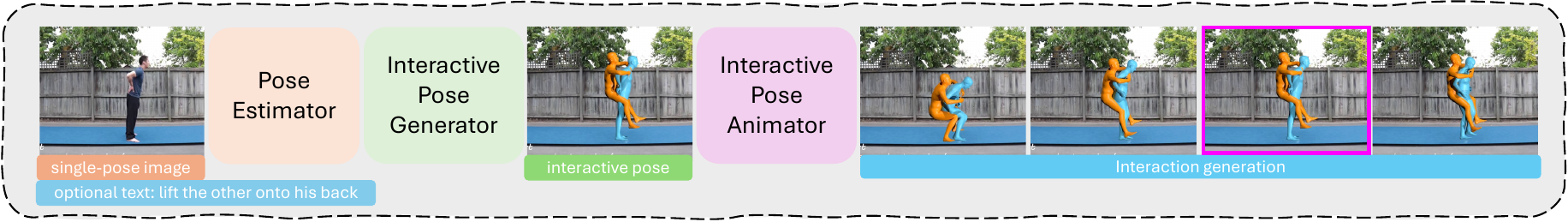} \\

         \textbf{Text-to-Interaction Synthesis} \\

   \includegraphics[width=0.9\linewidth]{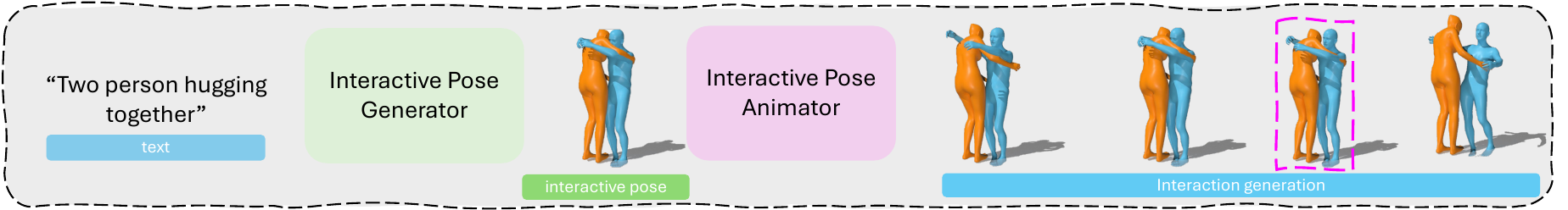} \\

    \end{tabular}
    }
\captionof{figure}{\textbf{Applications.} Our framework enables two-person image animation, single-person interaction generation, and text-to-interaction synthesis. For two-person images, we estimate interactive poses using an off-the-shelf model~\cite{muller2024generative}. For single-person images, we first estimate the pose by~\cite{cai2023smplerx} and generate its interactive counterpart. For text input, our unified pose generator could synthesize the pose directly. These poses are then fed into our animator to generate human dynamics.} 
\label{fig:application}
\vspace{-1.5em}
\end{table*}

Ponimator leverages interactive pose priors as intermediates for interaction animation, as shown in \cref{fig:archi}. We first introduce interactive poses and motion modeling (\cref{sec:prelim}). Then, we present the pose animator (\cref{sec:unvei}), which transforms interactive poses into motion, followed by the pose generator (\cref{sec:anchor}), which generates interactive poses from various inputs. Finally, in \cref{sec:app}, we explore Ponimator's applications to real-world images and text.

\subsection{Interactive Pose and Motion Modeling.}
\label{sec:prelim}

\noindent\textbf{Interactive pose and motion.} Our work defines interactive poses as the poses of two individuals in proximity and close contact. For person $a$, we use the SMPLX parametric body model~\cite{SMPL-X:2019} to model the pose $\bx^a = (\boldsymbol{\phi}^a, \boldsymbol{\theta}^a, \boldsymbol{\gamma}^a)$ and shape $\boldsymbol{\beta}^a \in \mathbb{R}^{10}$. 
Here, $\boldsymbol{\theta}^a \in \mathbb{R}^{21\times3}$ is the joint rotations, $\boldsymbol{\phi}^a \in \mathbb{R}^{1\times3}$ and $\boldsymbol{\gamma}^a \in \mathbb{R}^{1\times3}$ represents the global orientation and translation. The interactive pose of two individuals $a$ and $b$ is given as $\bx_I = (\bx_I^a, \bx_I^b)$. An interaction motion consists of a short pose sequence $\boldsymbol{\mathcal{X}}$ of length $N$, centered around an interaction moment, along with shape parameters $\boldsymbol{\theta}$ of both individuals, where $\boldsymbol{\mathcal{X}} = \{\mathbf{x}_i\}_{i=1}^{N}$, $\boldsymbol{\beta}=(\boldsymbol{\beta}^a, \boldsymbol{\beta}^b)$. $\boldsymbol{\mathcal{X}}$ includes an pair of interactive poses $\bx_I$ at interaction moment index $I$ within the sequence, and its nearby past poses $\bx_{1:I}$ and future poses $\bx_{I+1:N}$. An example of interactive pose and motion is shown in \cref{fig:interactive-pose}.

\noindent\textbf{Interaction motion modeling.} The interactive pose $\bx_I$ encodes rich \textit{temporal} and \textit{spatial} priors. 
As shown \cref{fig:interactive-pose}, interactive poses convey motion dynamics (top row) and spatial relationships (bottom row) between individuals. 
The strong prior make it easier for models to learn, whereas non-interactive poses lack clear interaction cues, making learning more challenging. Therefore, we model the interaction motion $(\boldsymbol{\mathcal{X}}, \boldsymbol{\beta})$ by anchoring on its interactive pose $\bx_I$.

\begin{equation}
\label{eq:decompose}
p(\boldsymbol{\mathcal{X}}, \boldsymbol{\beta}) = \tikzmarknode{x}{\highlight{red}{p(\boldsymbol{\mathcal{X}}; \highlight{green}{{\mathbf{x}_I}}, \boldsymbol{\beta})}} \cdot \tikzmarknode{s}{\highlight{blue}{p(\highlight{green}{\mathbf{x}_I}, \boldsymbol{\beta})}}
\end{equation}
\begin{tikzpicture}[overlay,remember picture,>=stealth,nodes={align=left,inner ysep=1pt},<-]
   \path (x.south) ++ (0,-0.5em) node[anchor=north east,color=red!67] (mean){temporal prior};
    \draw [color=red!57](x.south) |- ([xshift=-0.3ex,color=red]mean.south west);
    \path (s.south) ++ (0,-0.5em) node[anchor=north west,color=blue!67] (mean){spatial prior};
    \draw [color=blue!57](s.south) |- ([xshift=-0.3ex,color=blue]mean.south east);
\end{tikzpicture}

\noindent\textbf{Learning prior from diffusion model.}
Each prior's distribution in \cref{eq:decompose} is captured by a generative diffusion model~\cite{ho2020denoising} $G$, trained on high-quality mocap data. To model the underlying distribution of data $\mathbf{z}_0$, the diffusion model introduces noise $\boldsymbol{\epsilon}$ to the clean data $\mathbf{z}_0$ in the forward pass, following $\mathbf{z}_t = \sqrt{\bar{\alpha_t}}\mathbf{z}_0 + \sqrt{1-\Bar{\alpha}_t }\boldsymbol{\epsilon},\  \boldsymbol{\epsilon} \sim \mathcal{N}(0, \mathbf{I})$, where $\alpha_t \in (0, 1)$ are constants, $t$ is the diffusion timestep $t \in [0, T_\text{diffusion}]$. The model $G$ aims to recover clean input by $\hat{\mathbf{z}}_0 = G(\mathbf{z}_t, t, \bc)$ from the noisy observations $\mathbf{z}_t$ and condition $\bc$, optimizing the objective:
\begin{equation}
\label{eq:diffusion-loss}
    \mathcal{L}_D = \mathbf{E}_{\mathbf{z}_0, \bc, \boldsymbol{\epsilon} \sim \mathcal{N}(0,\mathbf{I}),t}[\| \mathbf{z}_0 - G(\mathbf{z}_t, t, \bc) \|_2^b ]
\end{equation}
During inference, the model iteratively predicts $G(\mathbf{z}_t, t, \bc)$ from $t=T_\text{diffusion}$ to $t=0$, gradually denoising the sample until it recovers the original clean data $\hat{\mathbf{z}}_0$.

\noindent\textbf{Close-proximity training data.} 
We collect large-scale training data from public mocap datasets, InterX~\cite{xu2024inter} and DualHuman~\cite{fang2024capturing}, without requiring contact annotations. Interactive poses are detected by spatial proximity, and if within a threshold, we extract the pose with its past and future frames to form a 3-second interaction clip. %

\begin{table*}
    \centering
    \footnotesize
    \setlength{\tabcolsep}{0.2em} %
    \resizebox{\linewidth}{!}{
        \begin{tabular}{c|cccc}

        {Input} & 
        \multicolumn{4}{c}{{Interaction Animation (left$\rightarrow$right: time steps)}} \\

        \visvideo{vis_buddi}{Karate_142}{05}{10}{15}{20}{0.19}[4]

        \visvideo{vis_buddi}{highfive_676142}{10}{15}{20}{25}{0.19}[4]

        \visvideo{vis_buddi}{happy_258387}{05}{10}{15}{20}{0.13}[4]

        \visvideo{vis_buddi}{kungfu}{05}{10}{15}{20}{0.19}[4]
        
        \end{tabular}
    }

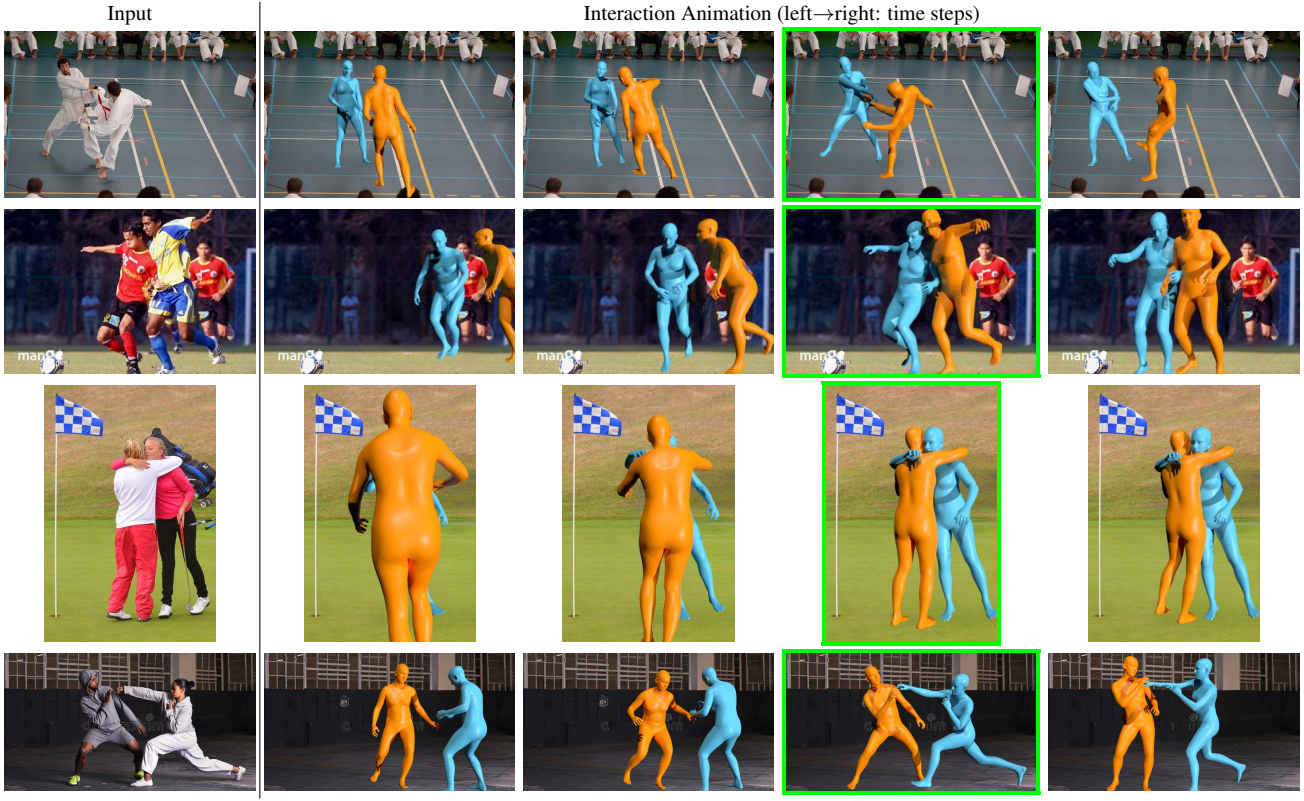
\captionof{figure}{\textbf{Interactive pose image animation} on FlickrCI3D dataset~\cite{fieraru2020three}. Left shows the input image, right shows the animated interaction motions. Interactive-pose frame is labeled in \textcolor{green}{green box}.}
\label{fig:exp-buddi-results}
\vspace{-1.5em}
\end{table*}

\subsection{Unveiling Dynamics from Interactive Poses}
\label{sec:unvei}
The interactive pose animator captures the temporal prior in $p(\boldsymbol{\mathcal{X}}; \mathbf{x}_I, \boldsymbol{\beta})$ given an interactive pose $\mathbf{x}_I$ and two person's shape  $\boldsymbol{\beta}$. The objective is to generate the motion sequences $\hat{\boldsymbol{\mathcal{X}}} = \{\hat{\mathbf{x}}_i\}_{i=1}^N$ where $\hat{\mathbf{x}}_I\approx \mathbf{x}_I$, as shown in \cref{fig:archi} (c).

\noindent\textbf{Interactive pose-centered representation.} We anchor the entire sequence on the interactive pose $\mathbf{x}_I$ and define the denoising target $\mathbf{z}_0$ as the motion residuals with respect to interactive poses $\mathbf{z}_0 = \{\mathbf{x}_i -\mathbf{x}_I\}_{i=1}^N$
This learning objective enforces model to learn the contextual dynamics strongly shaped by interactive poses.
During inference, we recover the predicted pose sequence $\{\hat{\mathbf{x}}_i\}_{i=1}^N$ by $\hat{\mathbf{z}}_0 + \mathbf{x}_I$.

We encode the interactive time index $I$ with a one-hot vector $\mathbf{m}_I \sim \texttt{OneHot}(I)\in \{0, 1\}^N$, where $\mathbf{m}_I^i = 1 \ \texttt{iff} \ i=I$. To better preserve the spatial structure of interactive pose at time $I$ in pose sequences, we apply an imputation strategy to the diffusion model, where the noise input $\mathbf{z}_{t}$ in~\cref{eq:diffusion-loss} is substituted with $\tilde{\mathbf{z}}_{t}$:
\begin{equation}
\label{eq:animator}
\tilde{\mathbf{z}}_{t} = (1- \mathbf{m}_I) \odot \mathbf{z}_{t} + \mathbf{m}_I \odot \mathbf{0}, \quad \bc= (\mathbf{m}_I, \mathbf{x}_I, \boldsymbol{\beta}),
\end{equation}
where $\odot$ denotes element-wise multiplication and  $\bc$ is the input condition. After imputation, noise is added to interactive poses (i.e., $\tilde{\mathbf{z}}_t + \mathbf{x}_I$) before fed into the network.

\noindent\textbf{Condition encoding.} The interaction time condition $\mathbf{m}_I$ is concatenated with the initial model input along the feature dimension. We encode the remaining conditions $(\mathbf{x}_I, \boldsymbol{\beta})$ by leveraging the SMPLX joint forward kinematics (FK) function $\texttt{FK}(\cdot, \cdot)$ to compute joint positions of interactive pose $\mathbf{j}_I=(\texttt{FK}(\mathbf{x}_I^a, \boldsymbol{\beta}_a), \texttt{FK}(\mathbf{x}_I^b, \boldsymbol{\beta}_b))$. Here, $\mathbf{j}_I$ inherently encodes both individuals' poses and shapes. It is further embedded through a single-layer MLP and injected into the model layers via AdaIN~\cite{huang2017arbitrary}.

\noindent\textbf{Architecture and training.} We adopt the DiT~\cite{peebles2023scalable} architecture as our diffusion model, built on stacked Transformer blocks~\cite{vaswani2017attention} that alternate spatial attention for human contact and temporal attention for motion dynamics. To train the model, besides diffusion loss $\mathcal{L}_D$ in \cref{eq:diffusion-loss}, we apply the SMPL loss $\mathcal{L}_{\text{smpl}}$ as the MSE between the denoised pose sequence and the clean input. We also use an interaction loss $\mathcal{L}_{\text{inter}}$~\cite{liang2024intergen} and a velocity loss ~\cite{tevet2023human}. $\mathcal{L}_{\text{vel}}$ encourages contact between individuals in close proximity, while $\mathcal{L}_{\text{vel}}$ ensures motion coherence. The total loss $ \mathcal{L} = \lambda_{D}\mathcal{L}_D + \lambda_{\text{smpl}}\mathcal{L}_{\text{smpl}} + \lambda_{\text{inter}}\mathcal{L}_{\text{inter}} + \lambda_{\text{vel}}\mathcal{L}_{\text{vel}}$. To improve robustness and generalization to noisy real-world poses, we apply augmentation by adding random noise to interactive pose $\mathbf{x}_I$. Please refer to \cref{sec:implement} for details.

\subsection{Interactive Pose Generator}
\label{sec:anchor}

\begin{table*}
    \centering
    \footnotesize
    \setlength{\tabcolsep}{0.2em} %
    \resizebox{\linewidth}{!}{
        \begin{tabular}{c|cccc}

        {Input} & 
        \multicolumn{3}{c}{{

        Interaction Generation (left$\rightarrow$right: time steps)}} \\

         \visvideo{vis_motionx}{Boxing_kicking_jumping_6_clip2}{10}{15}{20}{25}{0.19}[3][magenta]
        
         \visvideo{vis_motionx}{Ways_to_Stand_Karate_clip1}{05}{10}{15}{20}{0.19}[4][magenta]

          \visvideo{vis_motionx}{Ways_to_Stand_Groomsman_clip1}{05}{10}{15}{20}{0.19}[4][magenta]
          
         {"two person pose for a photo"} \\
        
        \visvideo{vis_motionx}{Ways_to_Pick_Up_a_Dollar_Sentimental_clip1}{05}{15}{20}{25}{0.19}[3][magenta]
         
         {"one person lift another one up"}  \\

  \\
    \end{tabular}
    }

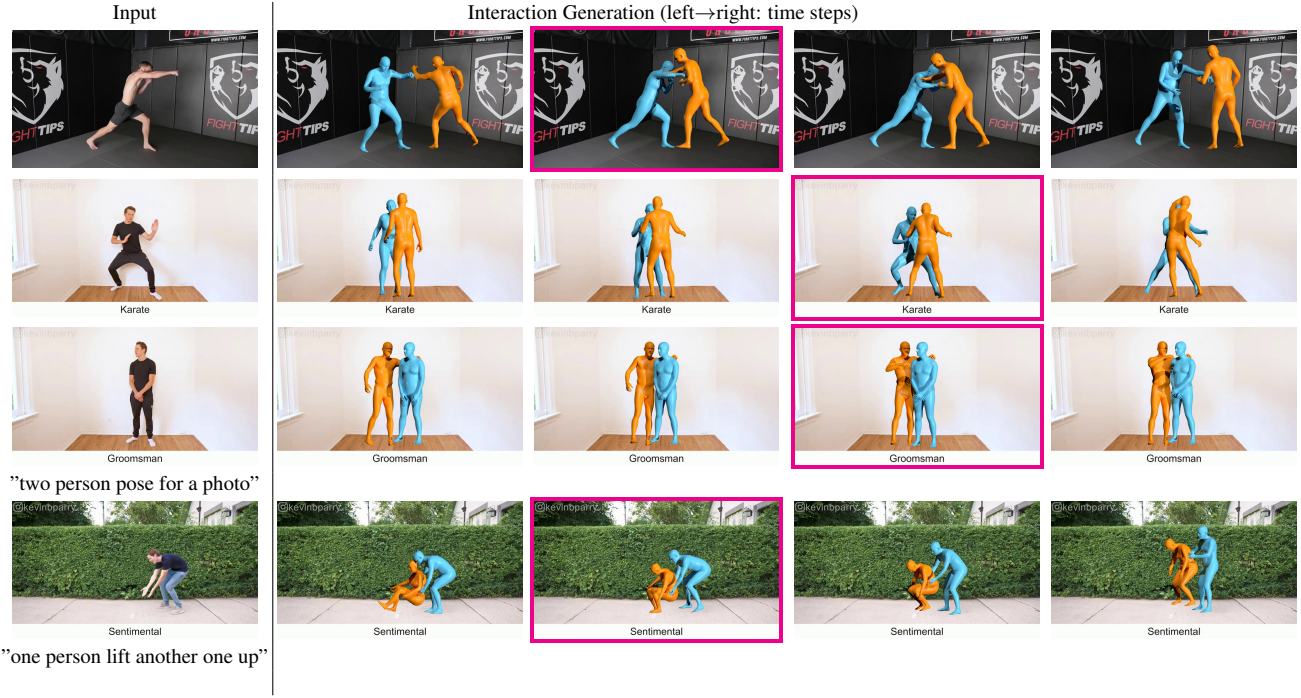
\captionof{figure}{\textbf{Single-person image interaction generation} on Motion-X~\cite{lin2024motion} dataset. Left shows the single person image input, right shows the generated two-person interaction dynamics. The generated interactive pose frame is labeled in \textcolor{magenta}{magenta box}. Top two rows display single-person pose inputs, while the bottom two show the same with accompanying text below the input image.} 

\label{fig:exp-motionx-results}
\vspace{-1.5em}
\end{table*}

The interactive pose generator models $p(\mathbf{x}_I, \boldsymbol{\beta})$ in \cref{eq:decompose}, leveraging the spatial prior to generate $\mathbf{x}_I, \boldsymbol{\beta} $ from various conditions, as shown in \cref{fig:archi}(a).

\noindent\textbf{Unified input conditioning.} Given various input conditions, including text $\bc$, single person pose $(\mathbf{x}_I^a, \boldsymbol{\beta}^a)$, or both, the model generates $\mathbf{z}_0^a = (\mathbf{x}_I^a, \boldsymbol{\beta}^a)$ and $\mathbf{z}_0^b = (\mathbf{x}_I^b, \boldsymbol{\beta}^b)$, which together form the diffusion target $\mathbf{z}_0 = (\mathbf{z}_0^a, \mathbf{z}_0^b)$ in \cref{eq:diffusion-loss}. To integrate these conditions into a unified model, we introduce two masks, $\mathbf{m}_c$ and $\mathbf{m}_a$, to encode the presence of text and pose conditions, respectively. These masks are sampled independently from a Bernoulli distribution with probability $p_{\text{condition}}$ during training. We modify the model input $\mathbf{z}_t$ and text condition $\bc$ to $\tilde{\bc}$ in \cref{eq:diffusion-loss} as:
\begin{equation}
\label{eq:generator}
\tilde{\mathbf{z}}_{t} = ((1-\mathbf{m}_a) \odot \mathbf{z}_{t}^a + \mathbf{m}_a \odot \mathbf{z}_0^a, \mathbf{z}_{t}^b), \quad
\tilde{\bc} = \mathbf{m}_c \odot \bc.
\vspace{-1em}
\end{equation}
This design enables the model to accommodate multiple combinations of conditions.

In SMPL, human shapes are coupled with genders $g\in {\{\text{male}, \text{female}, \text{neutral}\}}$. To enable a more generic shape condition, we instead use the global joint positions of rest pose $\bj_{\text{rest}}^{\{a, b\}}$, which inherently capture both shape and gender information, and define the diffusion target as $\mathbf{z}_0=(\mathbf{x}_I^{\{a,b\}}, \bj_{\text{rest}}^{\{a,b\}})$. After generation, we can recover $\boldsymbol{\beta}^{\{a,b\}}$ from $\bj_{\text{rest}}^{\{a,b\}}$ using inverse kinematics (IK).

\noindent\textbf{Architecture and training.} 
We use the same architecture as pose animator with modifications below.
(1) The text condition $\bc$ is encoded via CLIP~\cite{radford2021learning}, processed by two trainable Transformer layers, and injected by AdaLN~\cite{huang2017arbitrary}. (2) We retain spatial attention layers and remove temporal attentions. The model is trained with standard diffusion loss $\mathcal{L}_D$ in \cref{eq:diffusion-loss}, SMPL loss  $\mathcal{L}_{\text{smpl}}$, and bone length loss $\mathcal{L}_{\text{bone}}$ minimizes the MSE with ground-truth lengths in the SMPLX~\cite{SMPL-X:2019} kinematic tree. Total loss $\mathcal{L} = \lambda_D\mathcal{L}_D + \lambda_{\text{smpl}}\mathcal{L}_{\text{smpl}} + \lambda_{\text{bone}}\mathcal{L}_{\text{bone}}$. Please see \cref{sec:implement} for details.

\subsection{Applications}
\label{sec:app}
Our framework supports two-person interactive pose image animation, single-person pose interaction generation, and text-to-interaction synthesis, as shown in \cref{fig:application}. 

\noindent\textbf{Interactive pose image animation.}
As shown in 1st row of \cref{fig:application}, given a two-person image, we estimate the interactive pose $\hat{\mathbf{x}}_I$ using an off-the-shelf model~\cite{muller2024generative}. The estimated pose is fed into our interactive pose animator (\cref{sec:unvei}) to generate motions guided by the temporal prior in interactive poses. Our model provides flexible interaction timing control by adjusting $I$ in \cref{eq:animator}, where $I=0$ predicts future motion, $I=N$ reconstructs the past, and generally, $n=\frac{N}{2}$ enables symmetric animation. Open-world animation results are shown in \cref{fig:exp-buddi-results}.

\noindent\textbf{Single-person pose interaction generation.} As shown in the 2nd row of \cref{fig:application}, given a single-person image, we estimate the pose $\hat{\mathbf{x}}_I^a$ using off-the-shelf model such as ~\cite{cai2023smplerx} and feed it into our interactive pose generator (\cref{sec:anchor}). We set $\mathbf{m}_a = \mathbf{0}, \mathbf{m}_c = 0$ in \cref{eq:generator} as model input, disabling text input and allowing $\hat{\mathbf{x}}_I^a$ to generate its interactive counterpart $\mathbf{x}_I^b$ using the spatial prior in interactive poses. Alternatively, setting $\mathbf{m}_c = 1$ enables additional text conditioning. Once the interactive pose $\hat{\mathbf{x}}_I = (\hat{\mathbf{x}}_I^a, \hat{\mathbf{x}}_I^b)$ is obtained, it is fed into the interactive pose animator (\cref{sec:unvei}) to synthesize motion dynamics. Open-world results are presented in \cref{fig:exp-motionx-results}.

\noindent\textbf{Text-to-interaction synthesis.} As shown in 3rd row of \cref{fig:application}, given a short phrase, we generate the interactive pose $\hat{\mathbf{x}}_I$ by setting $\mathbf{m}_a = \mathbf{0}, \mathbf{m}_c = 1$ in \cref{eq:animator}. The generated $\hat{\mathbf{x}}_I$ is then passed to the pose animator to produce the corresponding motion. Examples for "two-person hugging together" and "push" are presented in \cref{fig:application,fig:exp-t2m-comparison}.

\section{Experiments}
\label{exp}

\begin{table}
    \centering
    \footnotesize
    \setlength{\tabcolsep}{0.2em} %
    \resizebox{1.05\linewidth}{!}{
        \begin{tabular}{c|cccc}

    {Interactive Pose} & 
        \multicolumn{4}{c}{{Interaction Animation (left$\rightarrow$right: time steps)}} \\

   \viscontact{vis_contact/vis_interx}{G020T006A030R003}{10}{15}{20}{25}[5cm 5cm 5cm 5cm]

     \multicolumn{5}{c}{{Inter-X test set~\cite{xu2024inter}}} \\

     \viscontact{vis_contact/vis_dh}{pair45_001740_1}{10}{15}{20}{25}[5cm 5cm 5cm 5cm]
     
    \multicolumn{5}{c}{{Dual-Human test set~\cite{fang2024capturing}}} \\

    \viscontact{vis_contact/vis_dd100}{Waltz_010_000_split0_00}{10}{15}{20}{25}[5cm 5cm 5cm 5cm]

    \multicolumn{5}{c}{{Duolando dataset~\cite{siyao2024duolando}}} \\

     \viscontact{vis_contact/vis_hi4d}{15_fight15_16_0000880}{10}{13}{15}{20}

    \multicolumn{5}{c}{{Hi4D dataset~\cite{yin2023hi4d}}} \\

     \viscontact{vis_contact/vis_interhuman}{3988}{10}{15}{20}{25}

    \multicolumn{5}{c}{{Interhuman dataset~\cite{liang2024intergen}}} \\
    
     \viscontact{vis_contact/vis_3people}{Ballet_003_001_split1_00}{05}{10}{15}{19}

      \multicolumn{5}{c}{Multi-person interaction animation} \\
    
    \end{tabular}
    }
\captionof{figure}{
\textbf{Interactive pose animation} on in-domain datasets (Inter-X\cite{xu2024inter}, Dual-Human~\cite{fang2024capturing}), out-of-domain dataset (Duolando~\cite{siyao2024duolando}, Hi4D~\cite{yin2023hi4d}, Interhuman~\cite{liang2024intergen}), and random composed multi-person pose. Each row: left—interactive pose, right—animation sequence. Our learned interactive pose prior is universal, generalizing across datasets and enabling multi-person interactions (6th row) without modification or retraining.}
\label{fig:exp-c2m-results}
\vspace{-1em}
\end{table}

\begin{table}[t]
     \centering
     \small
    {
    \setlength{\tabcolsep}{0.3em} %
    \renewcommand{\arraystretch}{1.3}
    \begin{tabular}{lcccc|gg}
        \toprule
        \textbf{Method} & \textbf{FID} $\downarrow$ & \textbf{Pre.} $\uparrow$ & \textbf{Rec.}$\uparrow$ & \textbf{Div. }$\rightarrow$ & \textbf{CR.}$\rightarrow$ & \textbf{Pene.}$\downarrow$ \\
        \midrule
        GT & 0.3 & 1.0 & 1.0 & 10.1 & 70.6 & 3.8\\
        \midrule
        MDM*~\cite{tevet2023human} & 62.6 & \textbf{0.79} & 0.20 & 9.8 & 66.4 & 5.3 \\

        ComMDM~\cite{shafir2024human} & 88.8 & 0.37 & 0.49 & 10.9 
        & 44.3 & 4.7 \\
        
        RIG~\cite{tanaka2023role} & 65.2 & 0.46 & 0.65 & 10.6
        & 44.3 & \textbf{4.3}\\
        
        InterGen~\cite{liang2024intergen} & 56.6 & 0.57 & 0.46 & \textbf{10.1}
        & 50.9 & \textbf{4.3 }\\
        
        Ours & \textbf{22.6} & 0.58 & \textbf{0.72} & 10.2 & \textbf{68.1} & 5.0 \\
        \bottomrule
    \end{tabular}
    }
    \caption{\textbf{Unconstrained interaction synthesis comparison} on Inter-X~\cite{xu2024inter} dataset. $\rightarrow$ means the closer to ground truth the better the result. Method in $*$ is adapted from ours for two-person interaction. 
    Our method largely outperforms others in motion quality and contact ratio, naturally ensuring physical contact and motion realism by anchoring on interactive poses.}
\label{table:exp_uncond}
\vspace{-0.5em}
\end{table}

\begin{table}[t]
     \centering
     \small
    {
    \setlength{\tabcolsep}{0.05em} %
    \renewcommand{\arraystretch}{1.3}
   \begin{tabular}{lcc|ggcc|gg}
   \toprule
     & \multicolumn{4}{c}{\textbf{Inter-X}} &
    \multicolumn{4}{c}{\textbf{Dual-Human}} \\
    \cmidrule(lr){2-5} \cmidrule(lr){6-9}
   \textbf{Method} & \textbf{FID}$\downarrow$ &   %
    \textbf{Div.}$\rightarrow$ & 
    \textbf{CR.} $\rightarrow$ & \textbf{Pene.}$\downarrow$ &
   \textbf{FID}$\downarrow$ & %
    \textbf{Div.}$\rightarrow$ & 
    \textbf{CR.} $\rightarrow$ & \textbf{Pene.}$\downarrow$ \\
    \midrule
     GT  & 0.3 & 10.1 & 70.6 & 3.8  & 2.1  & 12.0 & 70.4 & 3.4\\
     \midrule 
      InterGen* & 18.9 & 10.6 & 44.4 &   \textbf{4.3} & 88.8 & \textbf{11.9} & 44.3 & \textbf{4.1}\\
     w/o anchor & 7.1 & 9.8 & 67.3 & 5.1&
     36.9 & 11.6 & 70.7 & 4.5\\
     \textbf{-} time & 6.3 & 10.3 & 66.9 & 5.2 & 30.3 & 12.6 & 67.3 & 5.1  \\
     \textbf{-} joints & 5.6 & 10.0 & 67.6 & 5.1 & 29.9 & 12.3 & 70.2 & 4.4\\
     random-pose & 5.8 & \textbf{10.1} & 67.4 & 5.1 & 30.1 & 12.3 & 69.3 & 4.5\\
     ours & \textbf{5.0} & 9.9 & \textbf{68.5} & 5.1 & \textbf{24.2} & 11.8 & \textbf{70.4} & 4.5\\
    \bottomrule
    \end{tabular}
    }
    \caption{\textbf{Interactive pose animation comparison} on Inter-X~\cite{xu2024inter} and Dual-Human~\cite{fang2024capturing} dataset. InterGen* is adapted to take interactive poses input but lacks explicit interaction modeling, limiting its use of pose priors. Interactive pose anchoring, condition encoding, and interactive frames are crucial for the performance.}
\label{table:exp-contact-ablation}
\vspace{-1.5em}
\end{table}

\noindent\textbf{Implementation details.}
We extract interactive poses by detecting SMPL-X vertices contacts~\cite{muller2024generative} below a threshold in each mocap dataset within a 3s window. The interactive pose animator has 8 layers (latent dim 1024) and is trained using AdamW~\cite{loshchilov2017decoupled} (LR $1e\text{-}4)$. All loss weights are 1 except $\lambda_{\text{inter}}=0.5$.To handle real-world noise, we augment training by adding Gaussian noise (scale $0.02$) to interactive poses. At inference, DDIM~\cite{song2020denoising} samples 50 steps, generating $3s$ motions at $10$fps in $0.24s$ on an A100. The interactive pose generator follows a similar setup with $p_{\text{text}}=0.8$, $p_{\text{pose}}=0.2$, and a frozen CLIP-ViTL/14~\cite{radford2021learning} text encoder. The pose generation take $0.21s$. Models are trained for 4000 epochs with batch sizes of 256 (pose animator) and 512 (pose generator). Please see \cref{sec:implement} for details.

\noindent\textbf{Datasets.}
We train and test our model on two large-scale datasets: Inter-X~\cite{xu2024inter} ($11\text{k}$ sequences) and Dual-Human~\cite{fang2024capturing} ($2\text{k}$ sequences). We follow the official split for Inter-X and use a 3:1 training-testing split for Dual-Human, excluding non-interactive motion sequences.

\noindent\textbf{Metrics.}
We follow the evaluation metrics in ~\cite{shafir2024human, tevet2023human, raab2023modi}: \textbf{Frechet Inception Distance (FID)}, the feature distribution against ground truth (GT). We compute it by training a motion autoencoder to encode motion into features for each task; \textbf{Precision (Pre.)}, the likelihood that generated motions fall within the real distribution; \textbf{Recall (Rec.)}, the likelihood that real motions fall within the generated distribution; \textbf{Diversity}, the variance of generated motions. We also evaluate the physics plausibility via \textbf{Contact Frame Ratio (CR., $\%$)}—proportion of frames with two-person contact—and averaged \textbf{Inter-person Penetration (Pene}., cm).

\subsection{Effectiveness of Anchoring on Interactive Poses}
Previous works model human-human interaction dynamics either by finetuning on single-person motion priors with interaction data (e.g., ComMDM~\cite{shafir2024human}, RIG~\cite{tanaka2023role}) or by learning interaction dynamics from scratch (e.g., InterGen~\cite{liang2024intergen}). In this work, we model interaction dynamics by anchoring on proximal interactive poses. To evaluate the effectiveness of these approaches, we employ a simple task---unconstrained generation. We further adapt MDM~\cite{tevet2023human} to accommodate two-person motions in our setting. Ponimator seamlessly supports unconstrained generation by setting $\mathbf{m}_a=0$ and $\mathbf{m}_c=0$. Experimental results on our dataset collection from Inter-X~\cite{xu2024inter} are shown in \cref{table:exp_uncond}. We observe that previous methods~\cite{shafir2024human, tanaka2023role, liang2024intergen} struggle to synthesize close-contact interactions, while the adapted MDM*~\cite{tevet2023human} exhibits lower interaction motion quality. In contrast, by simply anchoring on interactive poses, our model achieves superior motion realism (FID of 22.6) and physical contact (contact ratio of 68.1).

\subsection{Interactive Pose Animation}
To evaluate the interactive pose animator, we compare against baselines and key ablations on Inter-X~\cite{xu2024inter} and Dual-Human~\cite{fang2024capturing} datasets in \cref{table:exp-contact-ablation}. We ablate key components of pose animator: \textbf{w/o anchor} removes interactive pose anchoring, replacing the denoising target $\mathbf{z}_0$ with $\{\mathbf{x}_i\}_{i=1}^N$; \textbf{-~time} removes the interaction time encoding $\mathbf{m}_I$; \textbf{-~joints} removes joints condition encoding; \textbf{InterGen}* replaces text conditions with interactive pose condition while keeping all other settings unchanged; \textbf{random-pose} uses random instead of interactive frames as anchor. All baselines are trained under the same setting. \cref{table:exp-contact-ablation} highlights the importance of interactive pose anchoring and interaction conditioning. InterGen* overlooks input poses, resulting in poorer performance. In contrast, our method explicitly models interaction and contact and achieves better results.

\noindent\textbf{Universal interactive pose prior.}
We visualize the animated motion in \cref{fig:exp-c2m-results} on in-domain datasets (Inter-X\cite{xu2024inter}, Dual-Human~\cite{fang2024capturing}) and out-of-domain datasets (Duolando~\cite{siyao2024duolando}, Hi4D~\cite{yin2023hi4d}, Interhuman~\cite{liang2024intergen}). Our approach generalizes to unseen subjects and interactions using the universal interactive pose prior. Our model is surprisingly capable of generating interactions beyond two persons without modification or retraining (see last row in \cref{fig:exp-c2m-results}). %

\noindent\textbf{Open-world two-person image animation.}
Our model generalizes to open-world images by extracting interactive poses from FlickrCI3D~\cite{fieraru2020three} dataset using~\cite{muller2024generative}. As shown in \cref{fig:exp-buddi-results}, it transforms static poses into realistic motion. 

\begin{table}[t]
     \centering
     \small
    {
    \setlength{\tabcolsep}{0.3em} %
    \renewcommand{\arraystretch}{1.3}
   \begin{tabular}{lccc|gg}
   \toprule
   \textbf{Method} & \textbf{FID}$\downarrow$ & \textbf{Div.}$\rightarrow$ & 
   \textbf{MModality}$\uparrow$ &
   \textbf{CR.} $\rightarrow$ & \textbf{Pene.}$\downarrow$ \\
    \midrule
 GT & 0.06 & 6.78 & - & 70.6 & 3.8 \\
 \midrule
 InterGen & 2.87 & 6.76 & 1.42 & 39.8 & \textbf{3.9} \\
 w/o anchor & 2.74 & \textbf{6.78} & 1.41 & 39.0 & 4.0 \\
Ours & \textbf{1.82} &\textbf{ 6.78} & \textbf{1.46 }&\textbf{ 45.9} & 4.3  \\
    \bottomrule
\end{tabular}
    }
    \caption{\textbf{Text-to-interaction synthesis} results on Inter-X \cite{xu2024inter} dataset. Our unified pipeline outperforms end-to-end w/o interactive pose as anchor method in short-term interaction synthesis.}
\label{table:exp_t2m}
\end{table}
\begin{table}
    \centering
    \footnotesize
    \setlength{\tabcolsep}{0.2em} %
    \resizebox{\linewidth}{!}{
        \begin{tabular}{cccccc}
        
        \rotatebox{90}{\hspace{0pt}InterGen~\cite{liang2024intergen}} &
        \vistm{vis_t2m/vis_intergen}{s03_Push_1}{05}{10}{15}{20}{25}[5cm 5cm 5cm 5cm]

        \rotatebox{90}{\hspace{0pt}W/o anchor } &
        \vistm{vis_t2m/vis_full}{s02_Push_10}{05}{10}{15}{20}{25}[5cm 5cm 5cm 5cm]

        \rotatebox{90}{\hspace{10pt}Ours} &
        \vistm{vis_t2m/vis_chi3d}{s03_Push_14}{05}{10}{15}{20}{25}[3cm 5cm 5cm 5cm] 
            
        \end{tabular}
    }
\captionof{figure}{\textbf{Text-to-interaction comparison} for "push". 
Anchored on interactive poses, our method achieves better contact and more realistic dynamics than InterGen~\cite{liang2024intergen} and the end-to-end baseline.
}

\label{fig:exp-t2m-comparison}
\vspace{-0.5em}
\end{table}

\subsection{Interaction Motion Generation}
We evaluate interaction motion generation on the Inter-X dataset~\cite{xu2024inter} using text and single-person poses.

\begin{table}[t]
     \centering
     \small
    {
    \setlength{\tabcolsep}{0.3em} %
    \renewcommand{\arraystretch}{1.3}
   \begin{tabular}{lcccc|gg}
   \toprule
   \textbf{Method} & \textbf{FID}$\downarrow$ & \textbf{Pre.}$\uparrow$ &
    \textbf{Rec.}$\uparrow$ & \textbf{Div.}$\rightarrow$ & \textbf{CR.}$\rightarrow$ & \textbf{Pene.}$\downarrow$ \\
    \midrule
 GT & 0.3 & 1.0 & 1.0 & 10.1 & 70.6 & 3.8  \\
 \midrule
 w/o anchor & 40.0 & 0.87 & 0.43 & 9.6 & 67.5 & \textbf{5.0} \\
 Ours & \textbf{27.8} & \textbf{0.91} & \textbf{0.48} & \textbf{9.7} & \textbf{73.3} & 5.2 \\
\bottomrule

\end{tabular}
    }
    \caption{\textbf{Single pose-to-interaction synthesis} results on Inter-X~\cite{xu2024inter} dataset. Compared to without anchor baseline, our method uses interactive poses for more effective interaction modeling.}
\label{table:exp-singlepose}
\vspace{-0.5em}
\end{table}

\begin{table}
    \centering
    \footnotesize
    \setlength{\tabcolsep}{0.2em} %
    \resizebox{\linewidth}{!}{
        \begin{tabular}{cc|cccc}

         & {Single Pose} & 
        \multicolumn{4}{c}{{Interaction Generation (left$\rightarrow$right: time steps)}} \\

         \rotatebox{90}{\hspace{0pt}W/o anchor} & \vispm{vis_p2m/vis_random}{G015T004A024R023}{10}{15}{20}{25}[3cm 5cm 3cm 5cm] \\ 
         
         \rotatebox{90}{\hspace{10pt}Ours} & \vispm{vis_p2m/vis_interx}{G015T004A024R023}{10}{15}{20}{25}[3cm 5cm 3cm 5cm]\\ 

    \end{tabular}
    }
\captionof{figure}{\textbf{Single pose-to-interaction comparison} on Inter-X dataset~\cite{xu2024inter}. Compared to the model without interactive pose anchors, our method generates more natural human interactions.}
\label{fig:exp-pose-compare-results}
\vspace{-0.5em}
\end{table}
\begin{table}
    \centering
    \footnotesize
    \setlength{\tabcolsep}{0.2em} %
    \resizebox{1.05\linewidth}{!}{
        \begin{tabular}{c|cccc}

    {Single Pose } & \multicolumn{4}{c}{{Interactive Generation (left$\rightarrow$right: time steps)}} \\
     \includegraphics[width=0.2\linewidth]{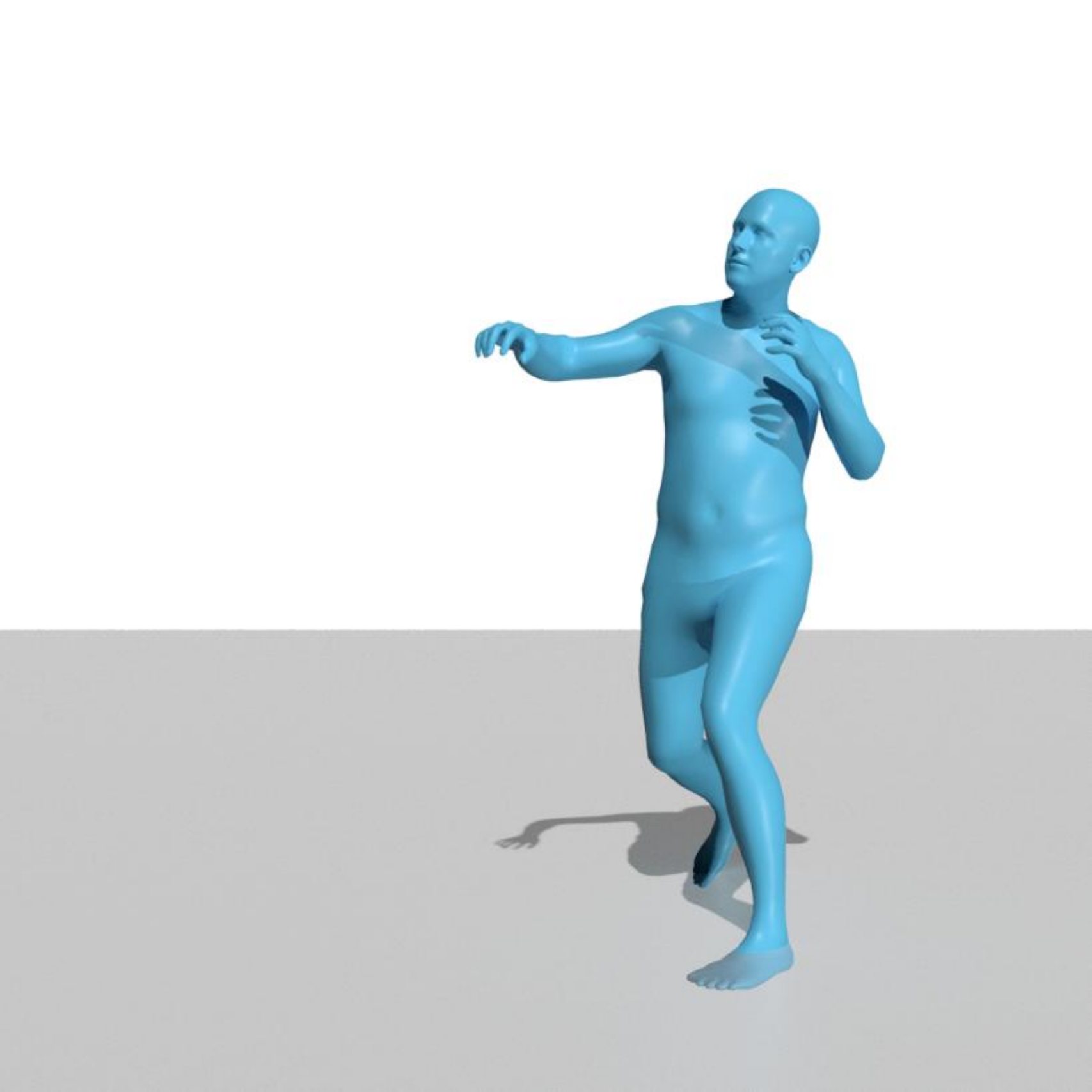} &
     \includegraphics[width=0.2\linewidth]{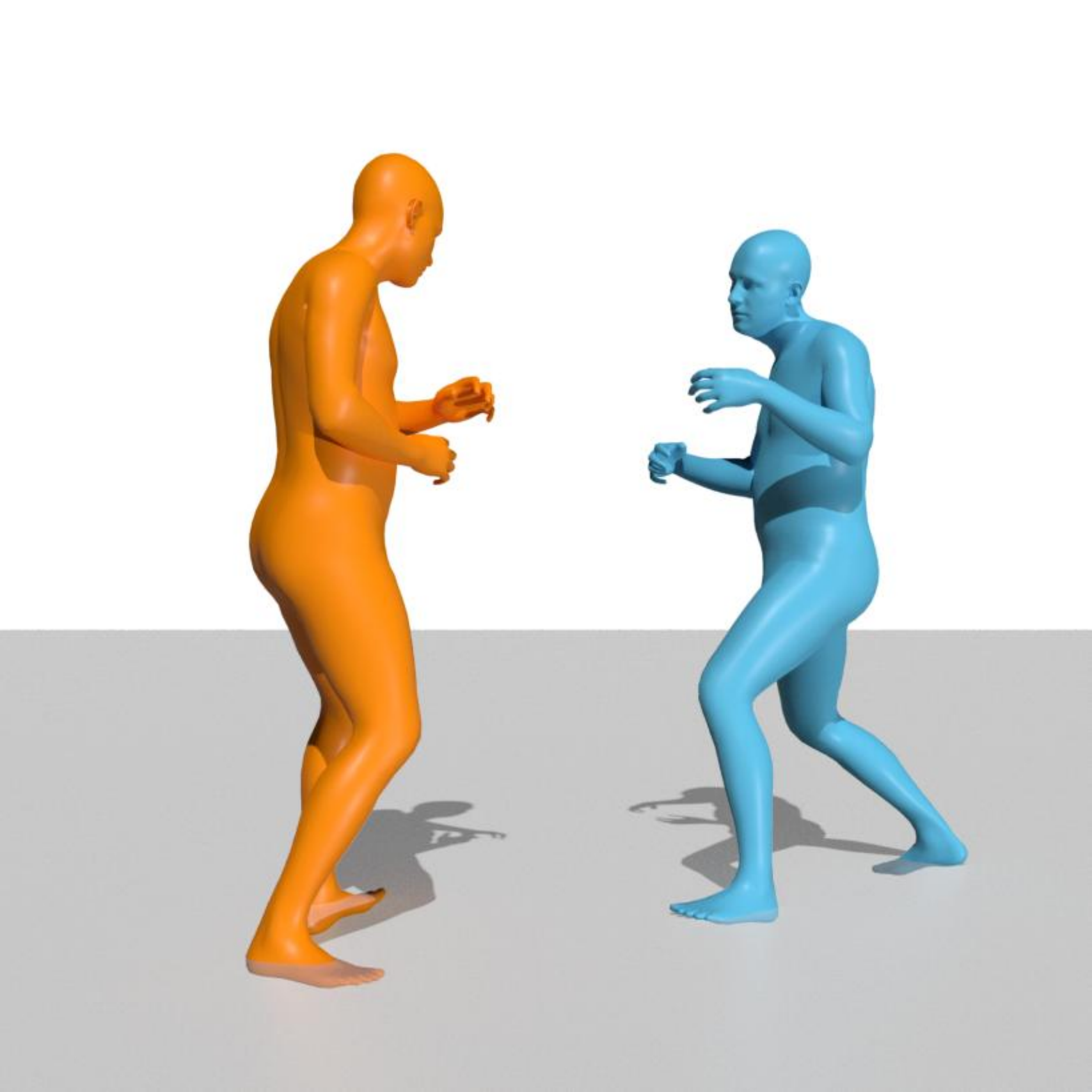} &
      \includegraphics[width=0.2\linewidth]{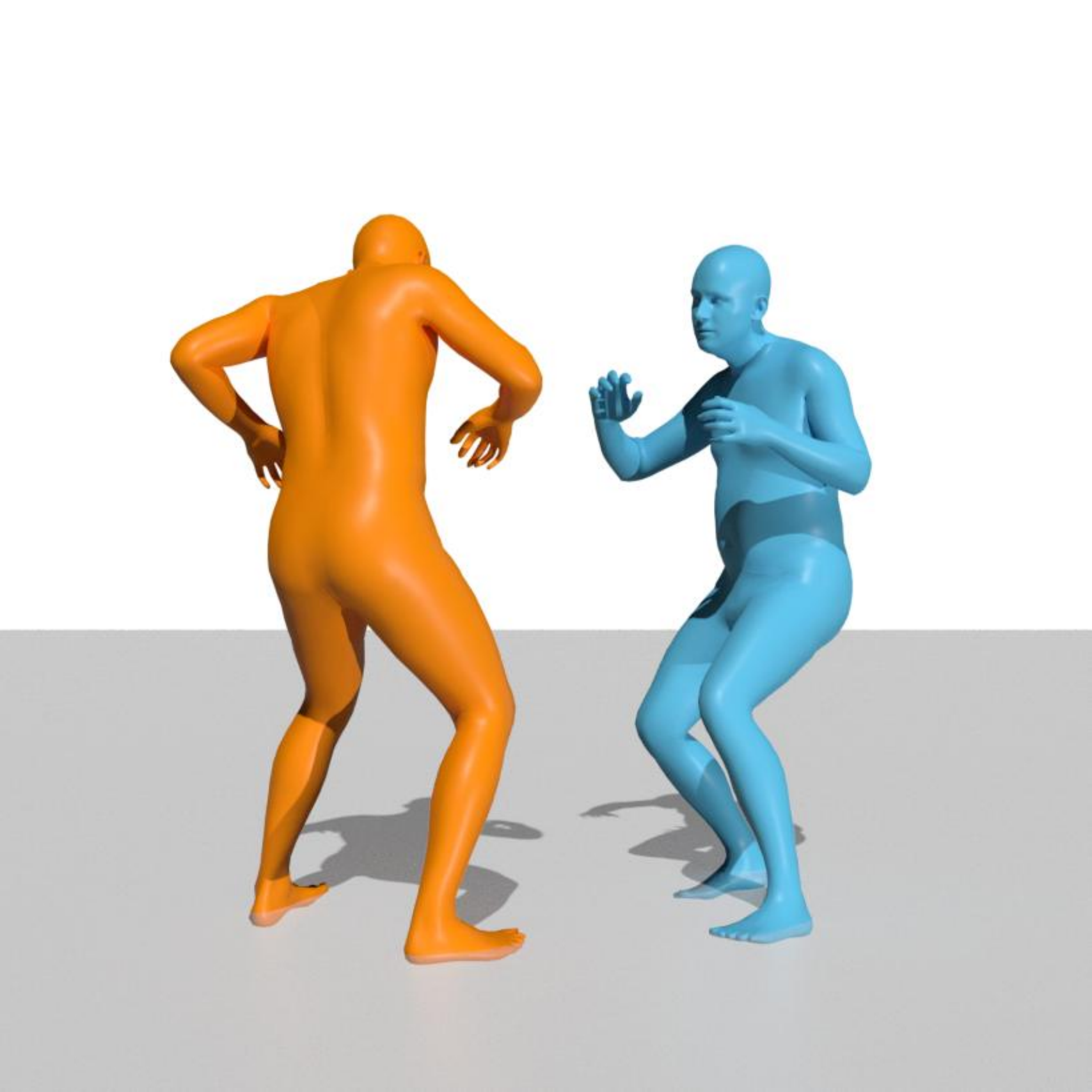} &
       \fcolorbox{magenta}{white}{
       \includegraphics[width=0.2\linewidth]{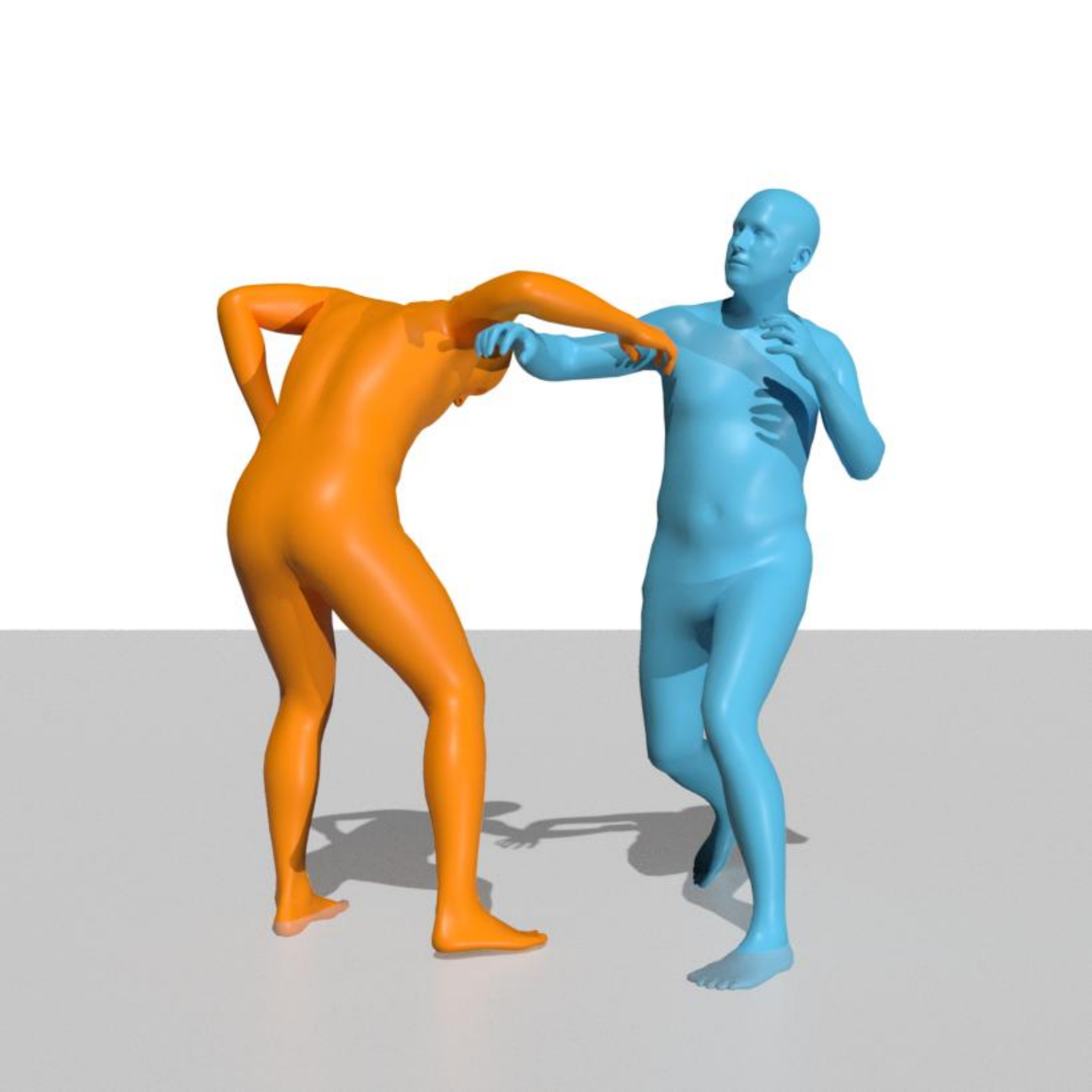}} &
        \includegraphics[width=0.2\linewidth]{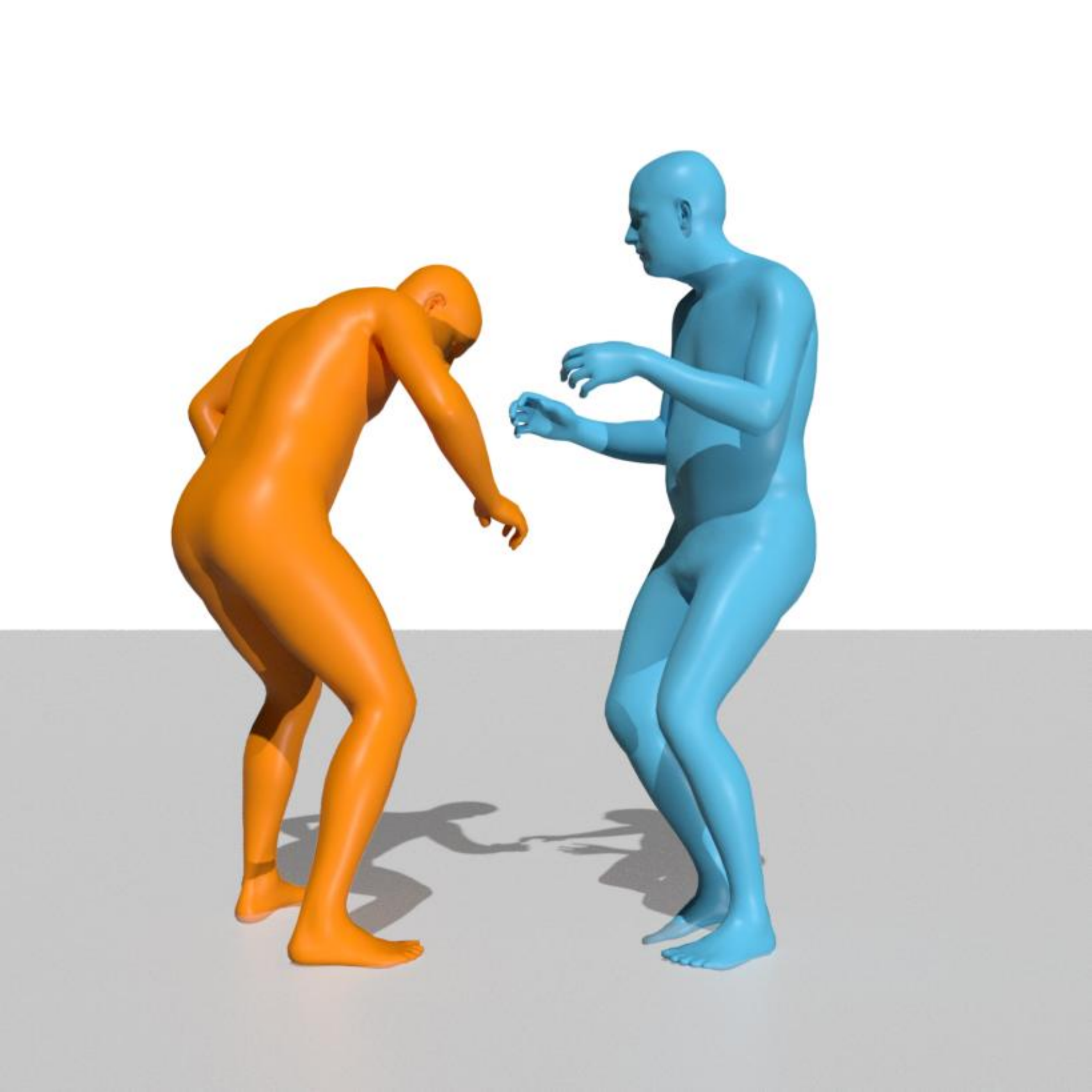} \\
        
    \includegraphics[width=0.2\linewidth]{src_figs/vis_p2m/vis_diverse/pair01_000242_1_no_text/single_pose_crop.pdf} &
     \includegraphics[width=0.2\linewidth]{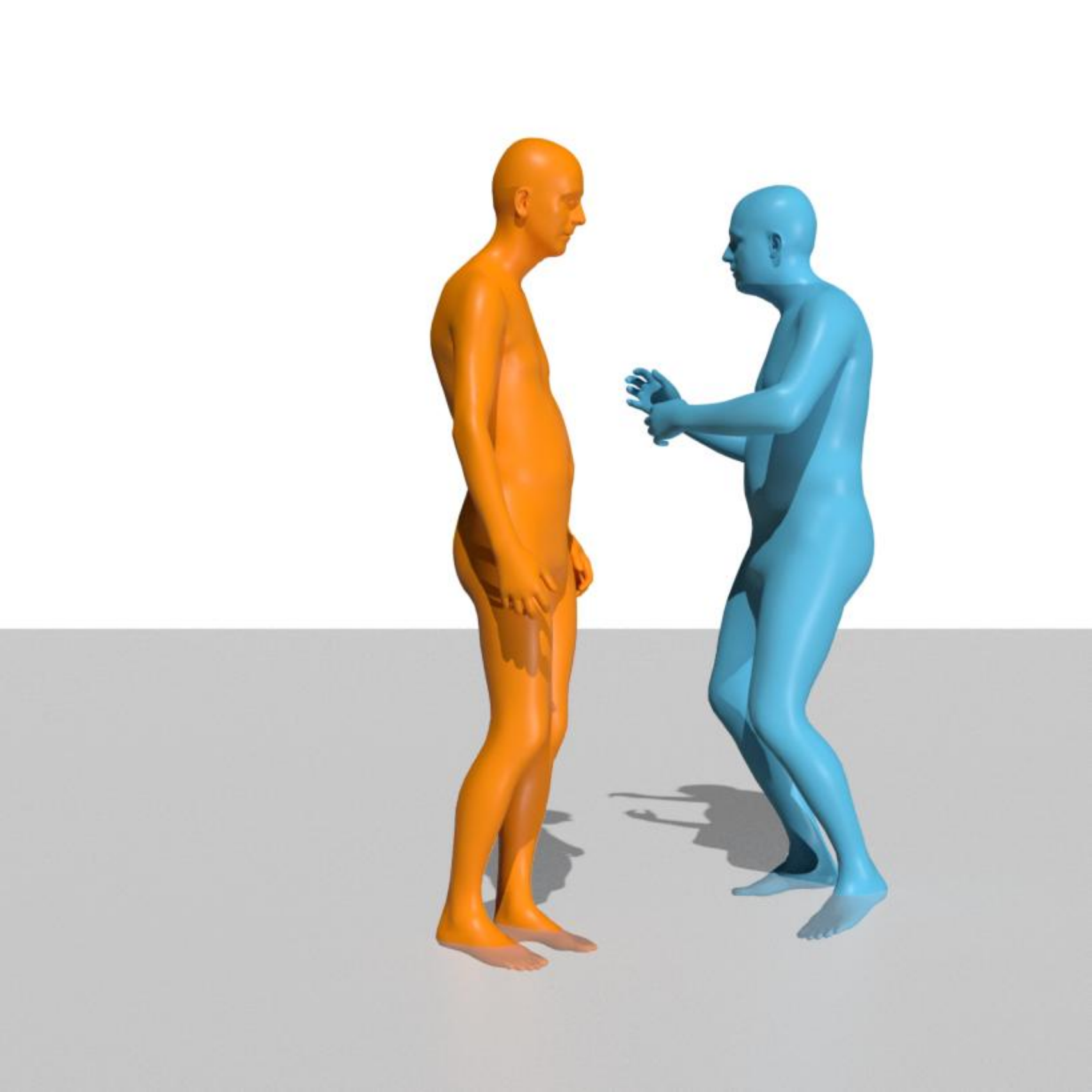} &
       \includegraphics[width=0.2\linewidth]{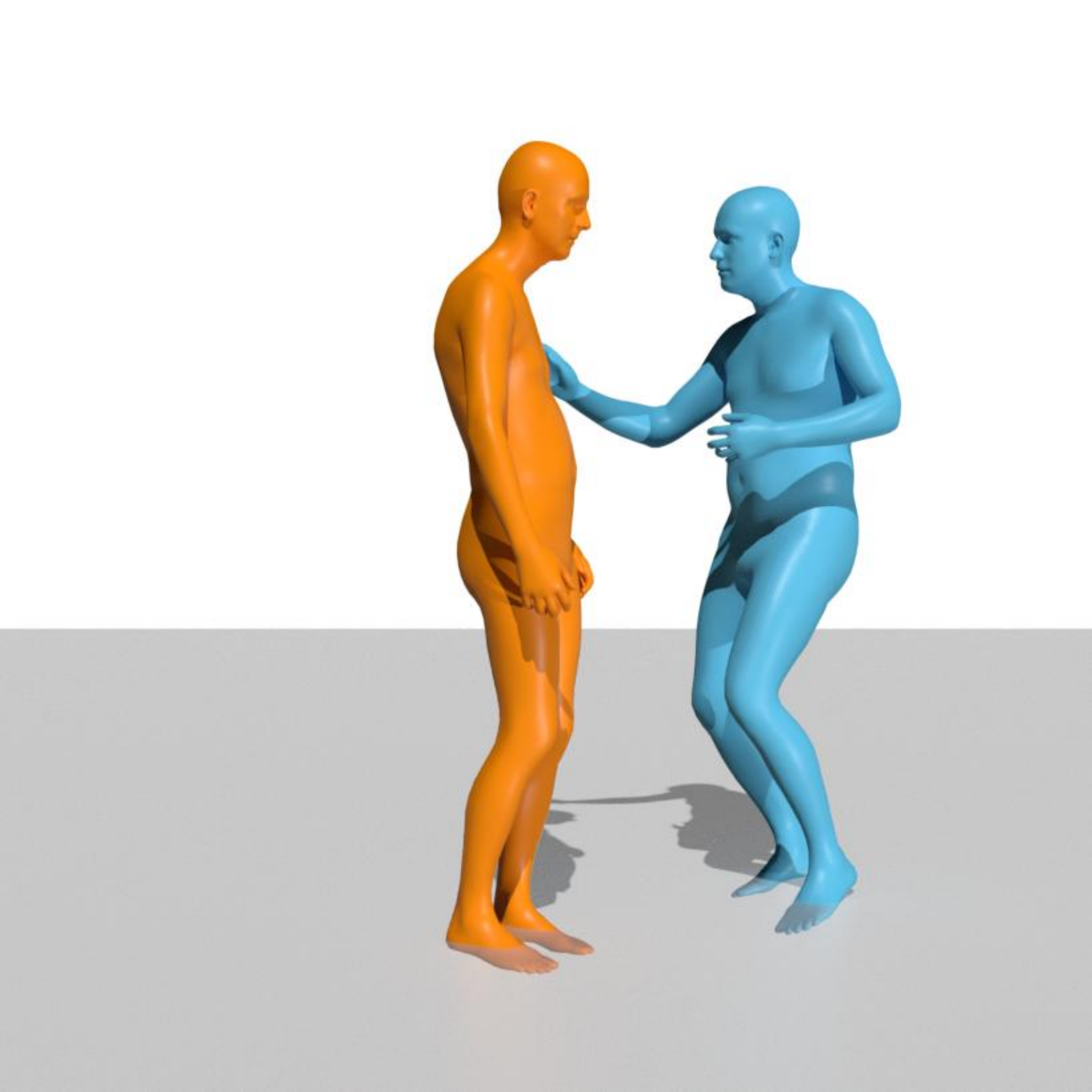} &
        \fcolorbox{magenta}{white}{
        \includegraphics[width=0.2\linewidth]{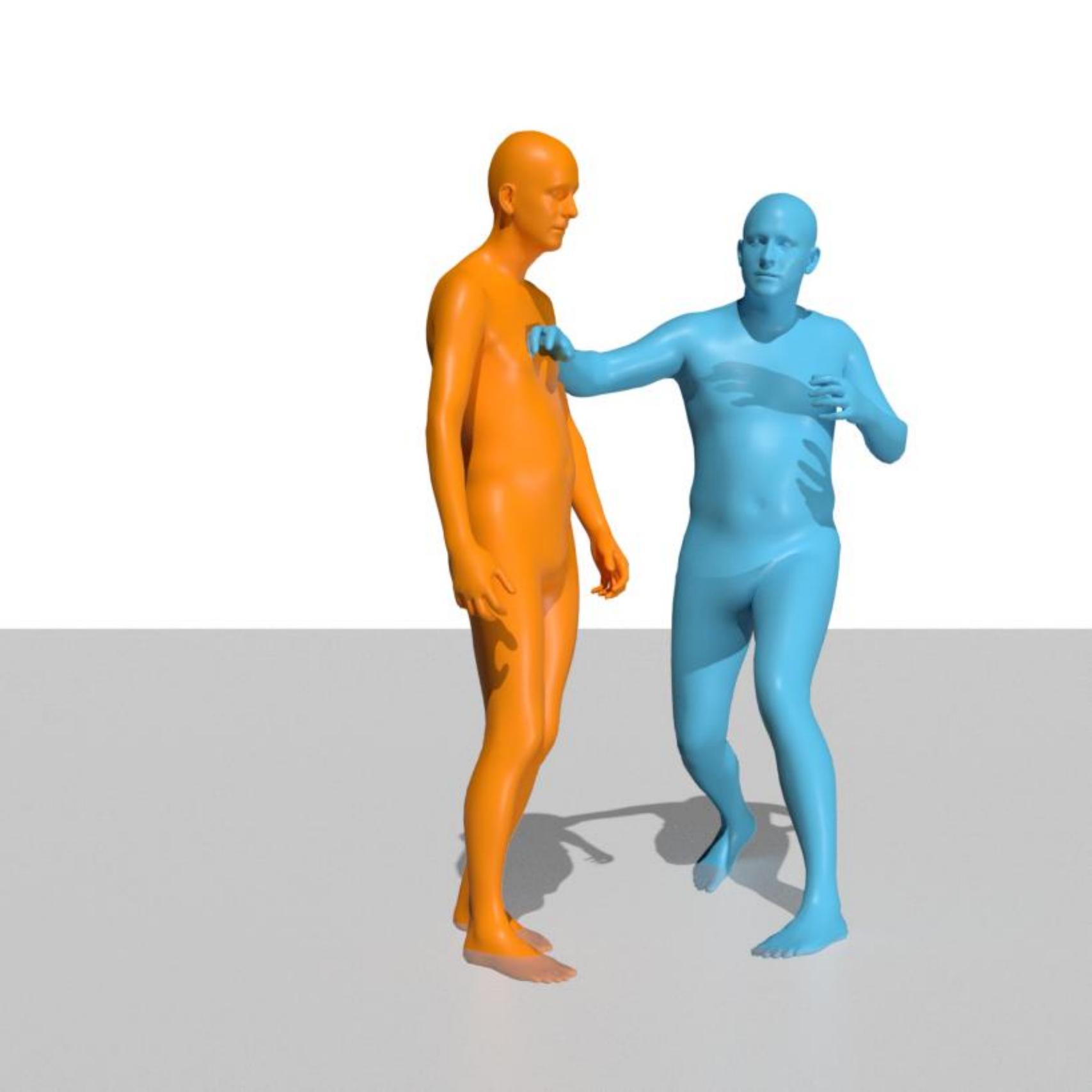}} &
         \includegraphics[width=0.2\linewidth]{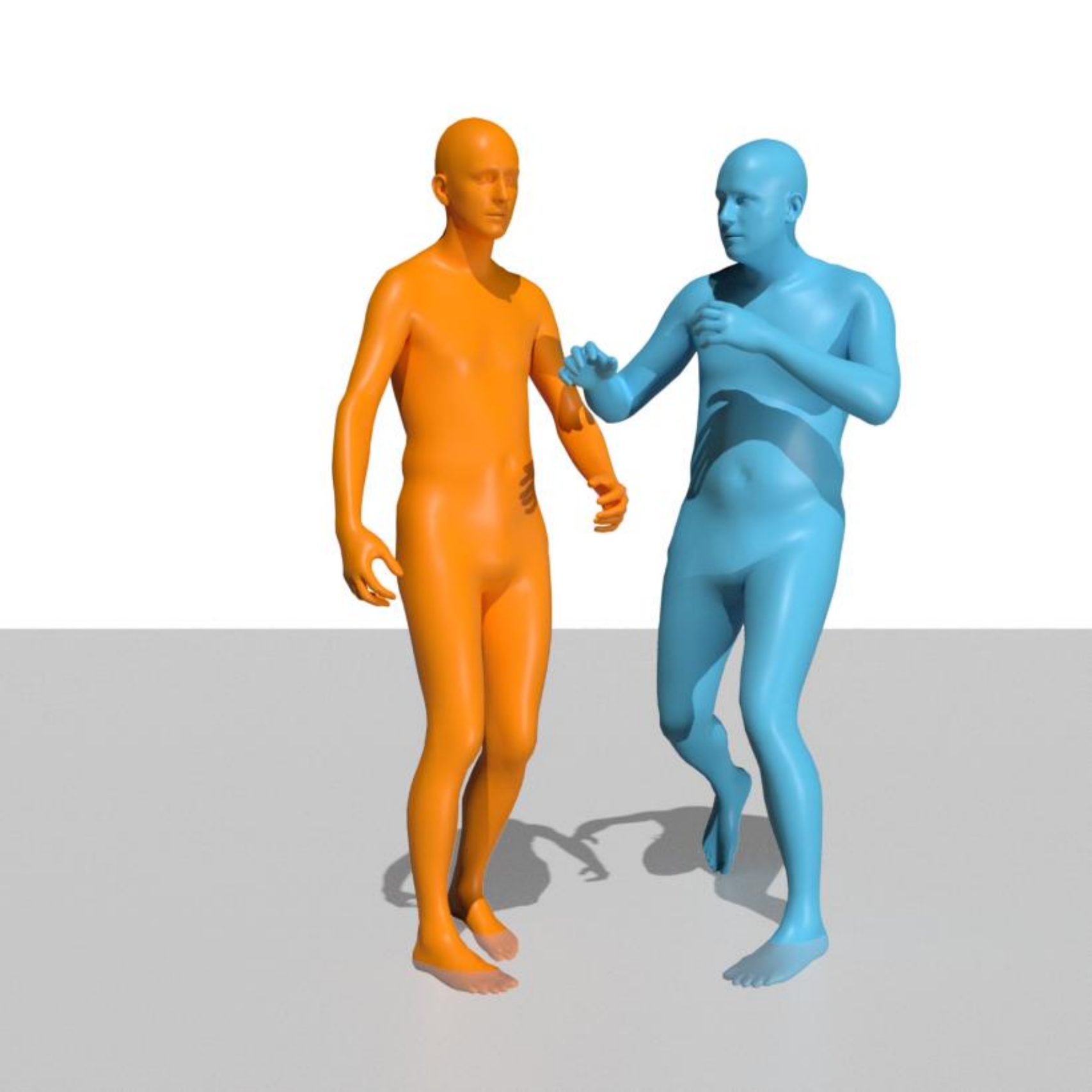} \\

         \includegraphics[width=0.2\linewidth]{src_figs/vis_p2m/vis_diverse/pair01_000242_1_no_text/single_pose_crop.pdf} &
         \includegraphics[width=0.2\linewidth]{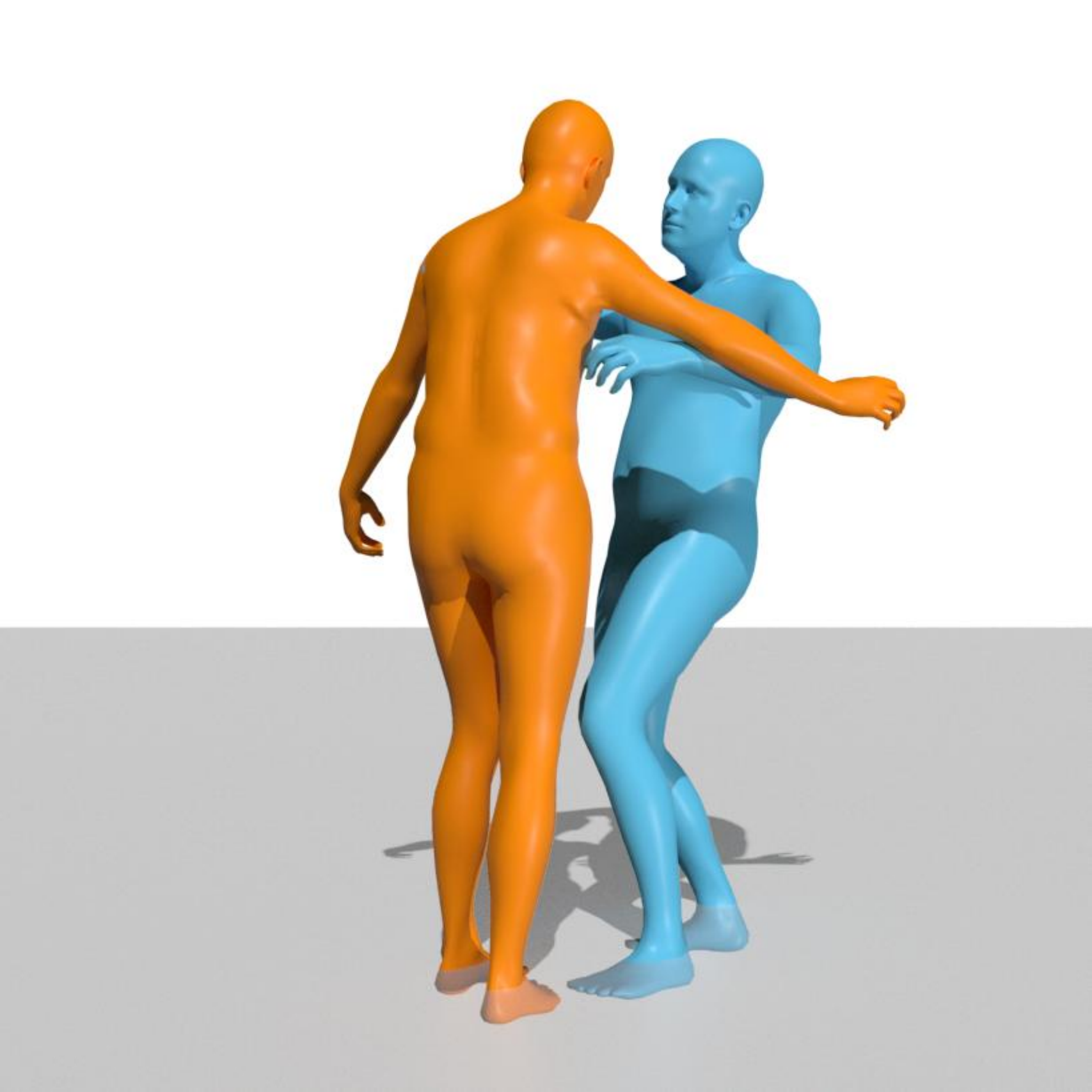} &
         \fcolorbox{magenta}{white}{
         \includegraphics[width=0.2\linewidth]{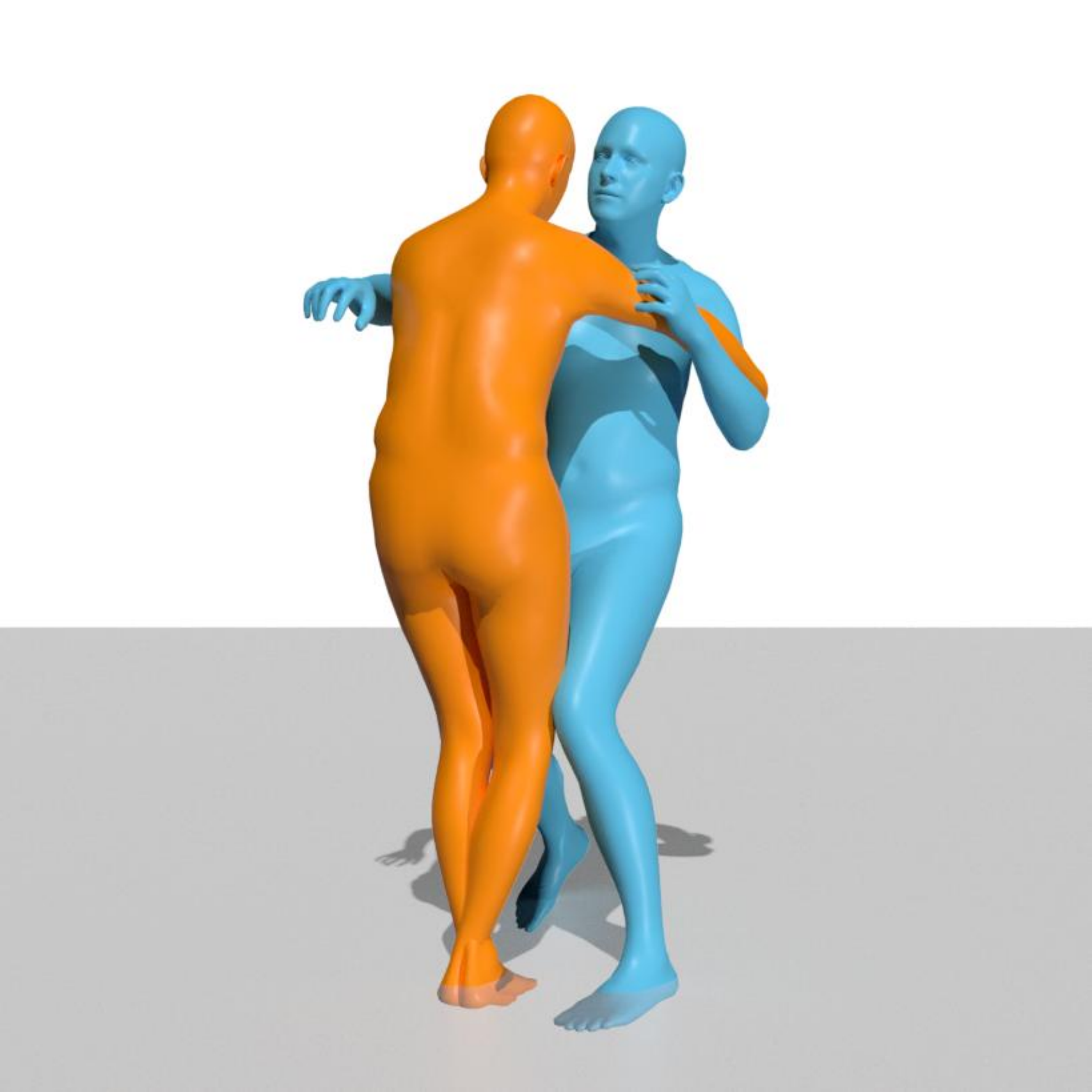}} &
         \includegraphics[width=0.2\linewidth]{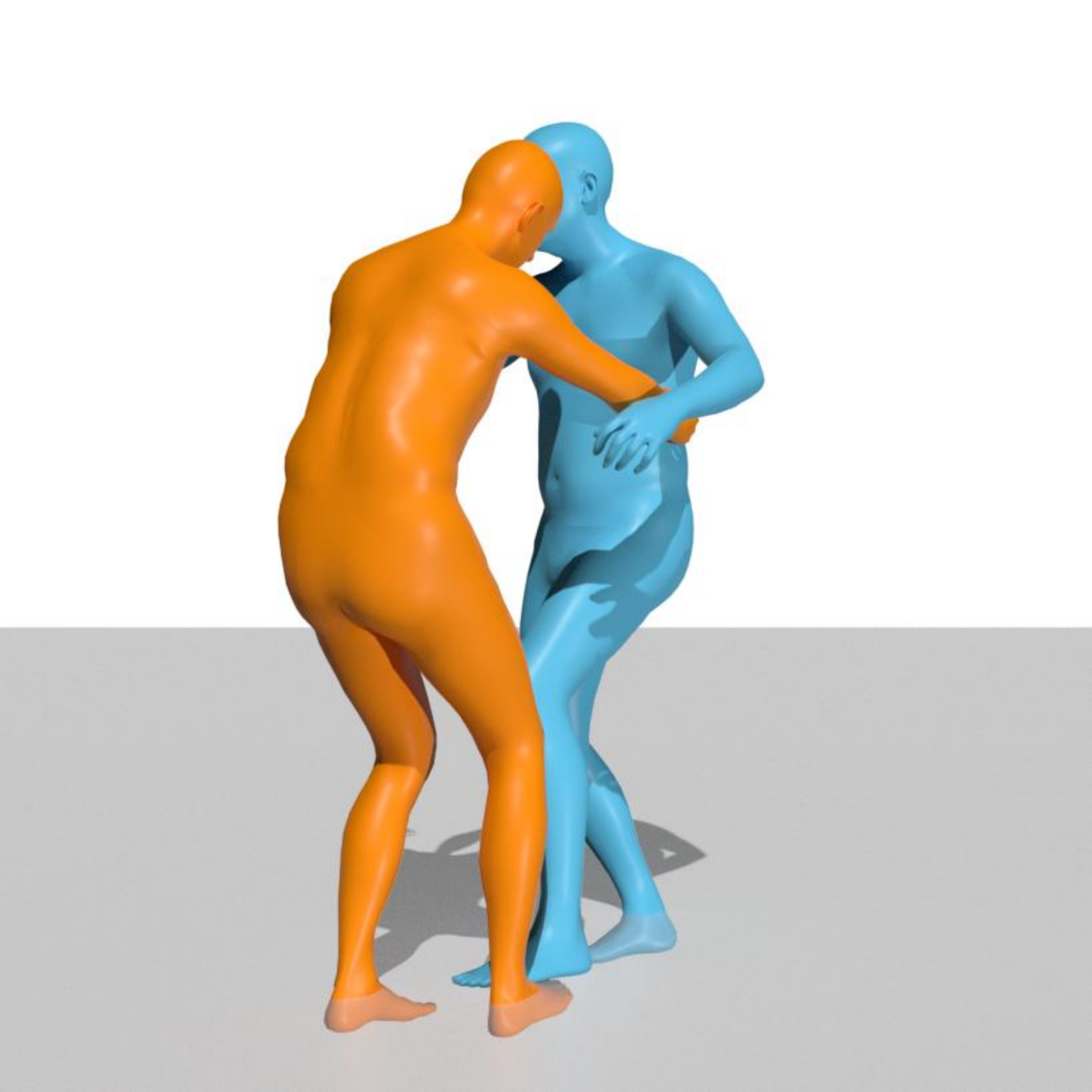} &
         \includegraphics[width=0.2\linewidth]{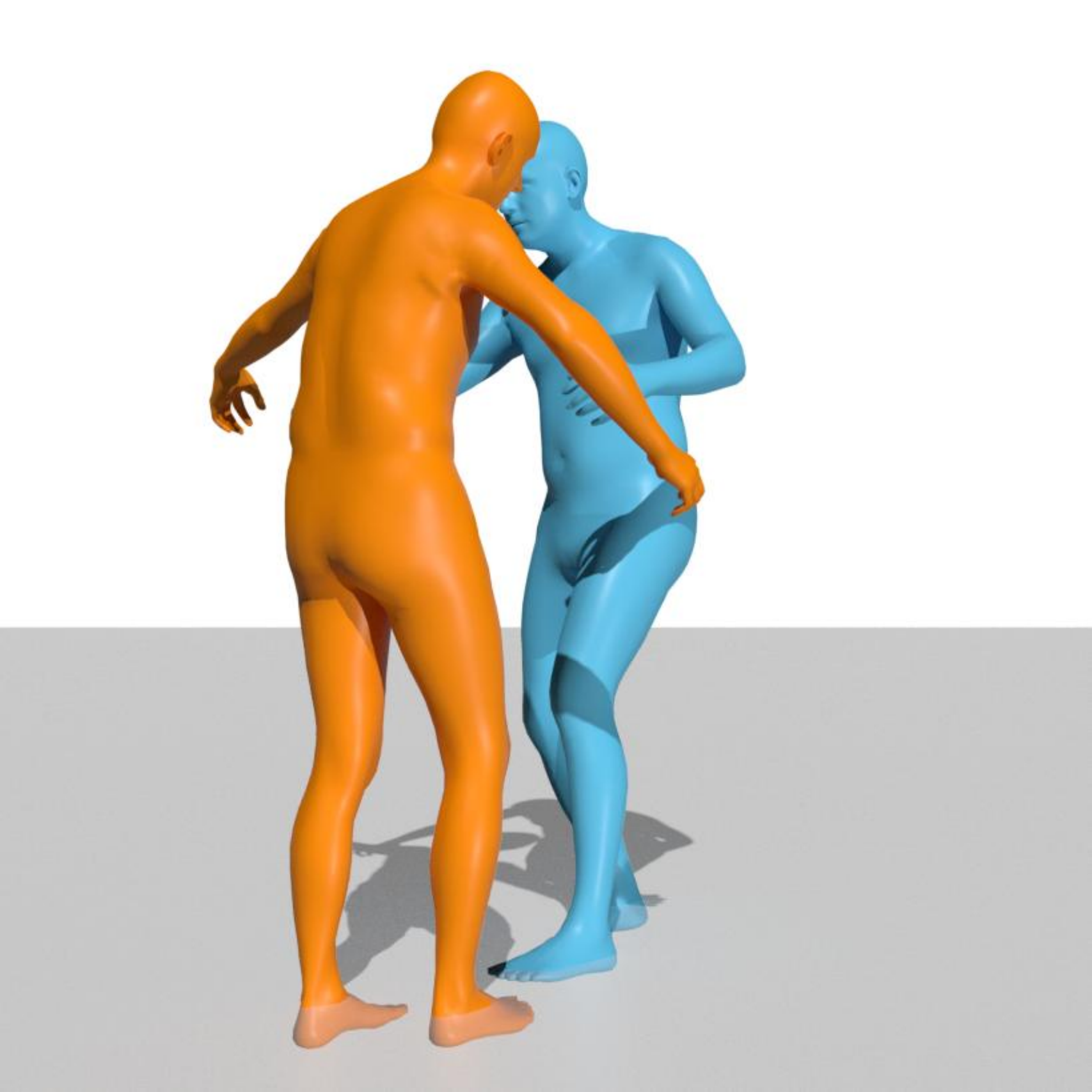} \\
         + "two person hug" \\
    \end{tabular}
    }
\captionof{figure}{\textbf{Diverse interactive motion generation}. From a single pose, our framework generates varied interactive poses (\textcolor{magenta}{magenta box}) and motions (1st, 2nd rows) and text-driven ones (3rd row).}
\label{fig:exp-p2m-diverse}
\vspace{-1.5em}
\end{table}

\noindent\textbf{Text-to-interaction synthesis}
We focus on 3s interaction generation, evaluating FID, Diversity, and \textbf{MModality}—the ability to generate diverse interactions from the same text~\cite{liang2024intergen, tevet2023human}.
We compare with InterGen~\cite{liang2024intergen} and an end-to-end w/o interactive pose baseline, both trained and tested on the same data. As shown in \cref{table:exp_t2m} and \cref{fig:exp-t2m-comparison}, they struggle with contact modeling, while ours excels in short-term interaction generation using interactive pose priors.

\begin{table*}
    \centering
    \footnotesize
    \setlength{\tabcolsep}{0.2em} %
    \resizebox{\linewidth}{!}{
        \begin{tabular}{c|cccc}

        {Input Image} & 
        \multicolumn{4}{c}{{Interaction Generation (left$\rightarrow$right: time steps)}} \\

         \visvideo{vis_motionx}{Back_Flip_Kungfu_wushu_Trim9_clip1}{05}{10}{15}{20}{0.19}[4][magenta] \\
    &
    \includegraphics[width=0.2\linewidth]{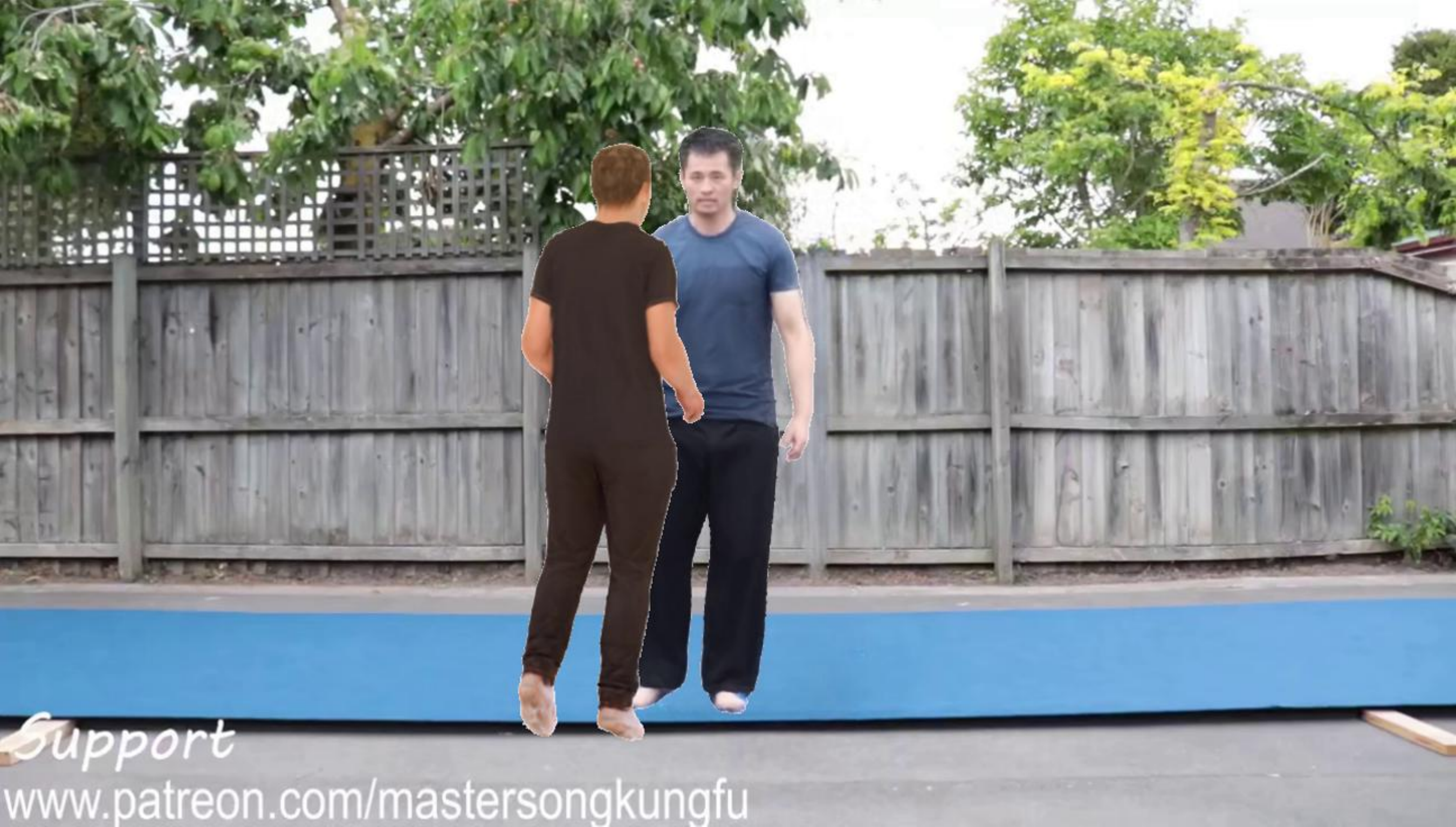} &
    \includegraphics[width=0.2\linewidth]{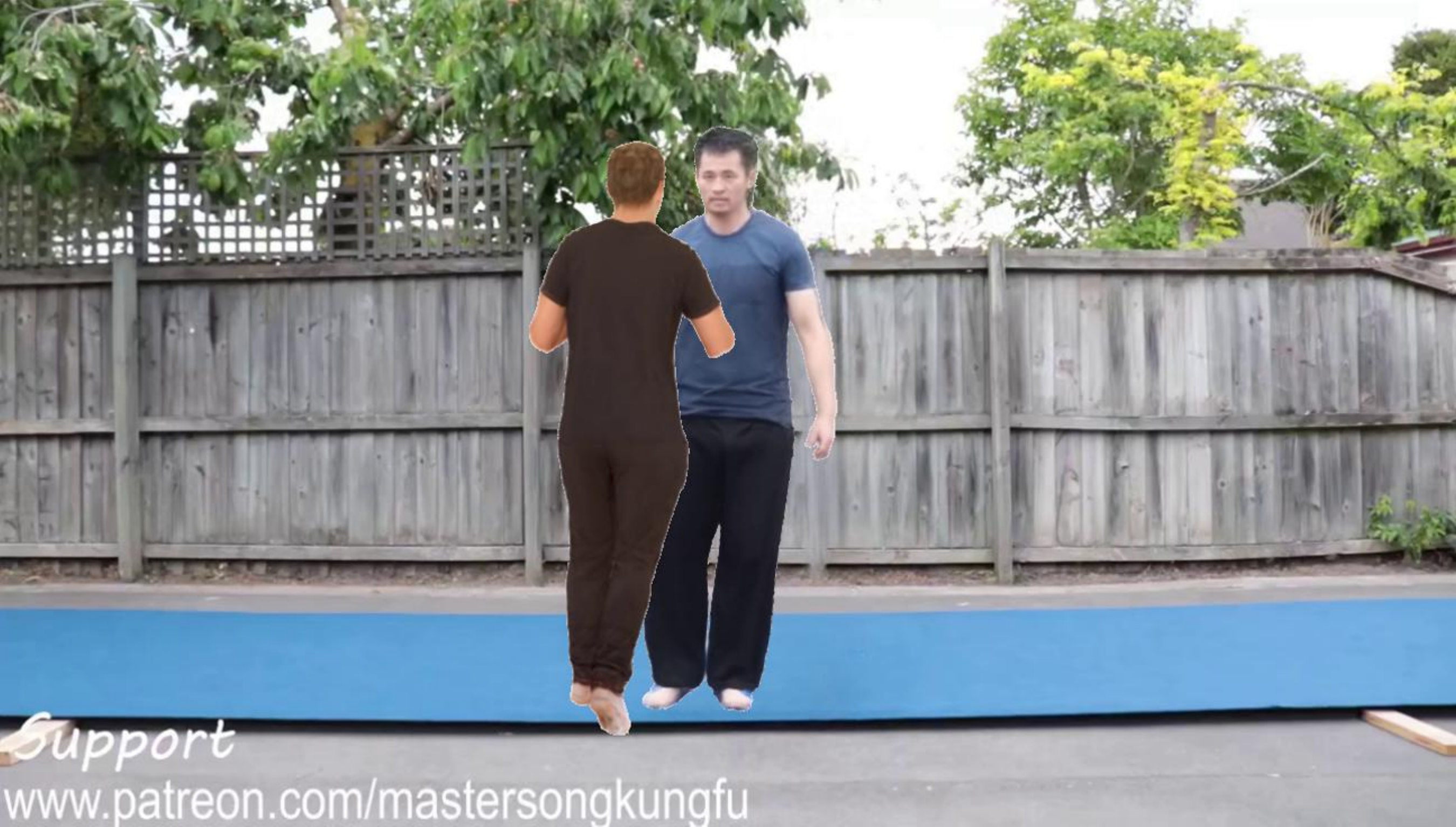} &
    \includegraphics[width=0.2\linewidth]{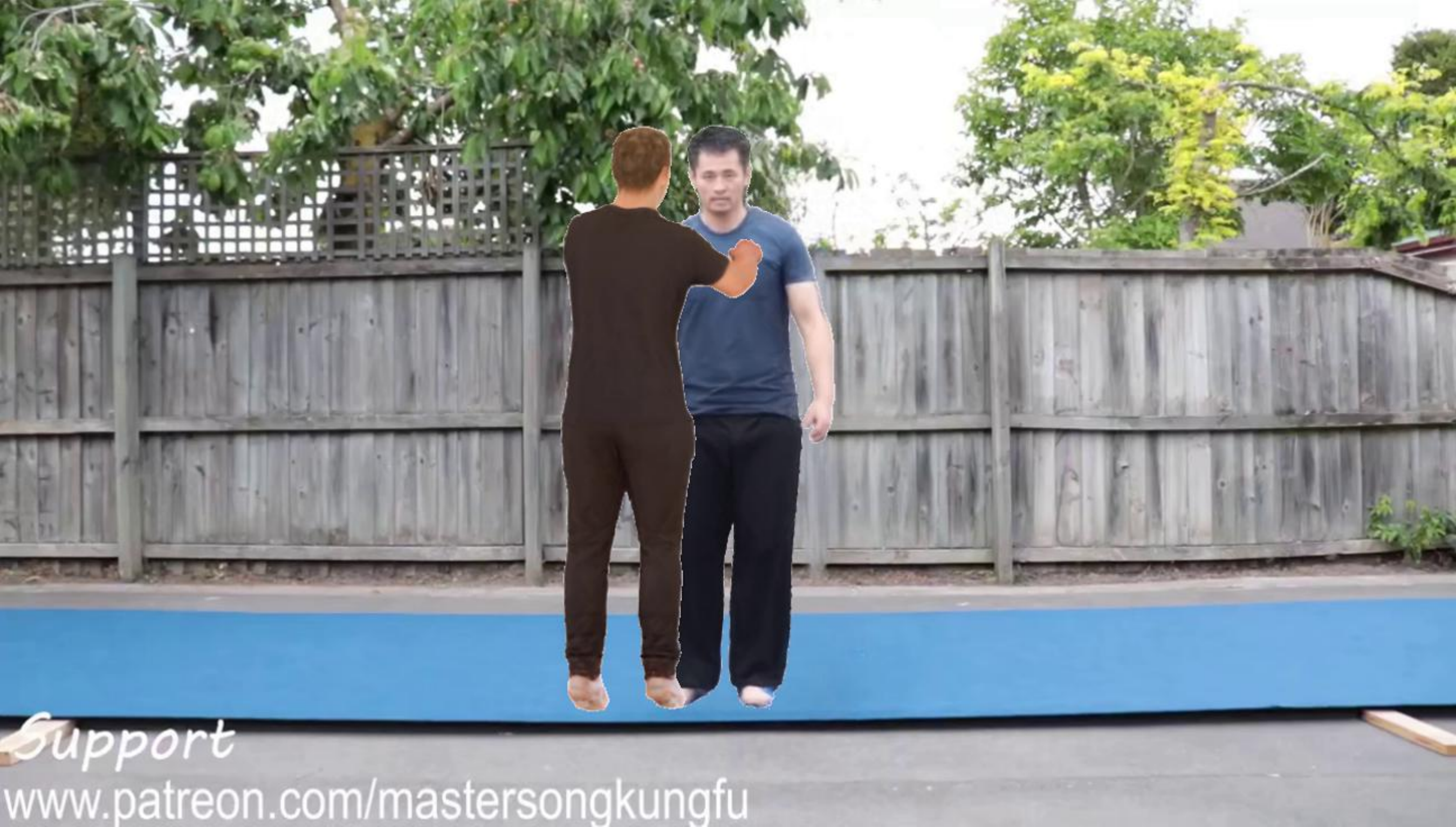} &
    \includegraphics[width=0.2\linewidth]{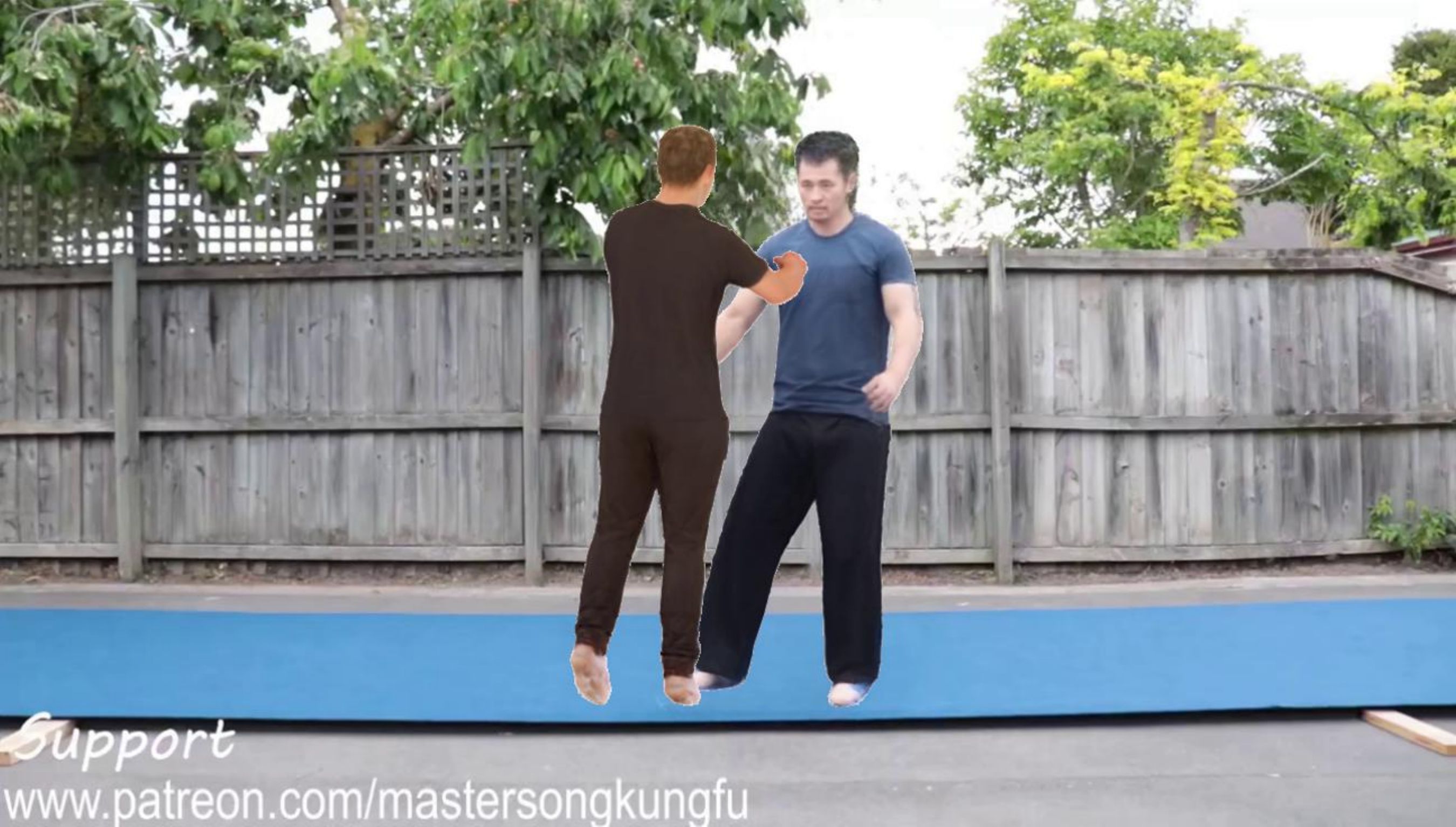} \\
         \hline

      \visvideo{vis_motionx}{Ways_to_Stand_Groomsman_clip1}{05}{10}{15}{20}{0.19}[4][magenta] \\

        &  \includegraphics[width=0.2\linewidth]{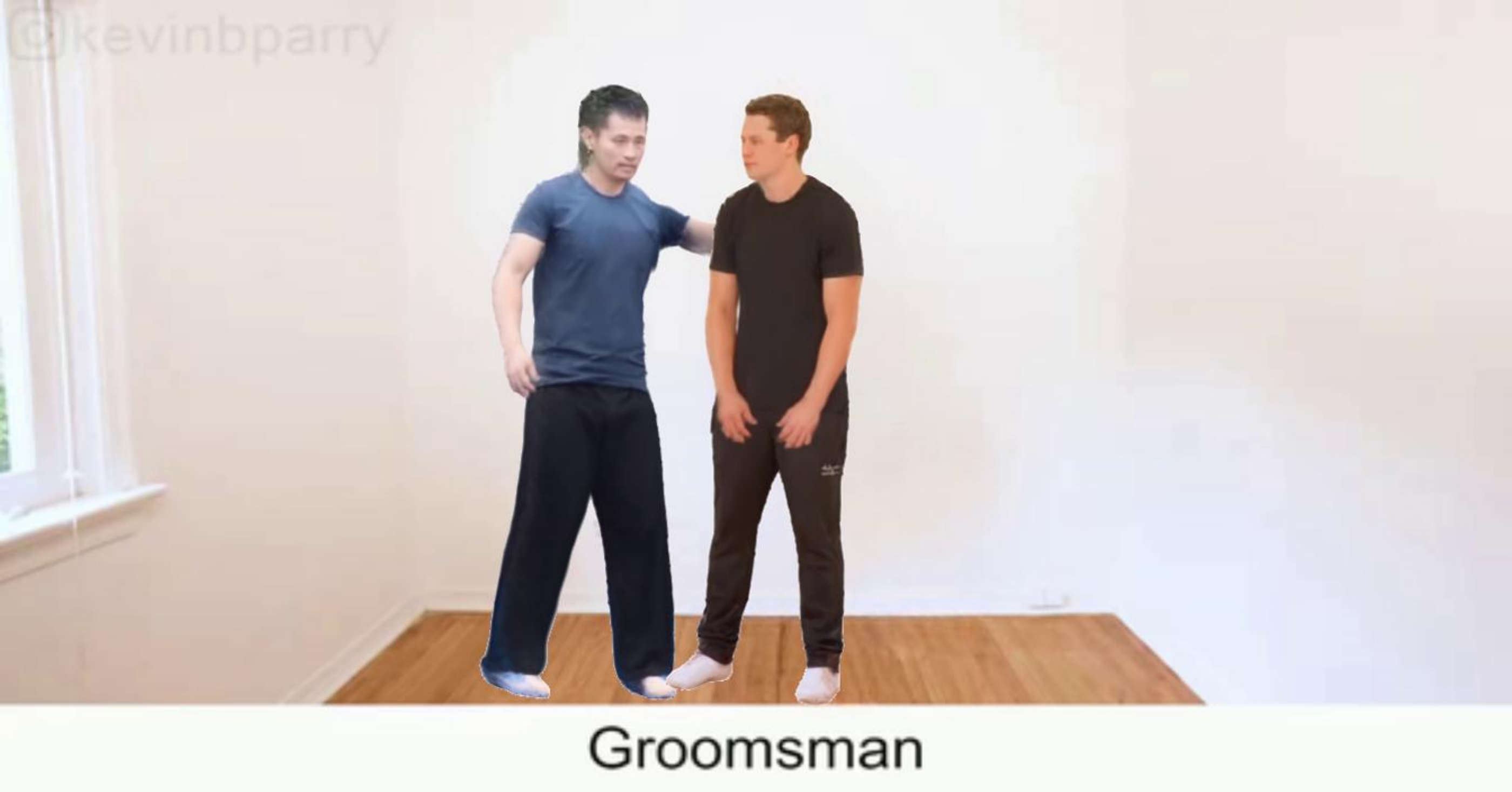} &

        \includegraphics[width=0.2\linewidth]{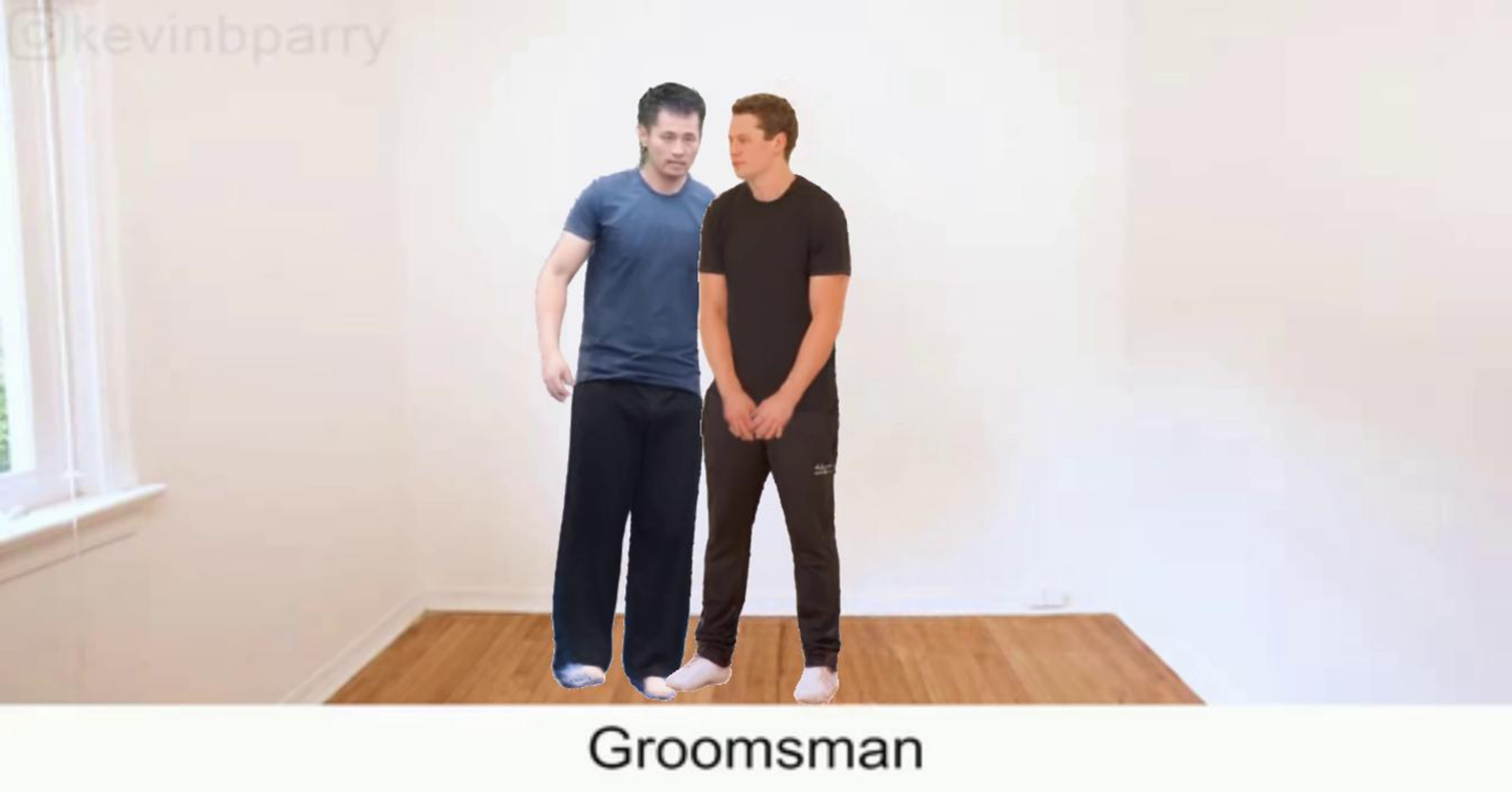} &

        \includegraphics[width=0.2\linewidth]{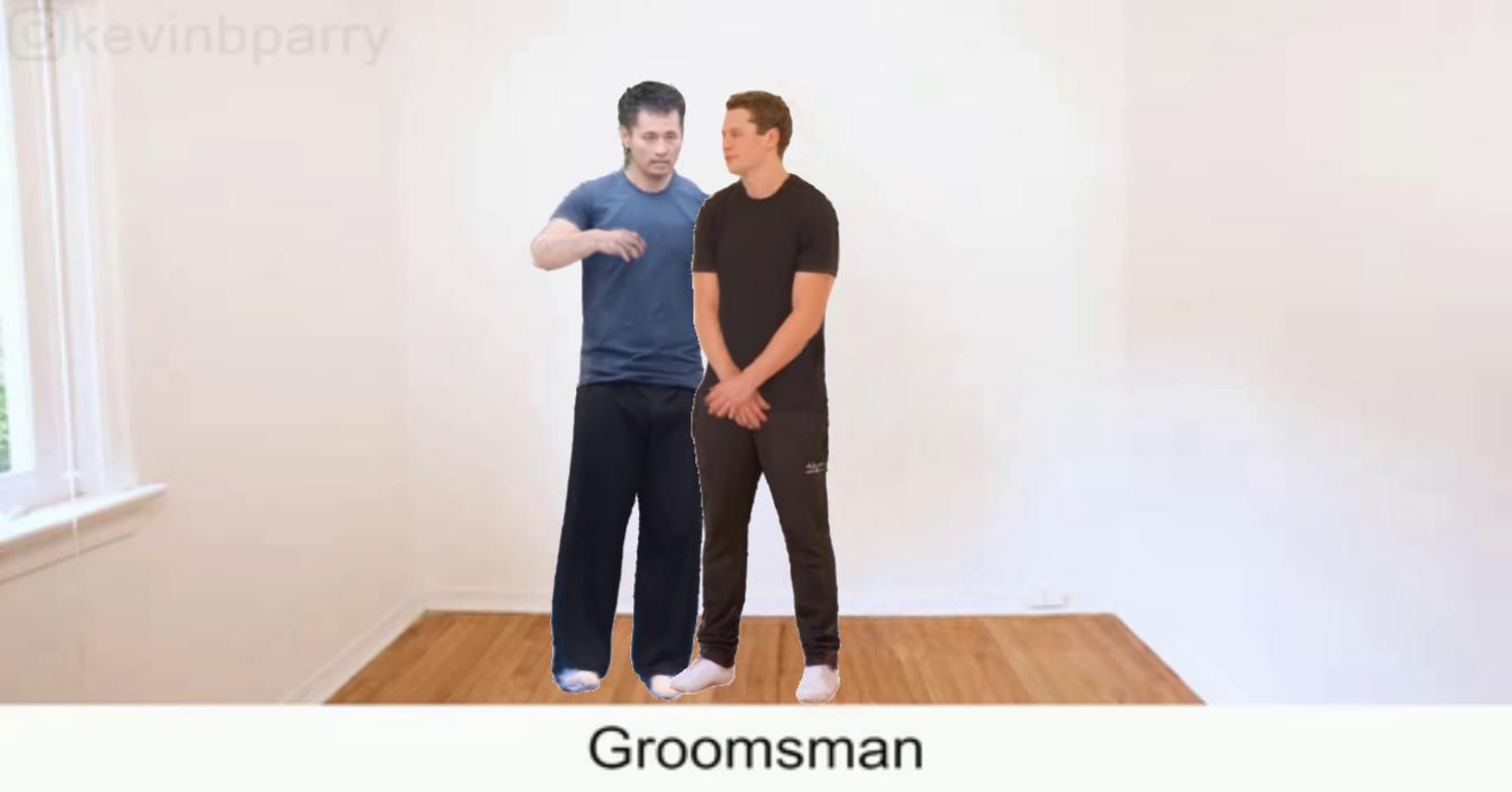} &

        \includegraphics[width=0.2\linewidth]{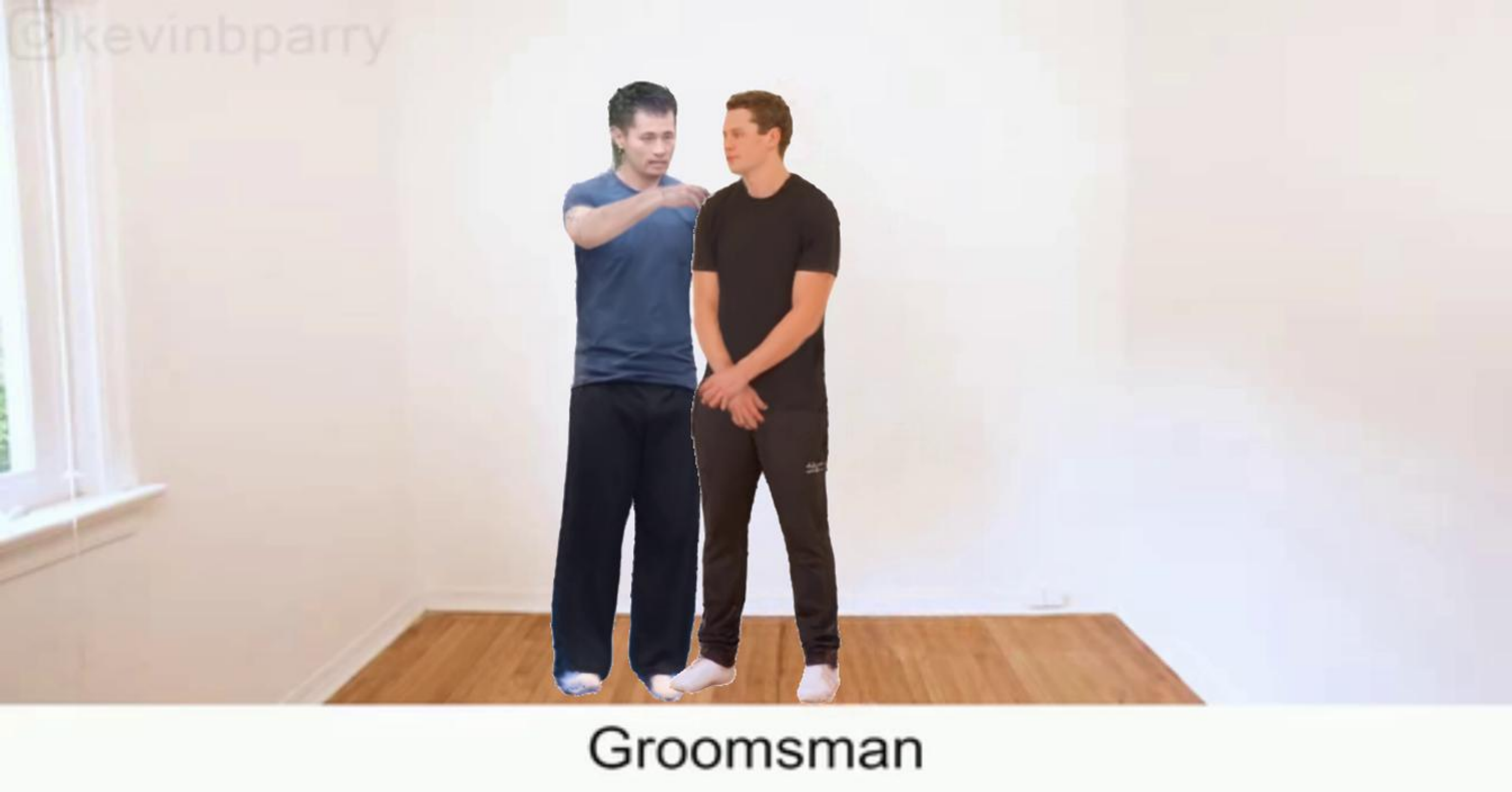} 
        
        \end{tabular}
    }
\captionof{figure}{\textbf{Interactive human video generation.} Given a single input image (left), our method generates interactive human motions that serve as intermediate results for video generation. We use an off-the-shelf human reconstruction model~\cite{qiu2025LHM} to recover textured humans from a single image. By pairing the generated motion with an arbitrary second person and applying the corresponding textures, we can produce realistic human interaction videos.}
\label{fig:video-gen}
\end{table*}

\noindent\textbf{Interaction synthesis from single pose}
We evaluate single pose-to-interaction synthesis on Inter-X~\cite{xu2024inter} dataset, comparing our method with an end-to-end without interactive pose baseline, which struggles in the large motion space, as shown in \cref{table:exp-singlepose} and \cref{fig:exp-pose-compare-results}. Our method leverages interactive poses to generate diverse motions under varying input conditions in \cref{fig:exp-p2m-diverse}.

\noindent\textbf{Open-world single-person image animation.} 
Our model generalizes to open-world single-person images by estimating poses~\cite{cai2023smplerx}, generating interactive counterparts, and animating motion. \cref{fig:exp-motionx-results} shows results on Motion-X~\cite{xu2024inter} dataset.

\subsection{Interaction Video Generation}
Our method generated interactive human motion could serve as intermediate outputs for downstream video generation. While existing video diffusion models~\cite{guo2023animatediff,ho2022video,blattmann2023align,jin2024pyramidal} can synthesize human videos, their motions often lack temporal consistency and realism. In contrast, our generated motions provide a stable and realistic foundation for interactive human video synthesis, either through pose-guided video diffusion models~\cite{xue2024follow,hu2024animate,zhu2024champ} or by texturizing motion sequences. As shown in \cref{fig:video-gen}, we use an off-the-shelf human reconstruction model~\cite{qiu2025LHM} to recover textured humans from a single image. The generated interactive motion is then paired with an arbitrary second person’s texture to produce realistic human interaction videos.

\subsection{Limitations}
Our method has few limitations: (1) it focus on short interaction segments; (2) it relies solely on human poses, ignoring scene context; (3) pose inaccuracies may cause contact errors and foot sliding; (4) close interactions may lead to inter-person penetration. Please refer to the ~\cref{sec:limitation} for more details.
 
\section{Conclusion}
\label{sec:conclusion}

We introduce Ponimator, which integrates a pose animator and generator for interactive pose animation and generation using conditional diffusion models. The animator leverages temporal priors for dynamic motion generation, the generator uses spatial priors to create interactive poses from a single pose, text, or both. Ponimator enables open-world image interaction animation, single-pose interaction generation, and text-to-interaction synthesis, exhibiting strong generalization and realism across datasets and applications.

{
    \small
    \bibliographystyle{ieeenat_fullname}
    \bibliography{arxiv}
}

\appendix
\clearpage
\maketitlesupplementary

\begin{abstract}
    
The supplementary material provides implementation details, limitation analysis, qualitative results and future work. In summary, we include

\begin{itemize}
    \item \cref{sec:implement}. Implementation details and model architecture of the interactive pose animator and generator. 
    \item \cref{sec:limitation}. Limitation analysis of our current approach.
    \item \cref{sec:qualitative}. Additional qualitative results of long interactive motion generation, complex interaction synthesis, two-person image animation, single-person image interaction generation, interactive pose animation, text-to-interaction motion synthesis, and single-pose-to-interaction motion synthesis.
    
\end{itemize}
\end{abstract}

\section{Implementation details}
\label{sec:implement}

\begin{table*}
    \centering
    \footnotesize
    \setlength{\tabcolsep}{0.2em} %
    \resizebox{\linewidth}{!}{
        \begin{tabular}{c|cccc}

        {Input} & 
        \multicolumn{4}{c}{{Interaction Animation (left$\rightarrow$right: time steps)}} \\

         \visvideo{limitation}{boys_31900}{05}{10}{15}{20}{0.16}[3]
        
         \visvideo{limitation}{Dancing_12847}{05}{10}{15}{20}{0.19}[4] 
         
          \visvideo{limitation}{get_water_2_clip1}{10}{15}{20}{25}{0.14}[3][magenta]
          
          \visvideo{limitation}{Ways_to_Jump_+_Sit_+_Fall_Temper_Tantrum_clip3}{10}{15}{20}{25}{0.19}[3][magenta] 
        
    \end{tabular}
    }

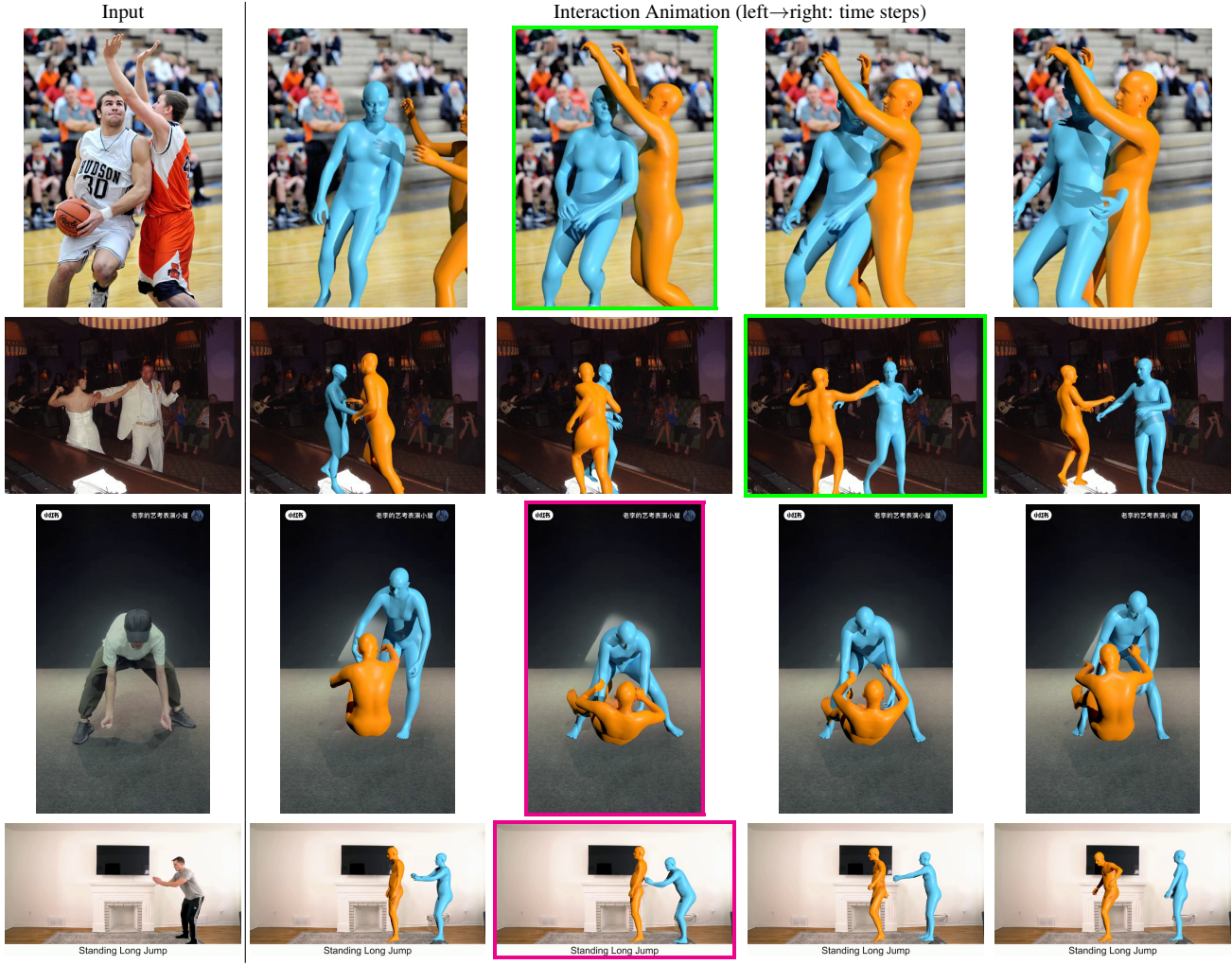
\captionof{figure}{\textbf{Method limitation analysis}. The first two rows show in-the-wild interactive pose animation results. In the first sample, severe interpenetration occurs as our method does not explicitly model penetration between two individuals. In the second, the generated motion is physically implausible due to the lack of scene context awareness, leading to collisions with the environment. The bottom two rows illustrate interaction motion generation from a single pose input. Due to inaccuracies in interactive pose generation, our method fails to produce realistic contact, resulting in unnatural motion.}
\label{fig:exp-limitation}
\end{table*}

\noindent\textbf{Interactive pose extraction.}
Given a two-person pose from a motion sequence, we determine close contact by measuring the minimum distance between their SMPL-X meshe vertices. Following~\cite{muller2024generative}, we downsample the mesh based on predefined contact regions and compute pairwise distances. If the smallest distance is below $1.3$cm, we classify the pose as a proximity pose—indicating contact between the individuals. This interactive pose is then used to train human interaction dynamics.

\noindent\textbf{Model architecture.}
Our pose animator and pose generator follow the DiT architecture~\cite{peebles2023scalable}, which consists of stacked Transformer blocks~\cite{vaswani2017attention}, each incorporating an attention mechanism and a feed-forward network (FFN). Both the animator and generator comprise 8 Transformer layers, with the animator utilizing both spatial- and temporal-attention blocks, while the generator employs only spatial attention. The model has a latent dimension of 1024, with 8-head multi-head attention, and uses the GELU activation function. The input motions are first encoded with positional encoding before being processed by Transformer blocks. The input has the shape $(B, P, N, D)$, where $B$ is batch size, $P = 2$ represents the number of individuals, and $N$ corresponds to number of frames, and $D$ is the dimension of diffusion target $\mathbf{z}_0$. Spatial attention operates along the $P$-dimension to model interactions between individuals, while temporal attention captures motion dynamics along the $T$-dimension. The model's output layer is a linear MLP, initialized with zero weights, which generates residual motion outputs. These residual motions are added to the interactive pose to produce the final output. Conditional information is incorporated into the model using Adaptive Instance Normalization~\cite{huang2017arbitrary}.

\noindent\textbf{Training.}
We apply training data augmentation to interactive poses in the interactive pose animator by adding random noise with a scale of $0.02$ to account for real-world inaccuracies in pose estimation. This ensures that even if the interactive pose estimator introduces noise, the animator can still produce reasonable results. This augmentation is performed online during training. Following prior work~\cite{guo2022generating, liang2024intergen}, we align one person's pose in the interactive pose to face the positive Z direction and center it at the origin. The interaction loss in the pose animator follows~\cite{liang2024intergen} and consists of a **contact loss**, which encourages contact between two individuals when their joints are close, and a **relative orientation loss**, which aligns their global orientations with the ground truth. The velocity loss $\mathcal{L}_{\text{vel}}$, following MDM~\cite{tevet2023human}, ensures motion coherence by minimizing the velocity difference between the generated motion and the ground truth. For diffusion training, we use a cosine scheduler with 1000 diffusion steps and DDIM sampling~\cite{song2020denoising} for 50 steps during inference. The model is trained with a learning rate of $1e\text{-}4$ and weight decay of $0.00002$ for 4000 epochs. The batch size is 256 for the interactive pose animator and 512 for the interactive pose generator. Training takes 2 days for the pose animator and 1 day for the pose generator on 4×A100 GPUs. 

\noindent\textbf{Inference speed comparison.}
Our interactive pose generation takes $0.21$s on a single A100 on average, the interactive pose animator generates $3$s motion at 10fps in $0.24$s, comparable to InterGen~\cite{liang2024intergen} which requires 0.76s for the same motion length.

\section{Limitation Analysis}
\label{sec:limitation}

Our method has the limitations below. The common failure modes are illustrated in \cref{fig:exp-limitation}. 

\noindent\textbf{Short motion modeling.}
Our method is mainly focus on short interactive motion segments. While our framework could support longer generation by interactive pose chaining as shown in \cref{fig:exp-c2m-long}, the benefit of interactive pose prior would diminish over time. In text-to-interaction synthesis, our framework prioritizes interactive motion-relevant information, which can result in partial rather than complete motion sequences when the input text describes extended human interactions. Moreover, our pose animator—taking only interactive poses as input—cannot fully capture the semantic context or temporal ordering in text (e.g., distinguishing “lifting up” from “putting down”). Incorporating text conditioning into the pose-to-interaction stage is a promising avenue for improving text-to-interaction–specific tasks. However, since our main focus is on pose-to-interaction animation without enforced text input, this ambiguity can be a strength, enabling multiple valid and physically plausible motion interpretations from the same interactive pose.

\begin{table*}[t]
    \centering
    \footnotesize
    \setlength{\tabcolsep}{0.2em} %
    \resizebox{\linewidth}{!}{
        \begin{tabular}{ccccccccc}
    
    \includegraphics[width=0.2\linewidth]{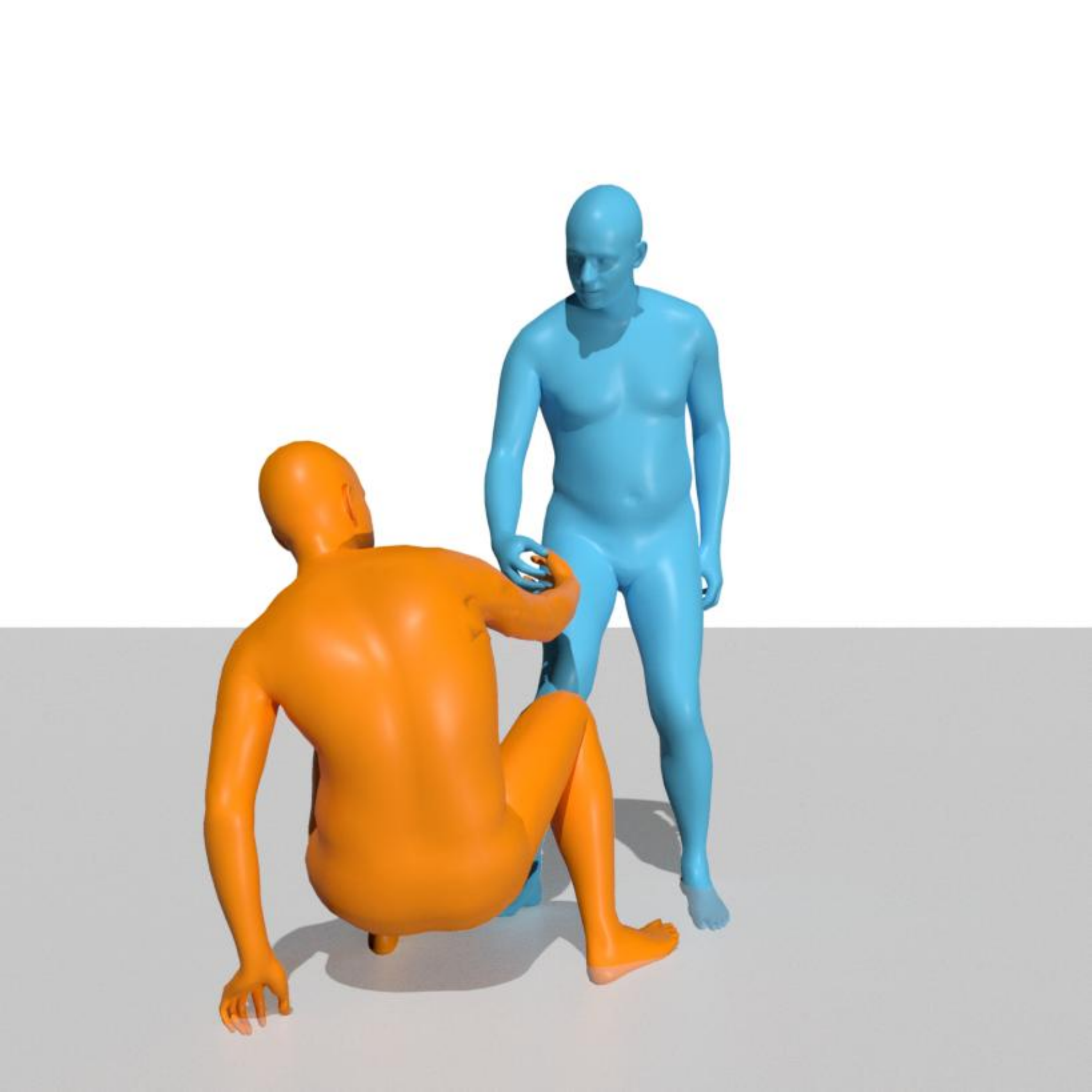} &
    \fcolorbox{magenta}{white}{\includegraphics[width=0.2\linewidth]{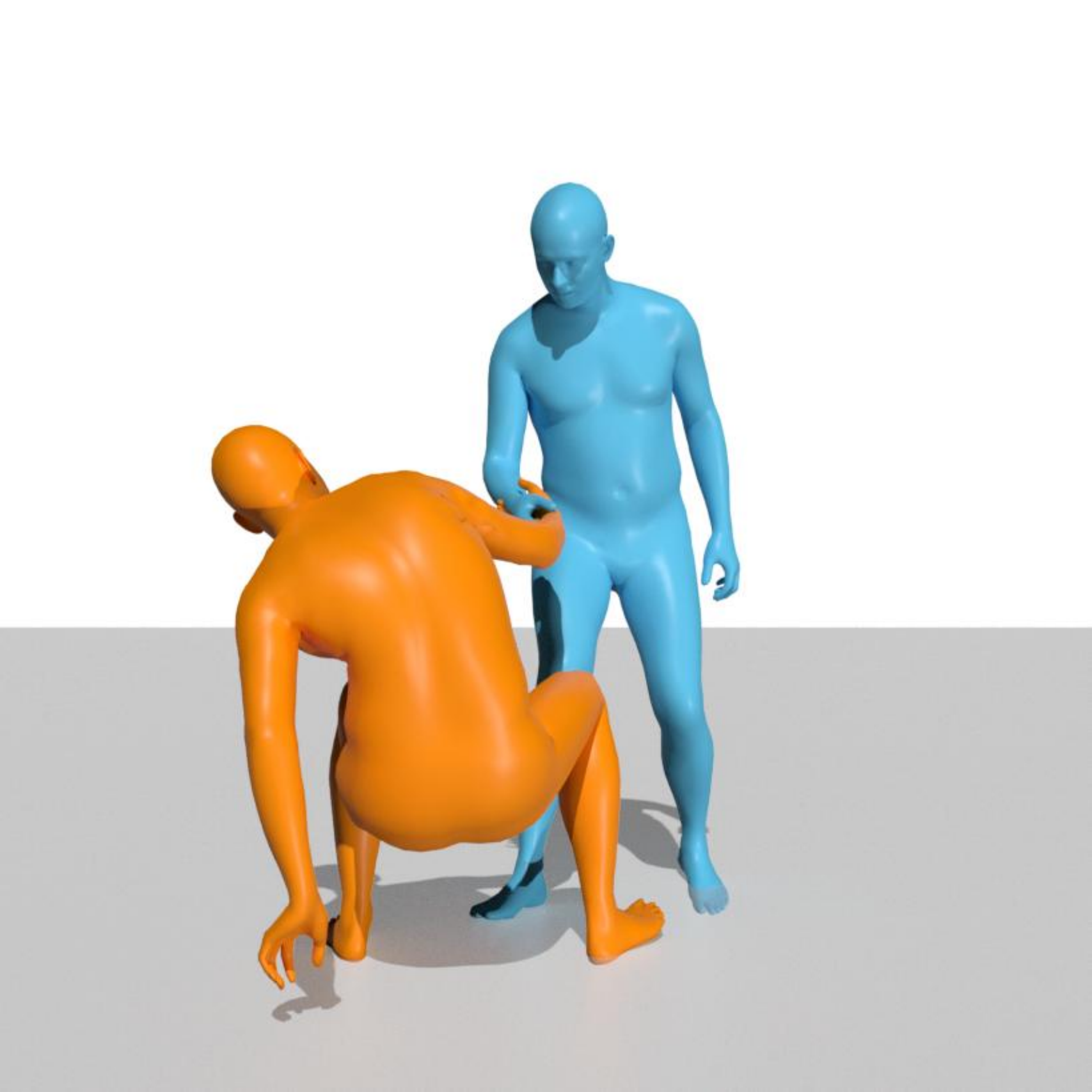}} &
    \includegraphics[width=0.2\linewidth]{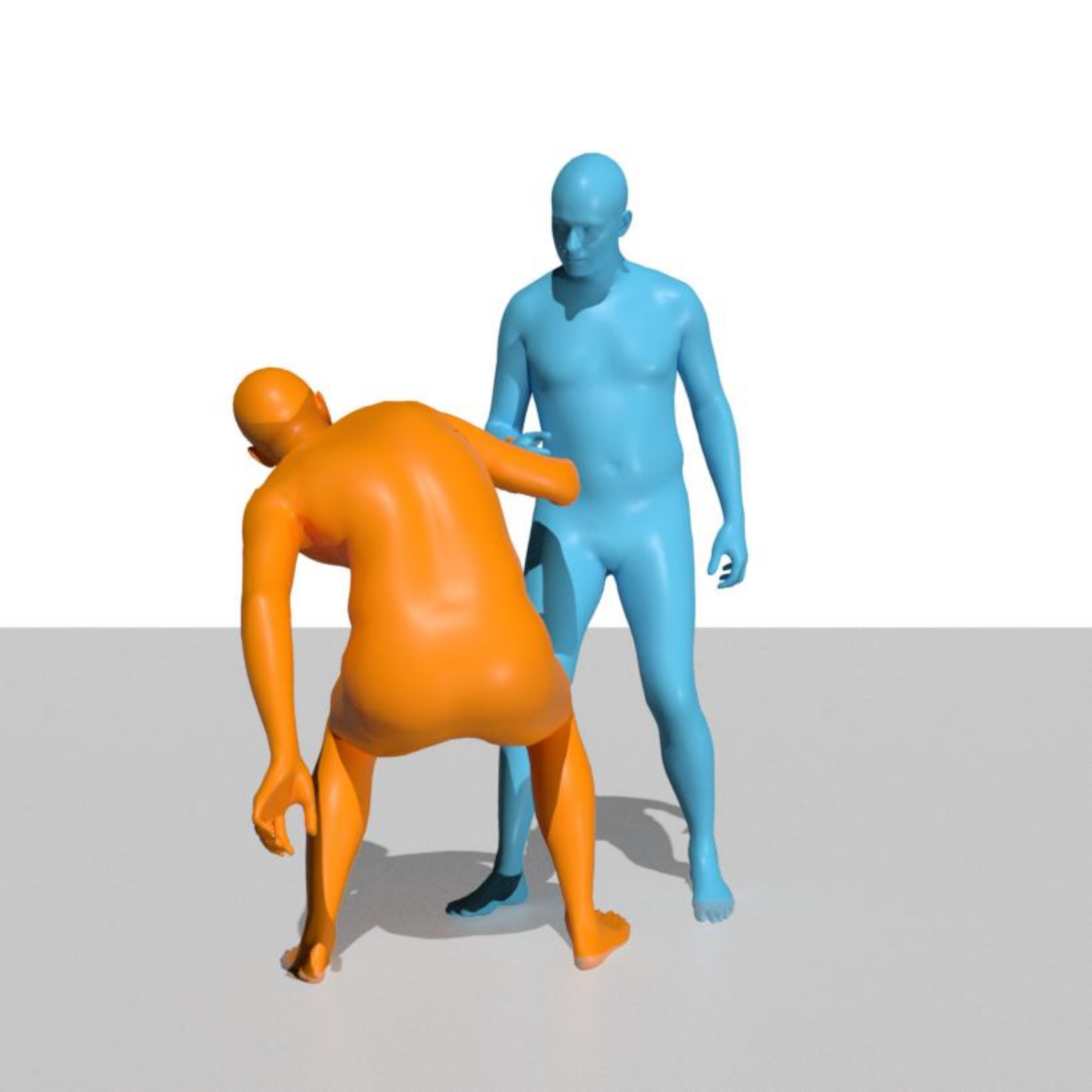} &
    \includegraphics[width=0.2\linewidth]{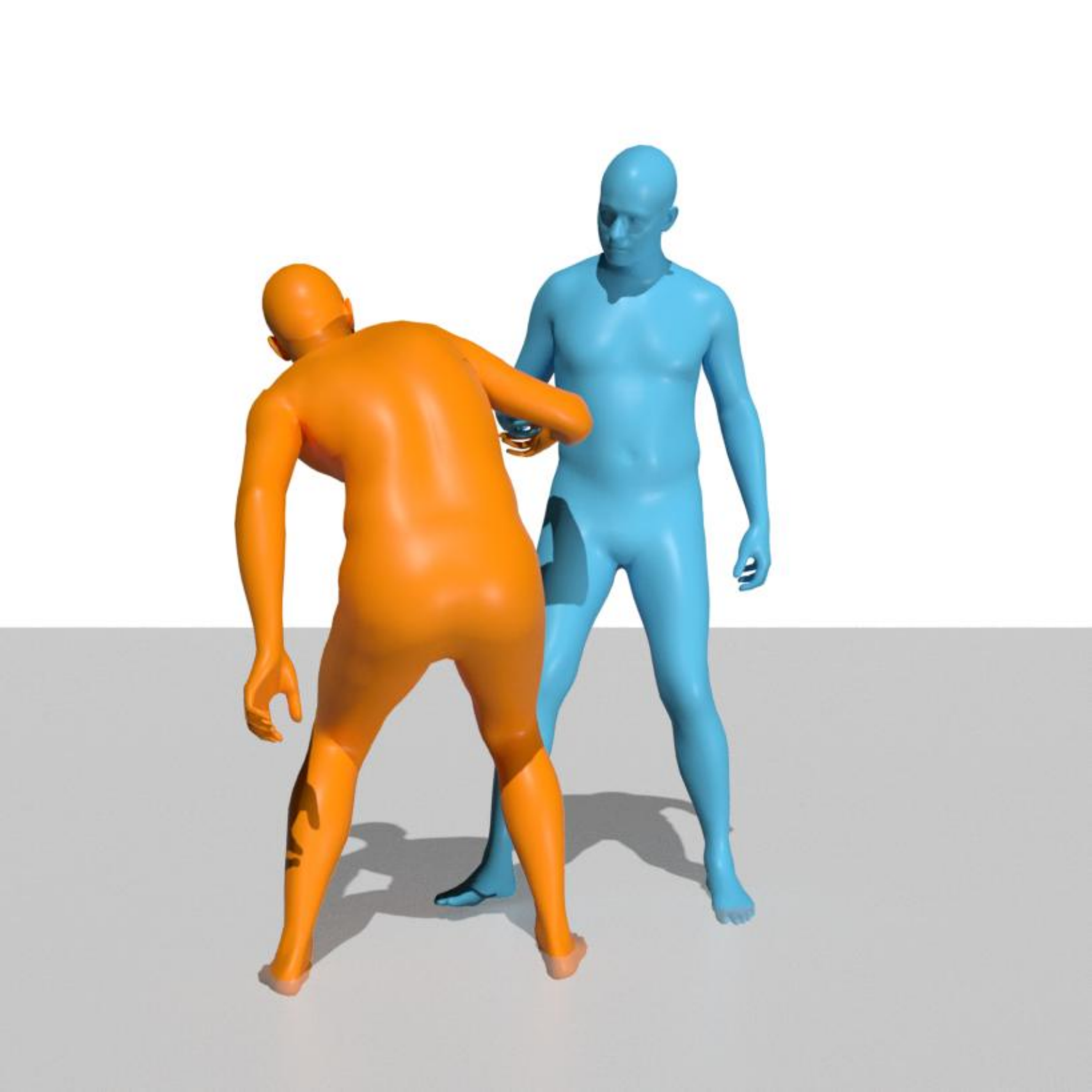} &

    & \fcolorbox{magenta}{white}{\includegraphics[width=0.2\linewidth]{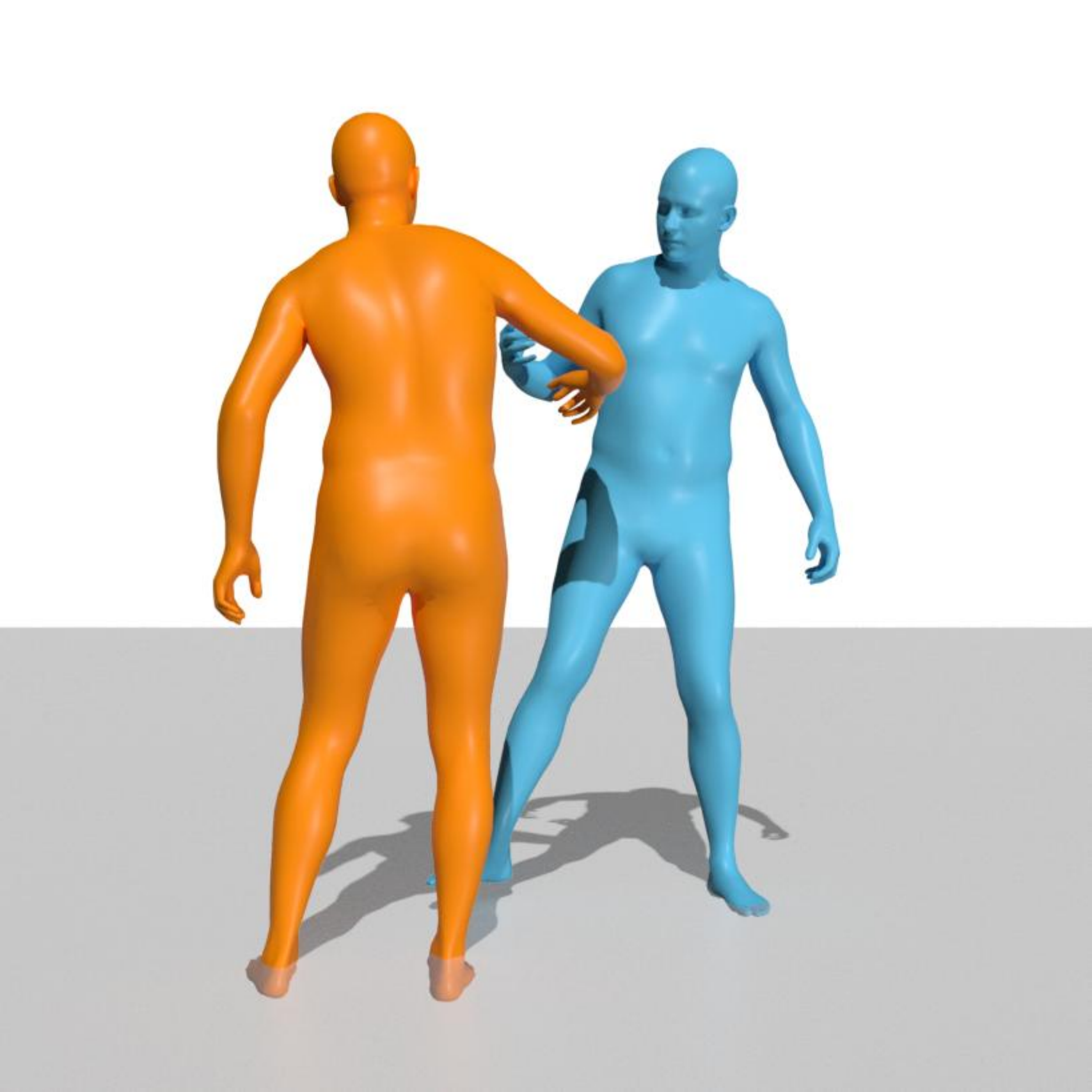}} &
    \includegraphics[width=0.2\linewidth]{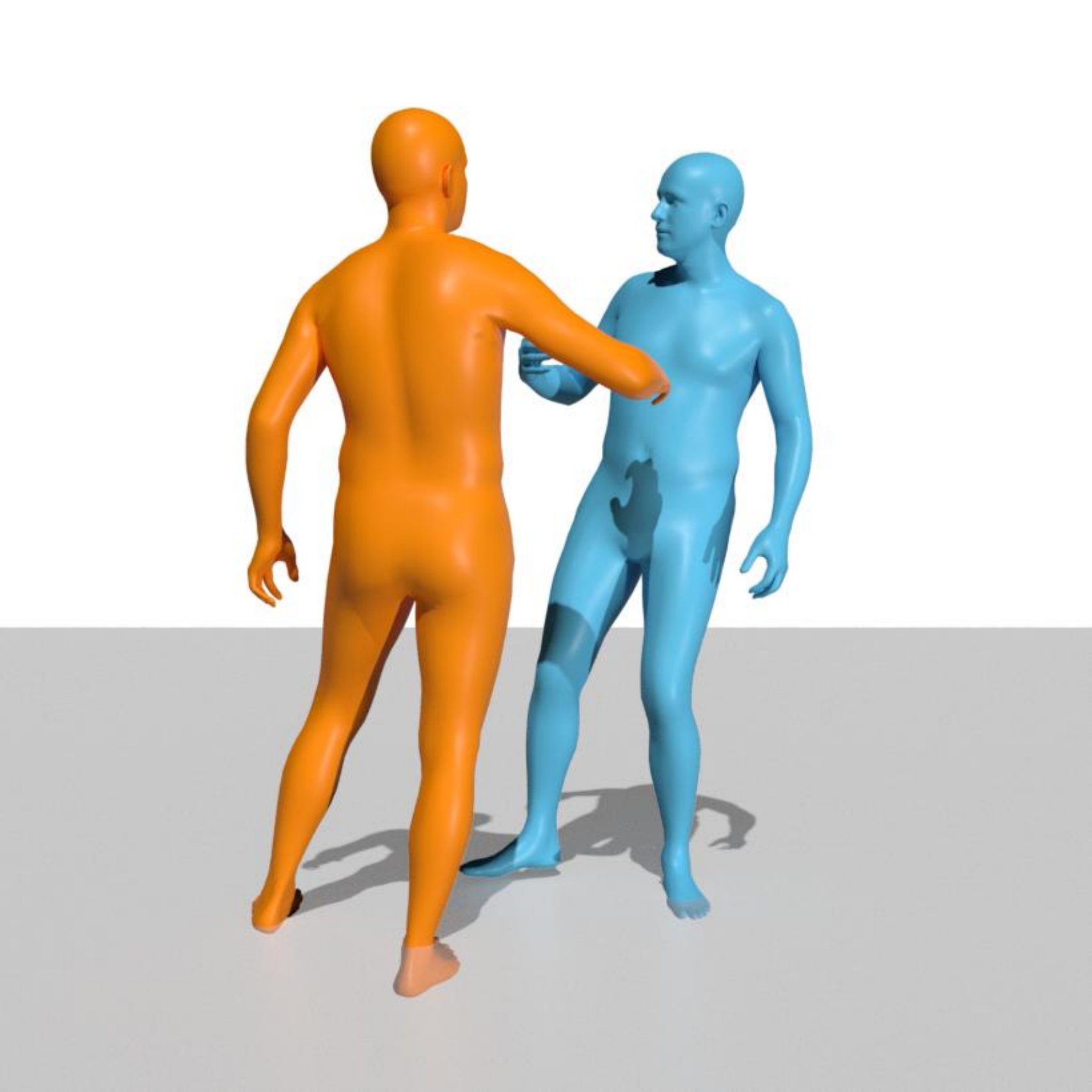} &
    \fcolorbox{magenta}{white}{\includegraphics[width=0.2\linewidth]{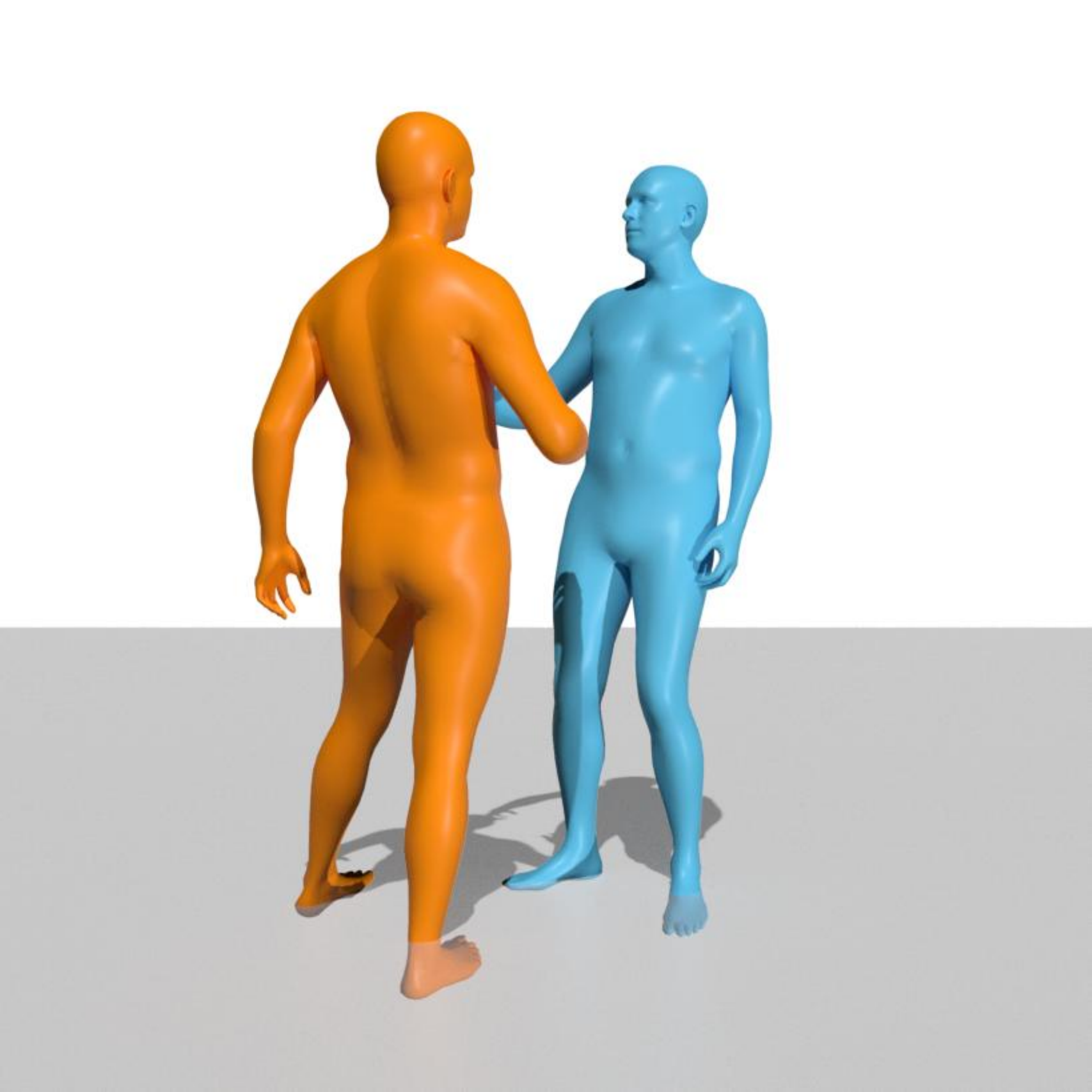}} &
    \includegraphics[width=0.2\linewidth]{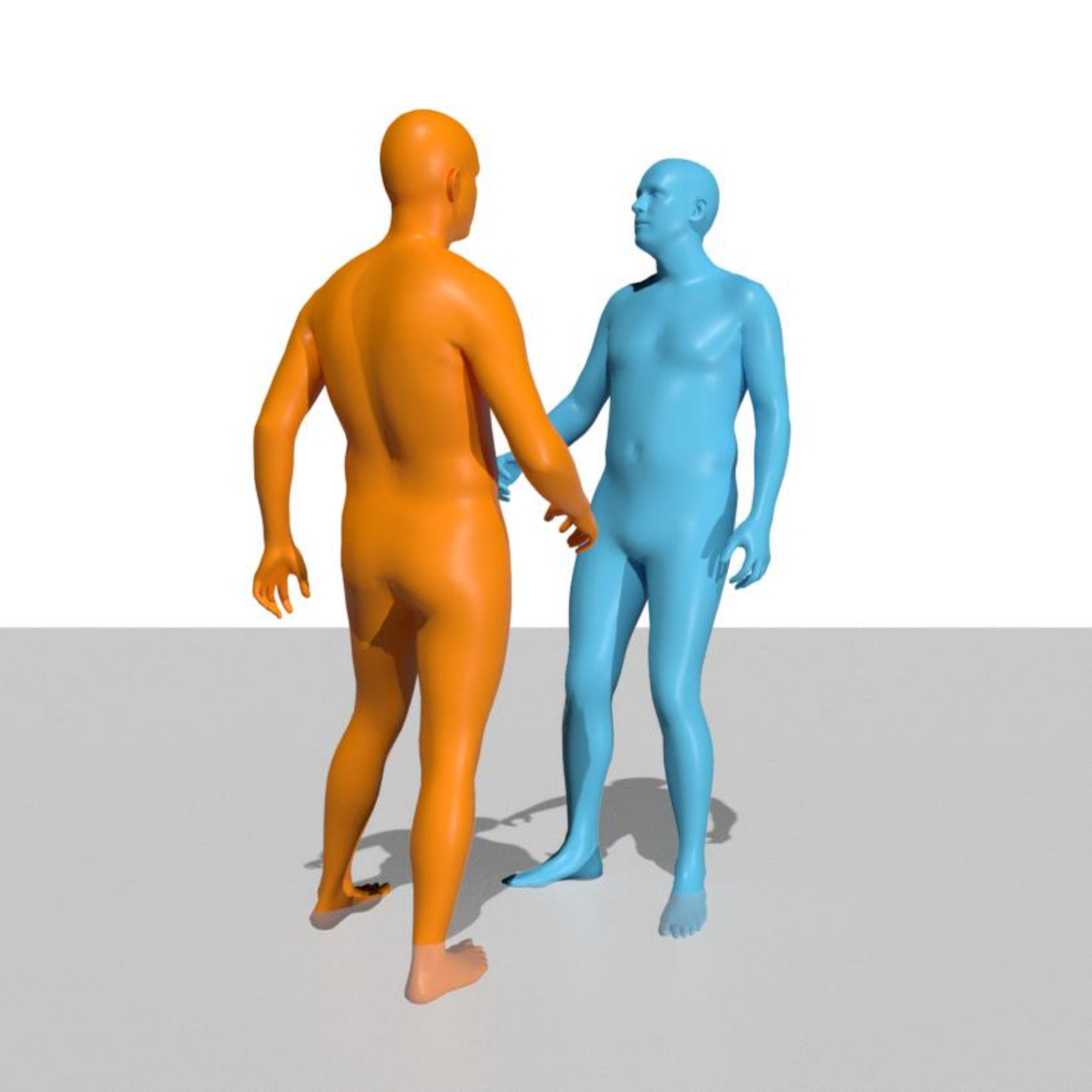} \\

        \end{tabular}
    }
\captionof{figure}{\textbf{Longer motion generation} by chaining interactive poses. We reuse the last generated pose as the next input, resetting interactive time to zero, enabling sliding-window synthesis of longer motions (key-frame in \textcolor{magenta}{magenta box}).}

\label{fig:exp-c2m-long}
\end{table*}

\begin{table*}
    \centering
    \footnotesize
    \setlength{\tabcolsep}{0.2em} %
    \resizebox{\linewidth}{!}{
        \begin{tabular}{c|ccccc}

        {Interactive Pose} & 
        \multicolumn{5}{c}{{Interaction Animation (left$\rightarrow$right: time steps)}} \\

    \includegraphics[width=0.2\linewidth]{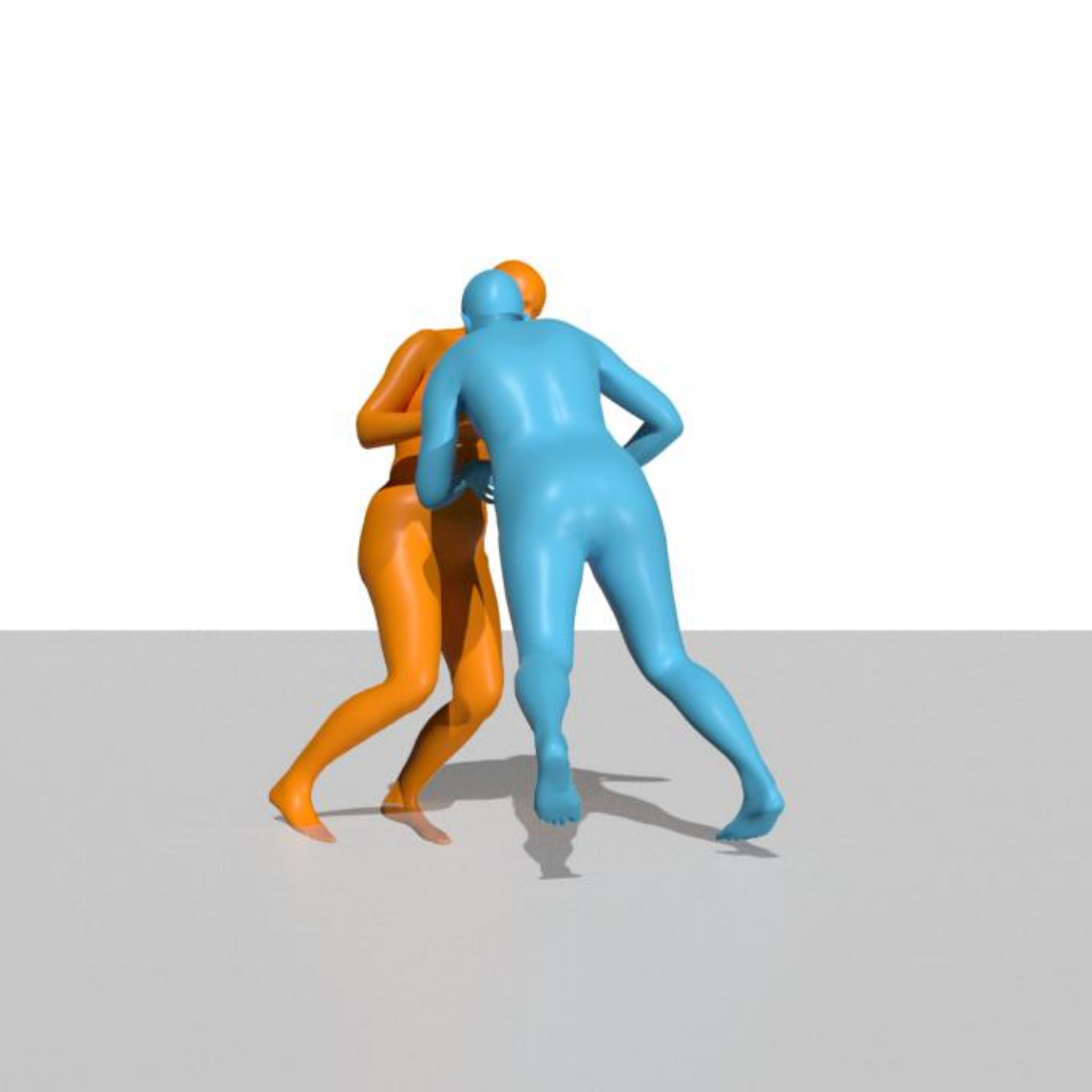} &
    \includegraphics[width=0.2\linewidth]{src_figs/vis_contact/vis_complex/Tai_Chi_Kungfu_25_clip1/00_crop.pdf} &
    \includegraphics[width=0.2\linewidth]{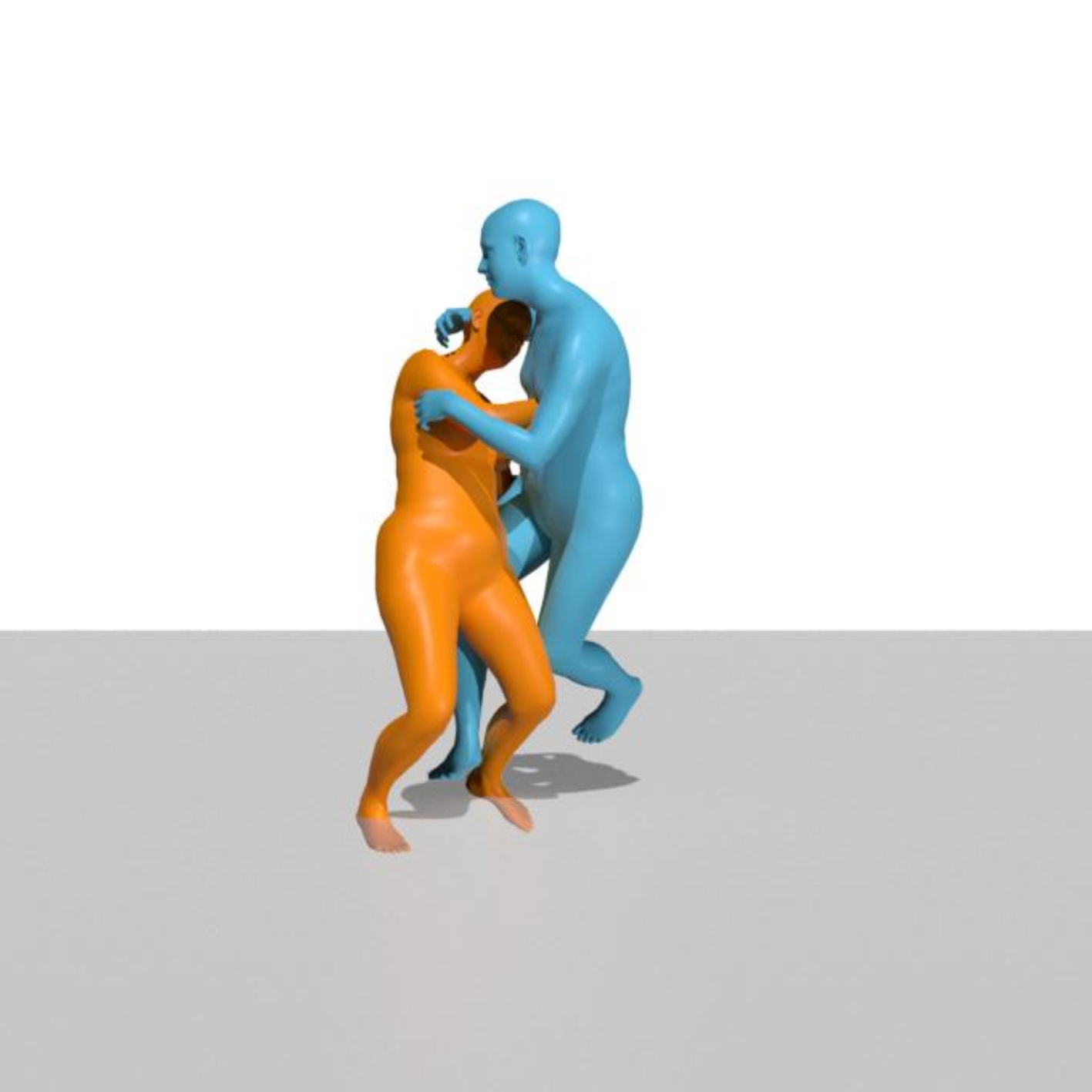} &
    \includegraphics[width=0.2\linewidth]{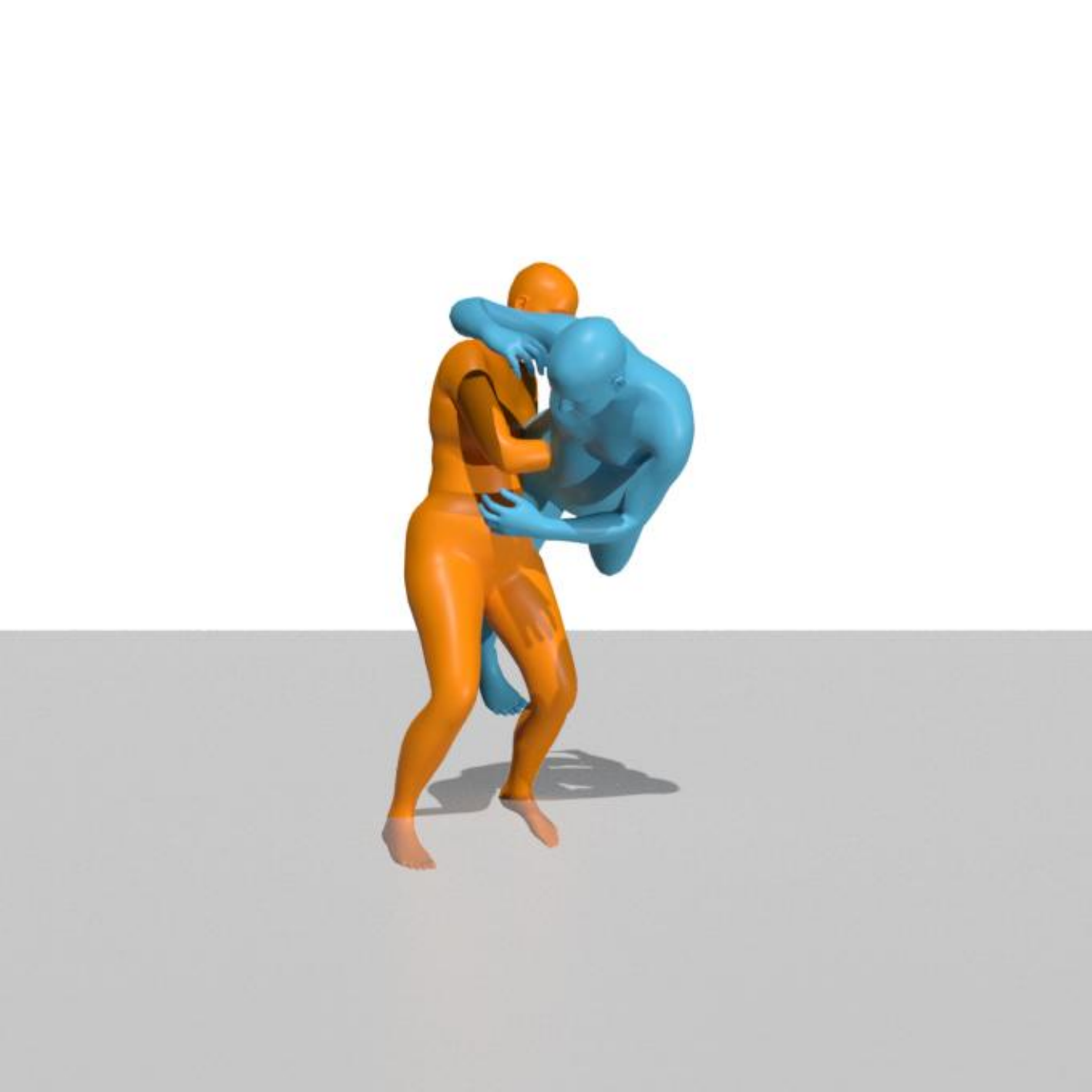} &
    \includegraphics[width=0.2\linewidth]{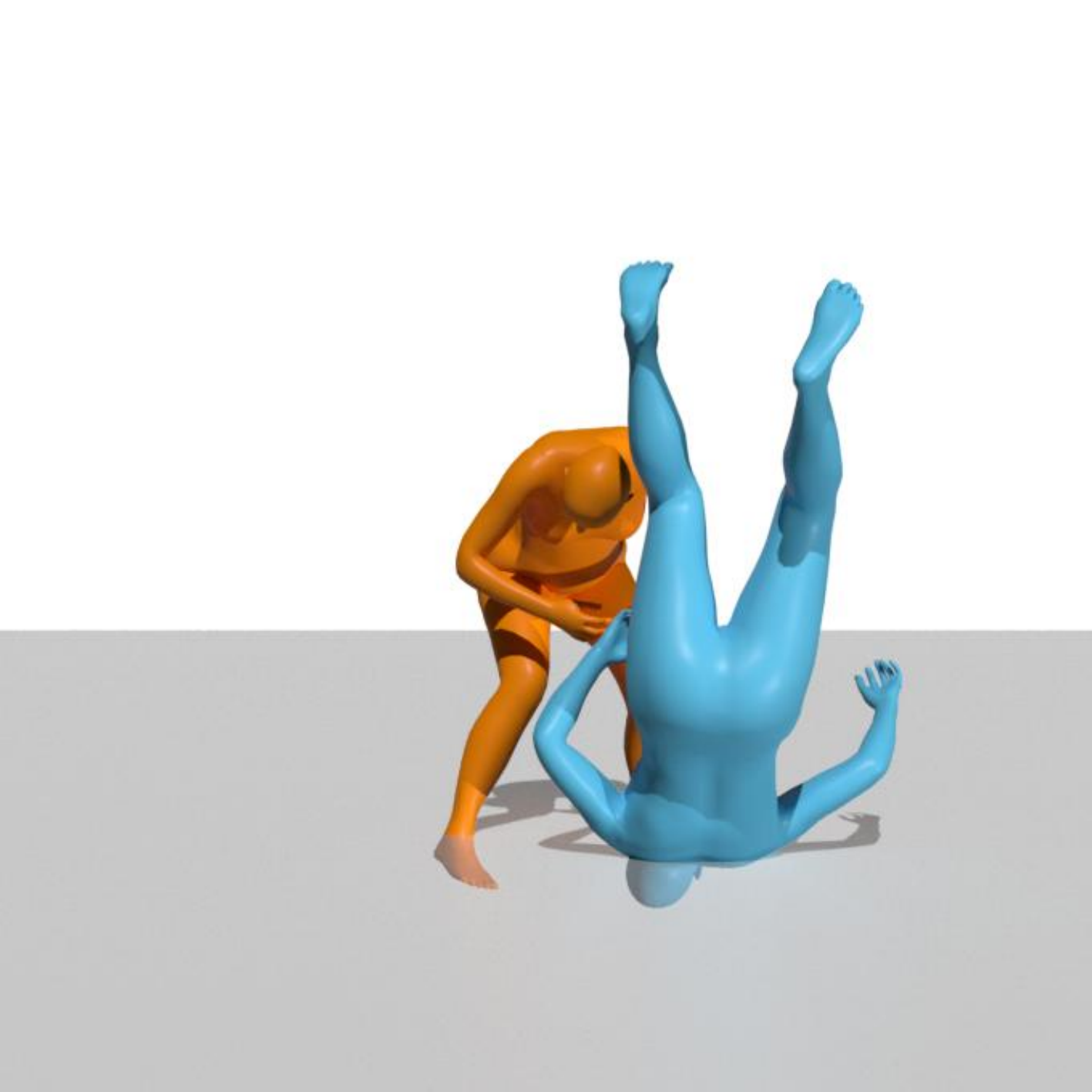} &
    \includegraphics[width=0.2\linewidth]{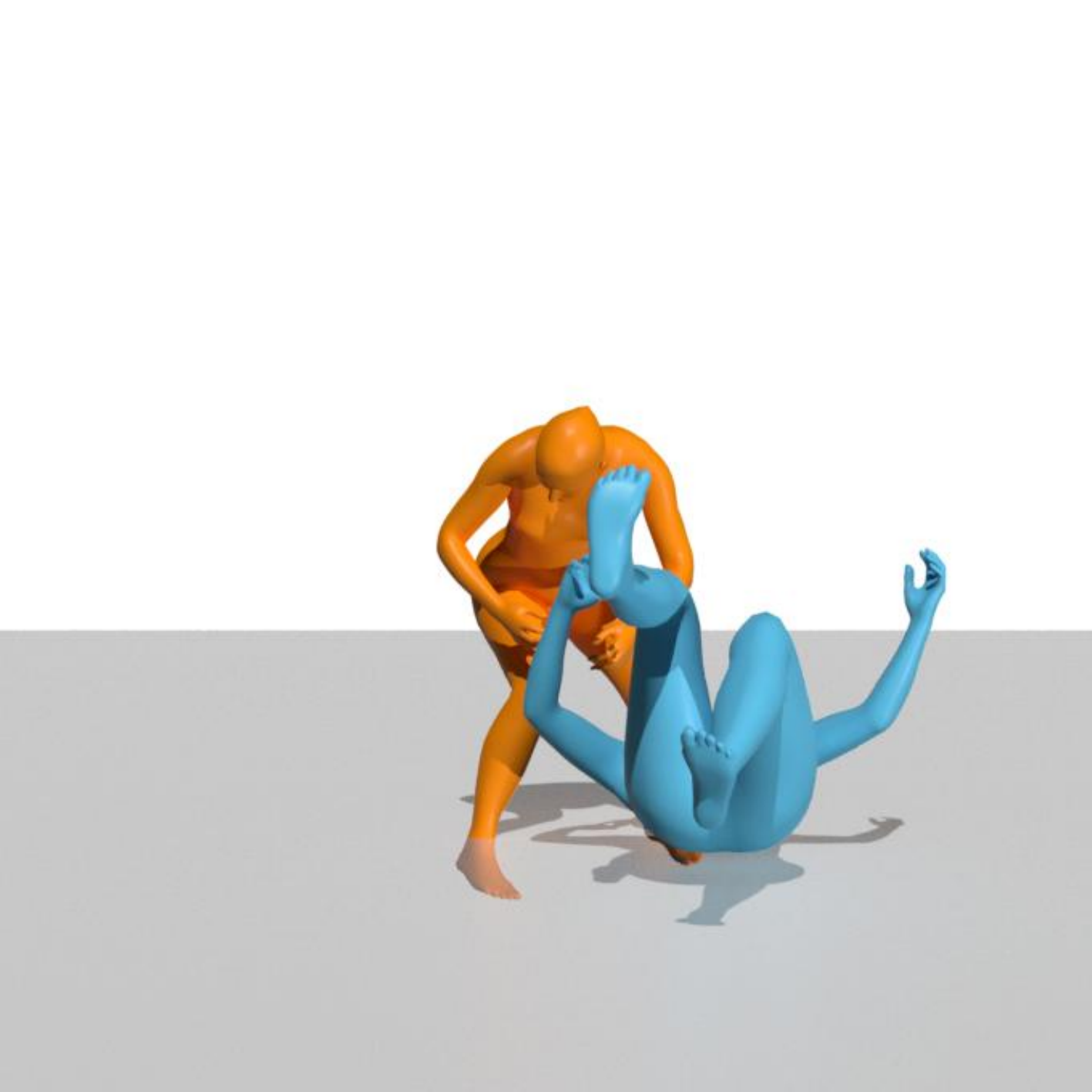} \\
    
    \includegraphics[width=0.2\linewidth]{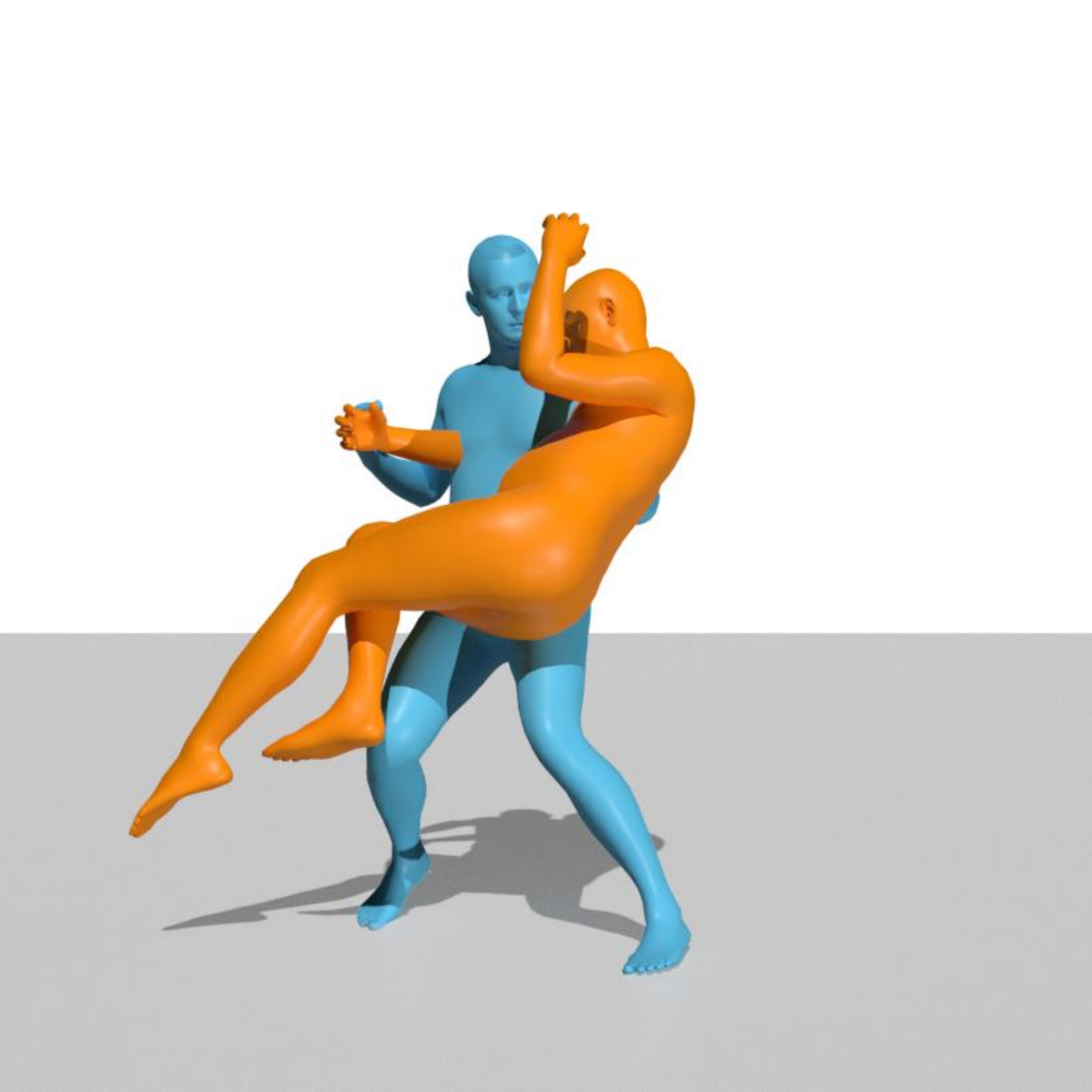} &
    \includegraphics[width=0.2\linewidth]{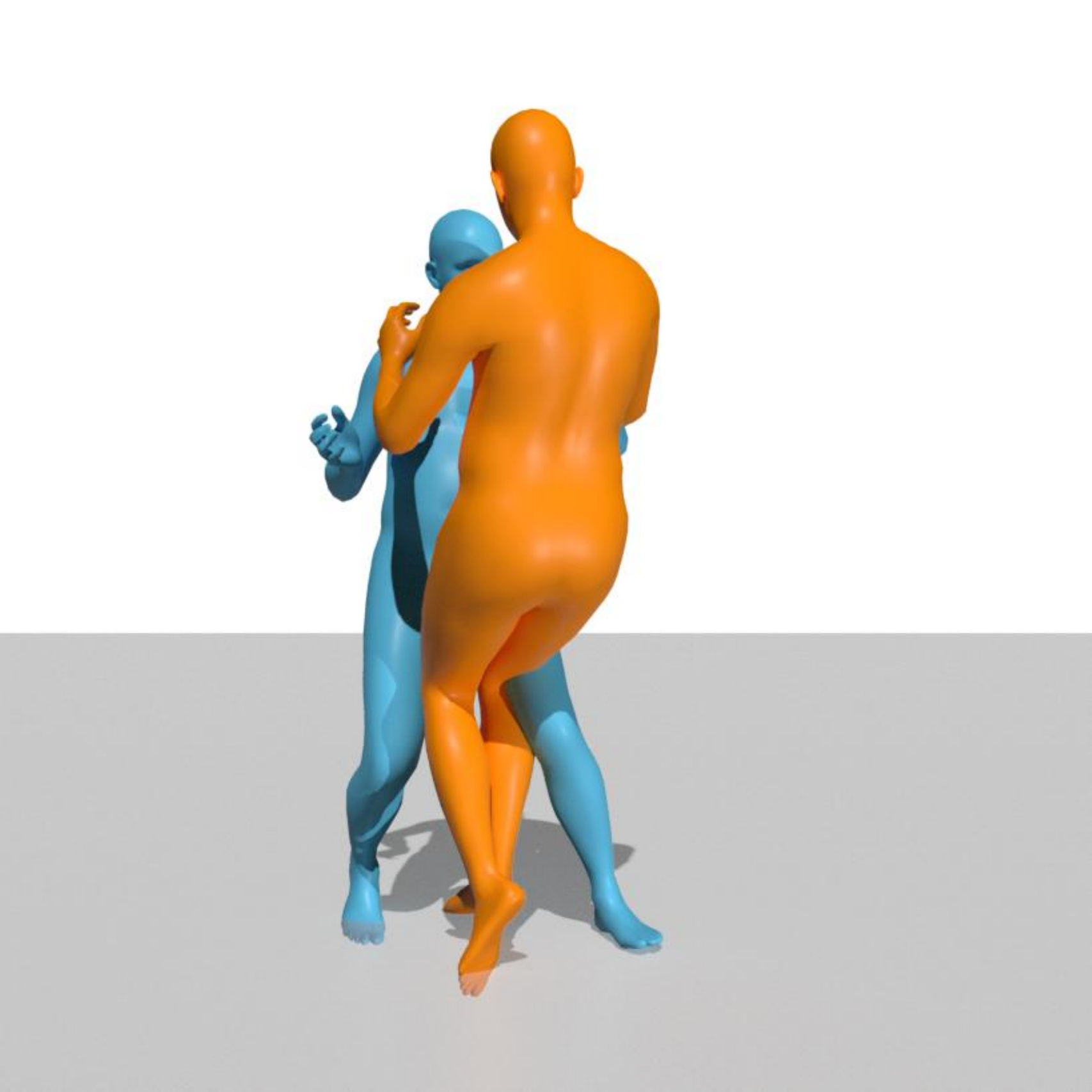} &
    \includegraphics[width=0.2\linewidth]{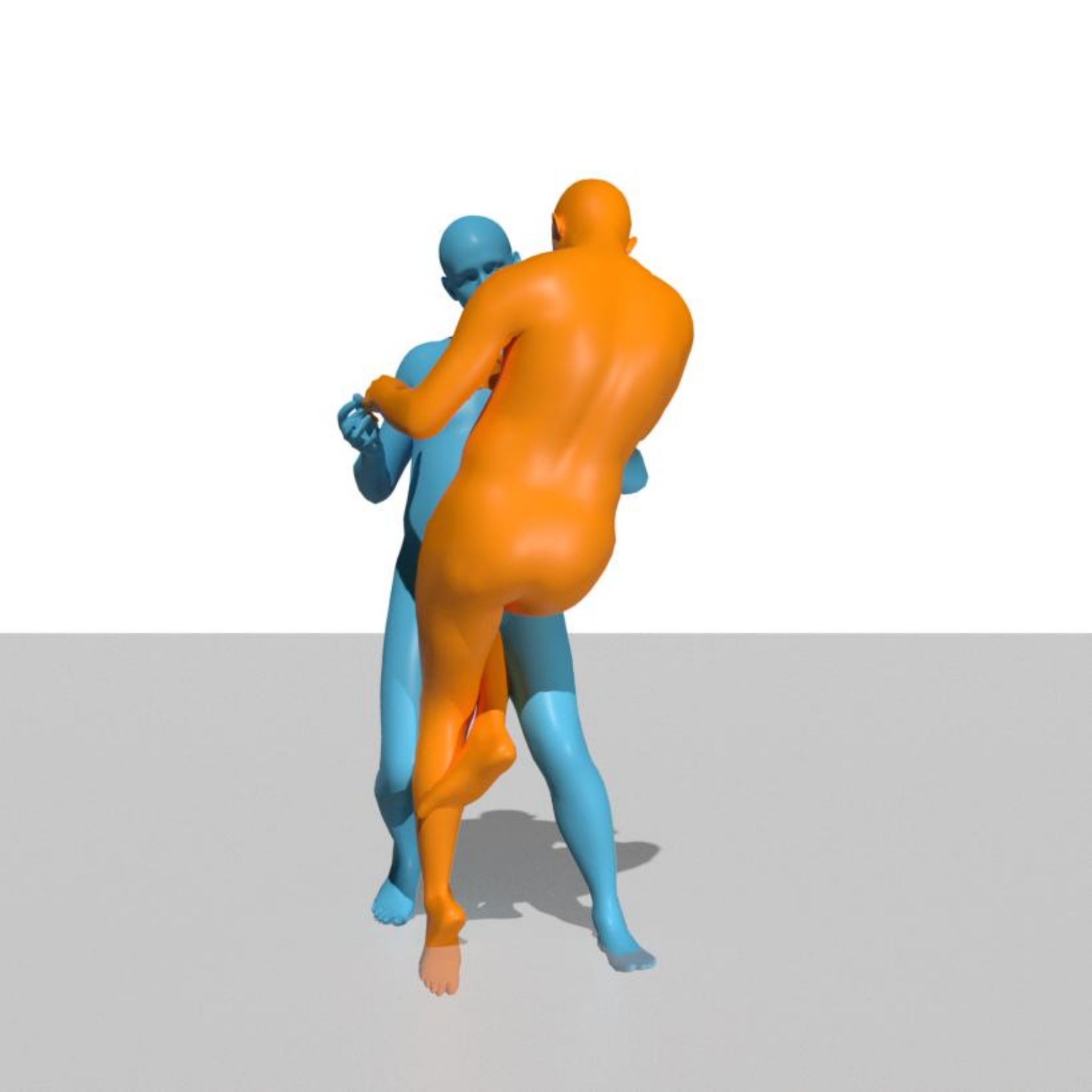} &
    \includegraphics[width=0.2\linewidth]{src_figs/vis_contact/vis_complex/pair00_000152_1_15/10_crop.pdf} &
    \includegraphics[width=0.2\linewidth]{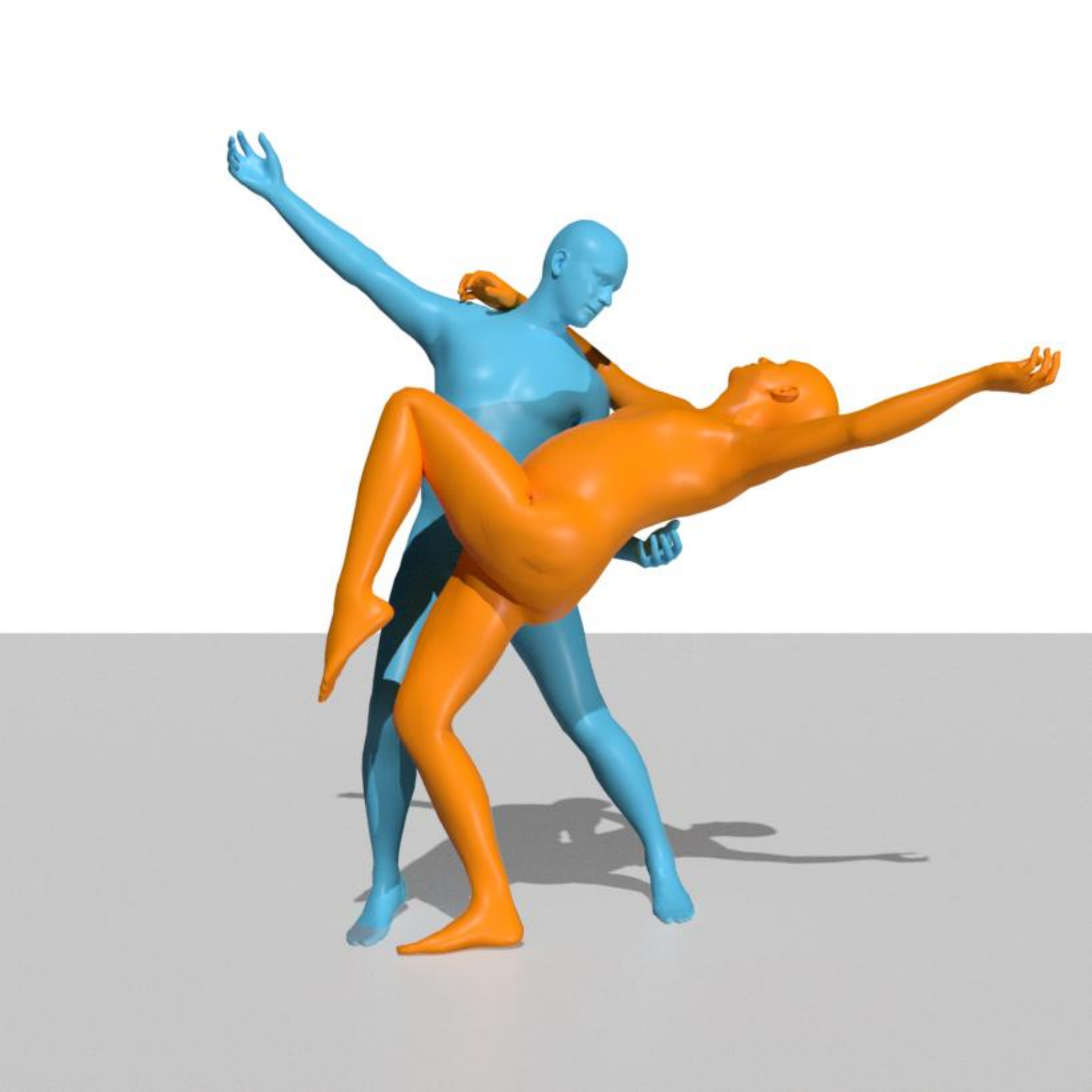} &
    \includegraphics[width=0.2\linewidth]{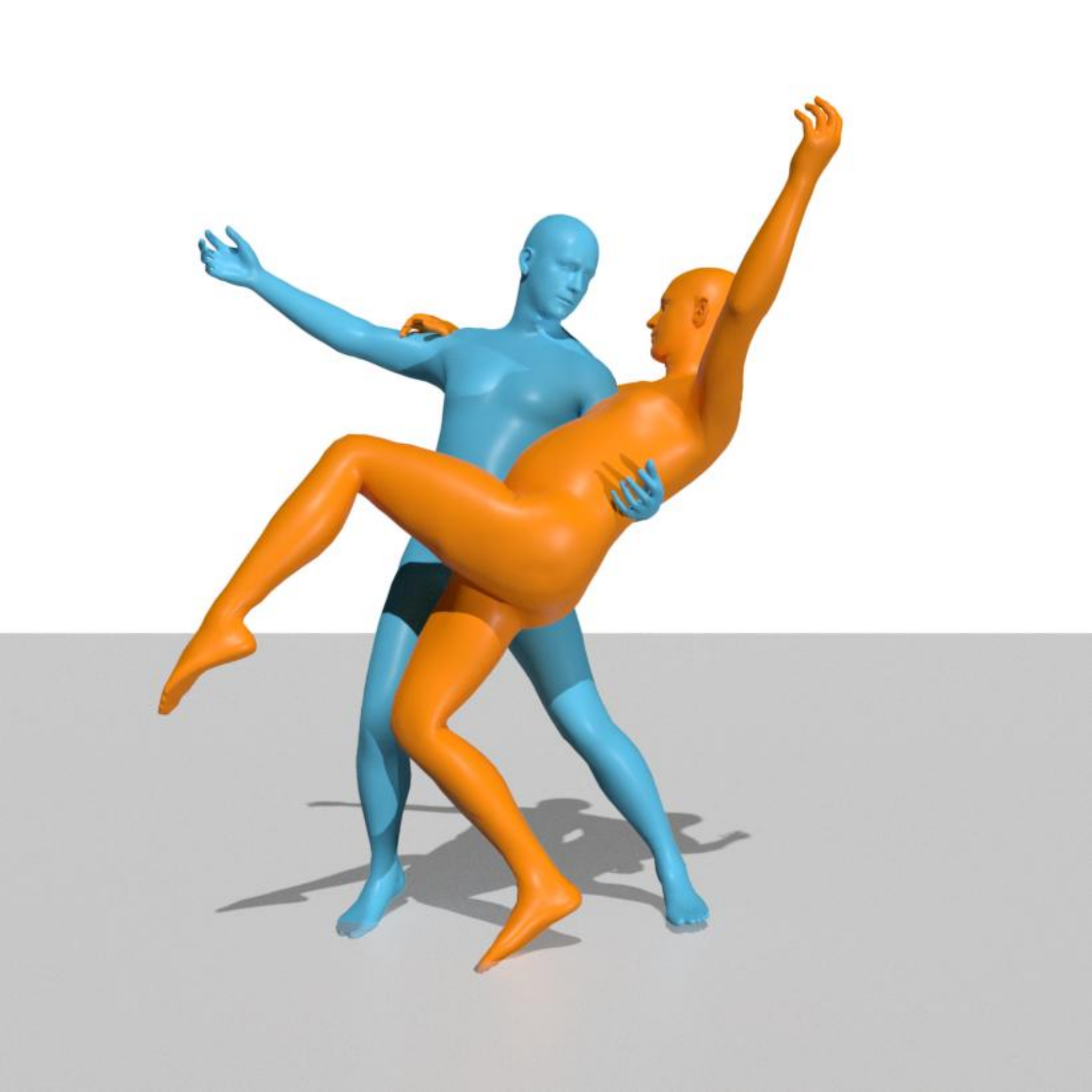} \\

    \end{tabular}
    }
\captionof{figure}{\textbf{Complex interactive pose animation}. Given an interactive pose, our pose animator can synthesize high-dynamics (1st row) and close-contact (2nd row) human-human motions, leveraging the strong interactive prior learned from high-quality mocap data.}
\label{fig:exp-contact-complex}
\vspace{-1em}
\end{table*}
\begin{table*}
    \centering
    \footnotesize
    \setlength{\tabcolsep}{0.2em} %
    \resizebox{\linewidth}{!}{
        \begin{tabular}{c|cccc}

        {Input} & 
        \multicolumn{4}{c}{{Interaction Animation (left$\rightarrow$right: time steps)}} \\

        \visvideo{vis_buddi}{Judo_2113}{00}{10}{15}{20}{0.19}[4] 
        
         \visvideo{vis_buddi}{Tango_796}{10}{15}{20}{25}{0.19}[3]
        
         \visvideo{vis_buddi}{938}{05}{10}{15}{25}{0.17}[4]
        
         \visvideo{vis_buddi}{Wrestling_1707}{05}{10}{15}{20}{0.19}[4]
       
         \visvideo{vis_buddi}{751}{05}{10}{15}{20}{0.16}[4]
        
         \visvideo{vis_buddi}{happy_56345}{05}{10}{15}{20}{0.18}[4]
        
        \end{tabular}
    }
\captionof{figure}{\textbf{Interactive pose image animation} on FlickrCI3D dataset~\cite{fieraru2020three}. Left shows the input image, right shows the animated interaction motions. Interactive-pose frame is labeled in \textcolor{green}{green box}. Our model generalizes well to in-the-wild interactive poses, producing realistic human-human interaction dynamics.}
\label{fig:supp-buddi-results}
\end{table*}
\begin{table*}
    \centering
    \footnotesize
    \setlength{\tabcolsep}{0.2em} %
    \resizebox{\linewidth}{!}{
        \begin{tabular}{c|cccc}

        {Input} & 
        \multicolumn{4}{c}{{Interaction Animation (left$\rightarrow$right: time steps)}} \\

        \visvideo{vis_motionx}{Ways_to_Catch_Twisted_Ankle_clip1}{00}{05}{10}{20}{0.19}[5][magenta]
        
         \visvideo{vis_motionx}{Aerial_Kick_Kungfu_wushu_15_clip2}{05}{10}{15}{20}{0.19}[4][magenta]

        \visvideo{vis_motionx}{tense_moment_clip5}{05}{10}{15}{20}{0.19}[4][magenta]
        
        \visvideo{vis_motionx}{dance_13_clip1}{05}{10}{15}{20}{0.19}[4][magenta]
        
         \visvideo{vis_motionx}{Back_Flip_Kungfu_wushu_Trim9_clip1}{05}{10}{15}{20}{0.19}[4][magenta]

         One person pushes the other \\

    \end{tabular}
    }
\captionof{figure}{\textbf{Single-person pose interaction generation} on Motion-X dataset~\cite{lin2024motion}. Left shows the single person image input, right shows the generated two-person interaction dynamics. The generated interactive pose frame is labeled in \textcolor{magenta}{magenta box}. The bottom row show the single-pose input with accompanying text input. Given different single-person poses, our interactive pose generator produces plausible interactive poses under  flexible conditions, while our interactive pose animator synthesizes realistic human-human motions. Our model demonstrates strong performance in challenging in-the-wild settings.}
\label{fig:supp-motionx-results}
\end{table*}
\begin{table*}
    \centering
    \footnotesize
    \setlength{\tabcolsep}{0.2em} %
    \resizebox{\linewidth}{!}{
        \begin{tabular}{c|cccc}

        {Interactive Pose} & 
        \multicolumn{4}{c}{{Interaction Animation (left$\rightarrow$right: time steps)}} \\

         \viscontact{vis_contact/vis_interx}{G031T000A004R022}{10}{15}{20}{25}[2cm 2cm 2cm 2cm]
         
         \viscontact{vis_contact/vis_interx}{G040T003A019R025}{10}{15}{20}{25}[2cm 2cm 2cm 2cm]

        \multicolumn{5}{c}{{Inter-X Dataset~\cite{xu2024inter}}} \\

         \viscontact{vis_contact/vis_dh}{pair01_000242_1}{10}{15}{20}{25}
         
        \viscontact{vis_contact/vis_dh}{pair09_000520_1}{10}{15}{20}{25}

        \multicolumn{5}{c}{{Dual-Human Dataset~\cite{fang2024capturing}}} \\

         \viscontact{vis_contact/vis_dd100}{Foxtrot_006_000_split0_00}{10}{15}{20}{25}[2cm 2cm 2cm 2cm]
         
        \viscontact{vis_contact/vis_dd100}{Ballet_005_003_split2_00}{10}{15}{20}{25}[3cm 3cm 3cm 3cm]

        \multicolumn{5}{c}{{Duolando Dataset~\cite{siyao2024duolando}}} \\

    \end{tabular}
    }
\captionof{figure}{\textbf{More interactive pose animation visualization} on Inter-X dataset~\cite{xu2024inter}, Dual-Human dataset~\cite{fang2024capturing}, Duolando dataset~\cite{siyao2024duolando}. Our pose animator generalizes well to out-of-domain interactive poses and synthesizes realistic dancing motions on the unseen Duolando two-person dancing motion dataset.}
\label{fig:supp-animation-results}
\end{table*}
\begin{table*}
    \centering
    \footnotesize
    \setlength{\tabcolsep}{0.2em} %
    \resizebox{\linewidth}{!}{
        \begin{tabular}{c|cccc}

        {Interactive Pose} & 
        \multicolumn{4}{c}{{Interaction Animation (left$\rightarrow$right: time steps)}} \\
    
        \viscontact{vis_contact/vis_interhuman}{2064}{10}{15}{20}{25}[6cm 6cm 6cm 6cm]
        \viscontact{vis_contact/vis_interhuman}{6596}{10}{15}{20}{25}        

    \end{tabular}
    }
\captionof{figure}{\textbf{Interhuman dataset ~\cite{liang2024intergen} interactive pose animation results}. We convert dataset provided SMPLH~\cite{loper2015smpl} to SMPLX~\cite{SMPL-X:2019} representation and select interactive poses from test motion sequences. Despite contact inaccuracies due to dataset conventions and pose variations, our model synthesizes reasonable motions, demonstrating the strong generalization capability of interactive poses for guiding human interaction animation.}
\label{fig:exp-interhuman-results}
\end{table*}
\begin{table*}
    \centering
    \footnotesize
    \setlength{\tabcolsep}{0.2em} %
    \resizebox{0.95\linewidth}{!}{
        \begin{tabular}{cc|cccc}

        & {Interactive Pose} & 
        \multicolumn{4}{c}{{Interaction Animation (left$\rightarrow$right: time steps)}} \\

         \rotatebox{90}{\hspace{0pt}InterGen~\cite{liang2024intergen}} & \viscontact{vis_contact/vis_intergen}{G040T003A019R025}{10}{15}{20}{25}

        \rotatebox{90}{\hspace{0pt}W/O anchor } & 
         \viscontact{vis_contact/vis_random}{G040T003A019R025}{10}{15}{25}{29}[5cm 3cm 0cm 3cm] 

        \rotatebox{90}{\hspace{10pt}Ours} &
         \viscontact{vis_contact/vis_interx}{G040T003A019R025}{10}{15}{20}{25}[5cm 5cm 0cm 5cm]

    \end{tabular}
    }
\captionof{figure}{\textbf{Interactive pose animation comparison} on Inter-X dataset~\cite{xu2024inter}. Compared to InterGen~\cite{liang2024intergen} and model trained with random poses, our method achieves better contact and human dynamics. Both baselines exhibit severe body penetration and less accurate contact, while our approach, guided by interactive poses, ensures more realistic interactions.}
\label{fig:exp-contact-compare-results}
\end{table*}
\begin{table*}
    \centering
    \footnotesize
    \setlength{\tabcolsep}{0.2em} %
    \resizebox{\linewidth}{!}{
        \begin{tabular}{ccccc}

         \vistm{vis_t2m/vis_interx}{G022T003A017R013}{05}{10}{15}{20}{25}[5cm 7cm 2cm 5cm]

        \multicolumn{5}{c}{Input Text: One person chases the other person} \\
         
        \vistm{vis_t2m/vis_interx}{G035T001A009R025}{05}{10}{15}{20}{25}

        \multicolumn{5}{c}{Input Text: One person sits down first, another sits on his/her lap} \\

        \vistm{vis_t2m/vis_interx}{G038T003A018R018}{05}{10}{15}{20}{25}[0cm 3cm 0cm 0cm]

         \multicolumn{5}{c}{Input Text: One person goes to the other person's ear and whispers to him/her} \\

             \vistm{vis_t2m/vis_chi3d}{s02_Handshake_10}{05}{10}{15}{20}{25}

        \multicolumn{5}{c}{Input Text: hand shake} \\
         
        \vistm{vis_t2m/vis_chi3d}{s04_Hug_5}{05}{10}{15}{20}{25}[0cm 2cm 0cm 0cm]

        \multicolumn{5}{c}{Input Text: hug} \\

        \vistm{vis_t2m/vis_chi3d}{s04_Posing_5}{05}{10}{15}{20}{25}

        \multicolumn{5}{c}{Input Text: posing} \\

    \end{tabular}
    }
\captionof{figure}{\textbf{More text-to-interaction motion synthesis results}. Our method synthesizes realistic two-person interactions from short phrases or single words.}
\label{fig:exp-text-results}
\end{table*}
\begin{table*}
    \centering
    \footnotesize
    \setlength{\tabcolsep}{0.2em} %
    \resizebox{\linewidth}{!}{
        \begin{tabular}{c|cccc}

        {Single Pose} & 
        \multicolumn{4}{c}{{Interaction Generation (left$\rightarrow$right: time steps)}} \\

         \vispm{vis_p2m/vis_interx}{G015T004A024R023}{10}{15}{20}{25}[5cm 2cm 5cm 2cm]
         
        \vispm{vis_p2m/vis_interx}{G026T003A016R008}{10}{15}{20}{25}[5cm 5cm 5cm 5cm]

        \vispm{vis_p2m/vis_interx}{G028T005A023R004}{10}{15}{20}{25}

        \multicolumn{5}{c}{{Inter-X Dataset~\cite{xu2024inter}}} \\

         \vispm{vis_p2m/vis_dh}{pair00_000019_1}{10}{15}{20}{25}[0cm 2cm 0cm 0cm]
         
        \vispm{vis_p2m/vis_dh}{pair20_000760_1}{10}{15}{20}{25}[0cm 2cm 0cm 0cm]

         \vispm{vis_p2m/vis_dh}{pair33_001172_1}{10}{15}{20}{25}

        \multicolumn{5}{c}{{Dual-Human Dataset~\cite{fang2024capturing}}} \\

    \end{tabular}
    }
\captionof{figure}{\textbf{Single-pose guided interaction motion synthesis result} on Inter-X~\cite{xu2024inter} and Dual-Human~\cite{fang2024capturing} datasets. The input single-person pose is shown on the left. Our method generates appropriate interactive poses from various inputs, capturing vivid underlying human dynamics.}
\label{fig:exp-pose-results}
\end{table*}

\vspace{-1em}
\noindent\textbf{Inter-person penetrations.}
While our method enhances contact in two-person interactions, it does not explicitly model interpenetration between individuals. Consequently, in close-contact scenarios—such as the first row in \cref{fig:exp-limitation}—some interpenetration may occur in the generated motion sequences. Achieving a balance between realistic contact and preventing interpenetration remains a challenging problem, as enforcing strict physical constraints could compromise natural motion quality. Addressing interpenetration modeling and ensuring physically plausible two-person interaction motion generation is an important direction for future work.

\noindent\textbf{Lack of scene awareness.}
When applied to in-the-wild two-person pose animation or motion generation, our method relies solely on human pose information and ignores the surrounding environment. As a result, generated motions may appear physically implausible in certain cases, such as the 2nd row of \cref{fig:exp-limitation}, where collisions occur. Moreover, interactive poses can sometimes be ambiguous, causing noticeable motion errors when used as the sole input. A more robust approach would integrate additional scene information (e.g. image features) to improve motion prediction and dynamics forecasting.

\noindent\textbf{Inaccurate contact.} 
The interactive pose estimator or our interactive pose generator may occasionally produce inaccurate interactive poses, resulting in poor human–human contact in the generated motions, as seen in the 3rd and 4th rows of \cref{fig:exp-limitation}. These inaccuracies result in unrealistic motion due to the lack of precise interactive pose inputs. Since the pose animator primarily models temporal dynamics and depends on the interactive pose for spatial information, it often cannot correct errors arising from inaccurate interactive poses. Additionally, our generated interaction motions may exhibit artifacts such as foot sliding, a common issue in human motion synthesis. While such artifacts can often be mitigated through post-processing, we do not apply any post-processing in our examples.

\section{Qualitative results}
\label{sec:qualitative}

\noindent\textbf{Longer interactive motion generation.}
Our framework is designed for short-term interaction generation but naturally extends to longer sequences. The pose animator takes an interactive pose together with an interactive time to synthesize both past and future motions centered on that pose. Longer sequences are produced by chaining segments in a sliding-window manner: the last generated pose of one segment is reused as the starting pose for the next, the interactive time index is reset to zero (beginning of the new segment), and generation continues. Repeating this process yields coherent long-term interactions, as shown in \cref{fig:exp-c2m-long}, where key-frames are labeled in \textcolor{magenta}{magenta box}.

\noindent\textbf{Complex interactive pose animation.} 
As shown in \cref{fig:exp-contact-complex}, beyond daily motions, our pose animator can synthesize complex interactive motions involving high dynamics (1st row) and close contact (2nd row) between two people, benefiting from the strong interaction dynamics learned from high-quality mocap data.

\noindent\textbf{Two person image human motion animation.}
We provide additional in-the-wild interactive pose animation results in \cref{fig:supp-buddi-results}. Given an interactive frame, we extract two-person poses using an off-the-shelf model~\cite{muller2024generative}, and animate the them with our interactive pose animator. To render the interaction, we use an off-the-shelf inpainting model~\cite{suvorov2022resolution} to remove the original individuals and overlay the generated motion. The results demonstrate that our model generalizes well to in-the-wild interactive poses, producing realistic human-human interactions.

\noindent\textbf{Single-person image human motion interaction generation.}
We present additional single-person image interaction motion generation results on the Motion-X dataset~\cite{lin2024motion} in \cref{fig:supp-motionx-results}. Given a single-person image, we first extract the pose using an off-the-shelf pose estimator~\cite{cai2023smplerx} and then generate interactive poses with our interactive pose generator. As shown, our model synthesizes plausible interactions from diverse single-person inputs. Finally, we apply our interactive pose animator to generate two-person dynamics, demonstrating its effectiveness in challenging in-the-wild scenarios.

\noindent\textbf{Interactive pose animation.}
We provide additional visualizations of interactive pose animation on the Inter-X dataset~\cite{xu2024inter}, Dual-Human dataset~\cite{fang2024capturing}, and Duolando dataset~\cite{siyao2024duolando} in \cref{fig:supp-animation-results}. Our model could successfully synthesize realistic dancing motions from out-of-domain interactive poses on the unseen Duolando dataset.

We further evaluate our method on the InterHuman dataset~\cite{liang2024intergen}, a more challenging out-of-distribution benchmark, with results shown in \cref{fig:exp-interhuman-results}. InterHuman provides SMPLH~\cite{loper2015smpl} annotations for two-person interactions, primarily for text-to-motion generation, but with less accurate contact. To fit our framework, we convert the SMPLH~\cite{loper2015smpl} representation to SMPLX~\cite{SMPL-X:2019} and extract interactive poses from the test sequences. Despite annotation noise and diverse pose distributions, our model produces realistic and coherent interactions, demonstrating strong generalization of the interactive pose prior.

We also provide a qualitative comparison with two baselines—InterGen* and the random-pose variant (see \cref{table:exp-contact-ablation})—in \cref{fig:exp-contact-compare-results}. InterGen~\cite{liang2024intergen} and the random-pose model exhibit poorer contact and more body penetration than ours, highlighting the effectiveness of interactive pose priors for realistic contact and interaction synthesis.

\noindent\textbf{Text-to-interaction synthesis.}
We present additional text-to-interaction motion synthesis results in \cref{fig:exp-text-results}. Our method effectively generates realistic two-person interactions from short phrases or simple words. By leveraging an intermediate interactive pose representation, our approach ensures consistent interaction and maintains accurate contact between the two individuals.

\noindent\textbf{Single pose-to-interaction motion synthesis.}
We present single pose-to-interaction motion synthesis results on the Inter-X~\cite{xu2024inter} and Dual-Human~\cite{fang2024capturing} datasets in \cref{fig:exp-pose-results}. As shown, our method generates appropriate interactive poses from various input poses while effectively capturing vivid underlying human dynamics.

\end{document}